\documentclass[10pt]{article}
\usepackage[letterpaper,top=0.75in,bottom=1in,left=0.75in,right=0.75in,columnsep=0.25in]{geometry}
\usepackage[utf8]{inputenc}
\usepackage[T1]{fontenc}
\usepackage[hyphens]{url}
\usepackage{graphicx}
\usepackage[numbers,sort&compress]{natbib}
\usepackage{caption}
\usepackage{amsmath}
\usepackage{amssymb}
\usepackage{multirow}
\usepackage{tabularx}
\usepackage{makecell}
\usepackage{bbding}
\usepackage[table]{xcolor}
\usepackage{subcaption}
\usepackage{booktabs}
\usepackage{adjustbox}

\renewcommand{\abstractname}{Abstract}
\newenvironment{wideabstract}{
  \begin{center}
  \bfseries \abstractname
  \end{center}
  \begin{quote}
}{
  \end{quote}
}
\newcommand{\cmark}{\scalebox{0.7}{\Checkmark}}
\newcommand{\xmark}{\scalebox{0.7}{\XSolid}}
\newcommand{\cascaptionof}[2][]{
  \captionof{figure}{#2}
  \if\relax\detokenize{#1}\relax\else\label{#1}\fi
}

\newcommand{\aqf}{air quality forecasting}

\newcommand{\LStream}{Structural Memory Stream}
\newcommand{\RStream}{Transient Dynamics Stream}
\newcommand{\RSAbb}{TDS}
\newcommand{\LSAbb}{SMS}
\newcommand{\Norm}{Adaptive Normalization}

\title{AirFlow: Context Preserving and Multi-Rate State Modeling\\ for Air Quality Forecasting}
\author{
Fan Yang\thanks{Equal contribution.}
\and Nan Chen\footnotemark[1]
\and Yijie Dong
\and Yuchen Zhang
\and Wei Zhang\thanks{Corresponding author: \texttt{cstzhangwei@zju.edu.cn}}\\
School of Software Technology, Zhejiang University\\
Hangzhou, Zhejiang 310000, China
}
\date{}
\begin{document}
\twocolumn[
\maketitle
\begin{wideabstract}
Accurate air quality forecasting is essential for public health and urban environmental management, but remains challenging because pollutant channels differ in periodicity and distribution drift, while their concentration trajectories contain both multi-scale dependencies and rapid changes. Recent methods have improved spatial dependency learning and meteorological covariate modeling. However, pollutant channels are still passed through the same normalization rule and temporal backbone, using a shared latent representation for channel-specific distributions and changes at different rates. To address this limitation, we propose AirFlow, a pollutant-aware dual-stream framework that operates on station multivariate observations without additional graph propagation or predefined signal decomposition. Specifically, AirFlow designs two novel blocks: (1) a statistic-guided normalization routing mechanism that selects a normalization path for each pollutant according to its 24-hour autocorrelation and distribution drift; and (2) a hierarchical dual-stream state model that combines multi-scale state space propagation with learnable response coefficients, where gated bidirectional cross-attention exchanges information and adaptively fuses the resulting representations. Experiments on real-world data from multiple cities show that AirFlow achieves the best performance in 34 of 36 metrics comparisons, with reductions of up to 11.11\% root mean square error over the state-of-the-art baseline. AirFlow also requires only 0.0483M parameters and 0.0215G FLOPs, achieving high forecasting accuracy with low computational overhead.
\end{wideabstract}
\vspace{1em}
]

\section{Introduction}
Air pollution remains a major concern for urban environmental management, public health protection, and early warning systems \cite{brunekreef2002air, zhang2012realtime}. Fine particulate matter and gaseous pollutants are closely associated with respiratory and cardiovascular diseases \cite{de2022ambient, cohen2017burden}, while severe pollution episodes increase the burden on healthcare systems and emergency response \cite{krupnick1991controlling}. Reliable \aqf{} is therefore essential for proactive risk assessment and policy intervention.

Spatial modeling remains valuable when inter-station transport can be supported by synchronized wind fields, emission inventories, terrain attributes, and boundary-layer conditions. In many operational monitoring systems, however, these variables are unavailable, coarsely sampled, or temporally misaligned with pollutant observations \cite{tai2010correlations, ding2009transport}. Under this setting, coordinates and historical correlations provide useful but incomplete spatial priors, because the effective influence between stations varies with atmospheric transport and emission conditions. Increasing the complexity of graph propagation therefore does not necessarily resolve the forecasting errors caused by unobserved environmental changes. These unobserved changes manifest in the measured trajectories as abrupt peaks, rapid rises or drops, and shifts in level and variability. Such changes are difficult because a forecasting model must preserve the historical context preceding a pollution episode while responding rapidly once the concentration regime changes. We therefore study a complementary question to increasingly elaborate spatial modeling: how to improve pollutant-aware temporal state evolution when transport context is incomplete.
\begin{figure*}[t]
  \centering
\includegraphics[width=0.8\textwidth]{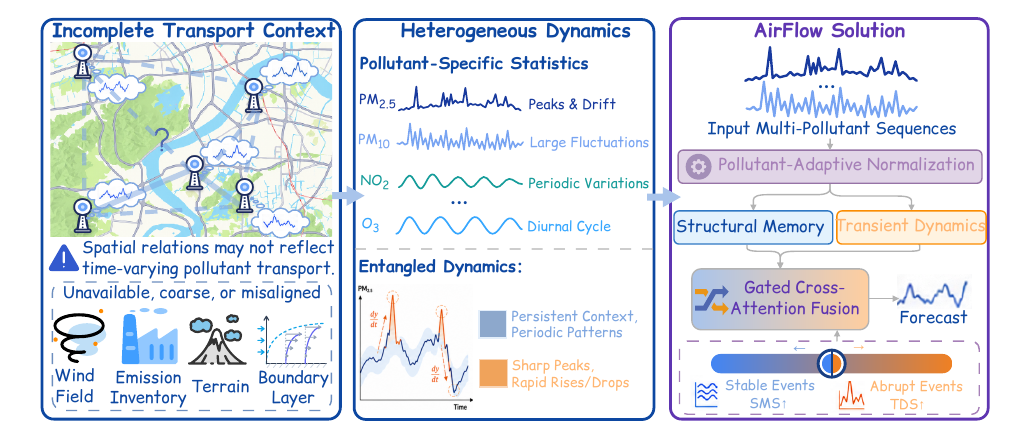}
  \caption{Motivation and framework of AirFlow for multivariate \aqf{}.}
  \label{fig:teaser}

\end{figure*}
This forecasting scenario poses two related modeling challenges. First, different pollutants exhibit distinct statistical characteristics. PM$_{2.5}$ and PM$_{10}$ are more susceptible to local emissions, dust, accumulation, and rapid removal, resulting in abrupt peaks, baseline drift, and non-stationary fluctuations \cite{huang2014secondary}. In contrast, NO$_2$ often shows traffic activity-related periodicity, while O$_3$ presents diurnal variation. Applying a shared normalization strategy may preserve useful magnitude relations for some variables while compressing local variations or amplifying distribution shifts in others \cite{revin, liu2022nonstationary}. Second, the temporal behavior of each pollutant is nonuniform. Periodic activity and gradual accumulation generate persistent dependencies \cite{gong2022multi}, whereas changes in emissions, dispersion, and removal conditions produce sharp rises, rapid drops, and short-lived peaks \cite{liu2025impacts, chen2020influence}. Accurate forecasting therefore requires both stable historical context and flexible responses to changing concentration conditions \cite{gu2022s4, lnn}. Figure~\ref{fig:teaser} summarizes the forecasting scenario, the modeling challenges, and the overall design of AirFlow.

Existing forecasting methods address these difficulties only partially \cite{geoman}. General sequence models usually process all pollutants using shared normalization and single state-update mechanism, limiting their ability to accommodate variable-specific statistics and nonuniform changes \cite{informer}. Decomposition-based methods separate sequences into trend, seasonal, frequency, or multi-scale components \cite{wu2021autoformer}, but their predefined or shared decomposition bases may not adapt well to pollutant concentration distributions, while component-wise processing may represent abrupt variations as residual or high-frequency parts without sufficiently preserving their connection to the preceding pollution state. These limitations motivate an adaptive framework that accommodates pollutant-specific distributions while combining contextual dependency modeling with state transitions.

AirFlow addresses air quality forecasting under incomplete transport context by coordinating two levels of pollutant heterogeneity: statistic-guided normalization preserves channel-dependent distribution information, while complementary state-transition pathways model persistent context and heterogeneous response rates without predefined decomposition or graph propagation. The main contributions of this work are summarized as follows:
\begin{itemize}
    \item  We propose AirFlow, a lightweight pollutant-aware framework that couples statistic-guided normalization routing with multi-rate latent state modeling. The routing mechanism preserves cross-window concentration references for statistically stable channels while reducing local level and scale shifts for drifting channels, avoiding a uniform preprocessing assumption across pollutants.
     \item  We design a novel hierarchical dual stream state model that combines multi-scale dependency modeling with heterogeneous hidden state transitions. Gated bidirectional cross-attention conditions rapid-response features on relevant historical context before adaptive fusion.
     \item Experiments across multiple cities and pollutants show accuracy gains and competitive computational overhead without additional graph propagation.
\end{itemize}

\begin{figure*}[!t]
  \centering
\includegraphics[width=0.8\textwidth]{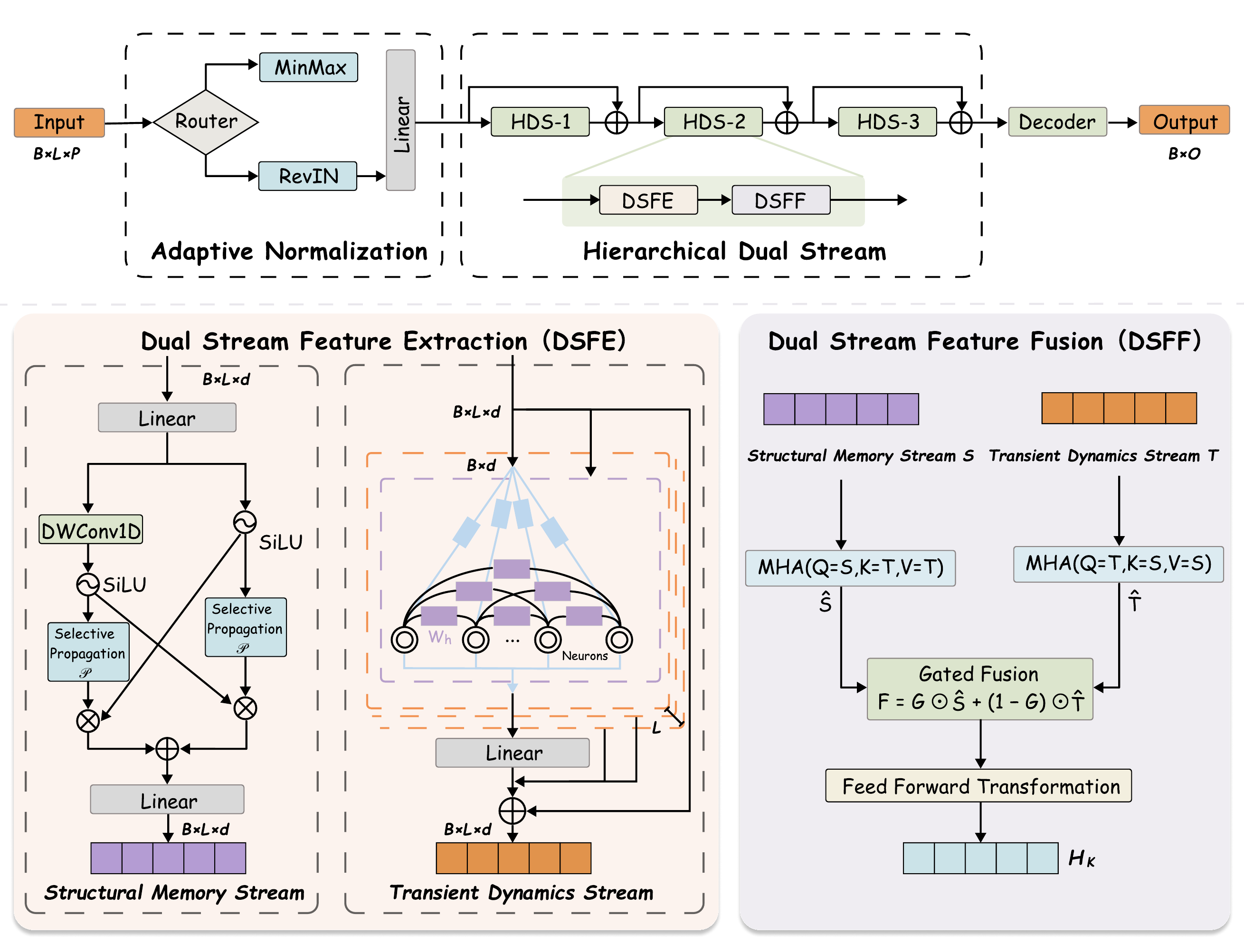}
  \caption{Overall architecture of AirFlow. \Norm{} handles heterogeneous statistics via statistic-guided router. Stacked Hierarchical Dual Stream (HDS) composed of Dual Stream Feature Extraction (DSFE) and Dual Stream Feature Fusion (DSFF). DSFE learns complementary representations through \LStream{} and \RStream{}, while DSFF fuses them via gated cross-attention. The final representations are decoded for forecasting.}
  \label{fig:overview}
\end{figure*}

\section{Related Work}
\subsection{Air Quality Forecasting}
Spatial inductive biases are widely adopted in air-quality forecasting to represent interactions across monitoring locations. PM$_{2.5}$-GNN incorporates geographical knowledge into graph construction, whereas GAGNN models dependencies at both city and city-group levels \cite{pm25gnn,gagnn}. AirFormer develops scalable spatial attention for time-varying correlations over a nationwide monitoring network \cite{airformer}. DGN-AEA further constructs dynamic relations from adaptive edge attributes \cite{dgnaea}, while DSGT constructs them from time-varying station observations and auxiliary features \cite{dsgt}. MGSFformer combines multi-granularity observations with spatiotemporal correlations \cite{mgsfformer}. These methods improve spatial representation through increasingly flexible relation construction, while the learned dependencies remain tied to the signals used to infer inter-station relations.

Beyond learned spatial relations, recent studies incorporate atmospheric knowledge into the forecasting process. MasterGNN processes air quality and weather stations modeling and suppresses observation noise accumulation caused by spatial-temporal propagation \cite{mastergnn}. CauAir models the causal association between weather covariates and Air Quality Index (AQI) \cite{cauair}. Physics-guided methods further encode transport dynamics. AirPhyNet represents diffusion and advection with graph-based differential equation networks, Air-DualODE combines open system physical dynamics with a data-driven Neural ordinary differential equation (ODE) branch, and AirDDE models propagation delays through multifactor historical features and delay function grounded in diffusion and advection processes \cite{airphynet,airdualode,airdde}. These studies demonstrate that spatial and meteorological context provides valuable information for \aqf{}. However, recent advances have emphasized sophisticated spatial dependency modeling, while the heterogeneous temporal patterns of pollutant concentrations remain underexplored. Additional spatial dependencies enrich contextual information, but they may also compensate for, rather than resolve, insufficient modeling of abrupt changes and non-stationary patterns. This motivates us to strengthen pollutant-aware temporal modeling and verify whether competitive or better performance can be achieved without constructing additional graph propagation.

\subsection{Normalization and Temporal Modeling}
Normalization methods address non-stationarity by controlling which statistics are removed, estimated, and restored. Reversible Instance Normalization (RevIN) applies reversible instance mean and variance normalization, whereas SAN estimates evolving statistics over local temporal slices \cite{revin,san}. SIN selects statistics according to local invariance and global variability \cite{sin}; FAN and DDN extend normalization to frequency-domain and dual-domain distribution changes \cite{fan,ddn}. These methods show that normalization choices embody different assumptions about distribution change, motivating pollutant-wise routing rather than one fixed operator across channels.

Temporal forecasting models capture sequence dependencies through different computational mechanisms. Convolutional neural network (CNN)-based models extract local patterns with temporal convolutions, whereas recurrent neural network (RNN)-based models propagate historical information through recurrent hidden states \cite{timesnet}. Transformer-based models capture long-term interactions with attention \cite{informer}, while selective state space models provide input-dependent state propagation with linear sequence complexity \cite{mamba,mamba2}. Liquid neural network (LNN)-based models regulate hidden-state transitions through differential dynamics \cite{lnn,cfc}. To address non-stationarity and temporal patterns at different scales, recent models further introduce adaptive decomposition, multi-resolution processing, and frequency-domain representations \cite{amd,timekan,timeemb, sempo}. These works suggest that component construction and latent-state modeling are complementary design strategies. AirFlow follows the latter to model multi-scale context and rate-varying changes without relying on a fixed component partition.
\vspace{3mm}
\section{Methodology}
\label{sec:method}

\subsection{Problem Formulation}
\label{sec:overview}

Let $\mathcal{S}$ denote the set of monitoring stations. For station
$s\in\mathcal{S}$ and forecasting origin $t$, the input
$\mathbf{X}_{s,t}\in\mathbb{R}^{L\times P}$ contains $L$ observations
of $P$ pollutants collected at the same station. For a fixed target
pollutant, AirFlow learns
\begin{equation}
\widehat{\mathbf{y}}_{s,t}
=
f(\mathbf{X}_{s,t};\boldsymbol{\Theta})
\in\mathbb{R}^{O},
\label{eq:problem}
\end{equation}
where $O$ is the forecasting horizon and $\boldsymbol{\Theta}$ denotes
the model parameters. Training sequences are generated by sliding a
window over each station, and the model is shared across all stations.
The target pollutant is fixed in each training run.

AirFlow is designed for two properties of multi-pollutant
observations. Pollutants differ in periodicity and distribution shift,
making a single normalization rule unsuitable for all channels. Their trajectories also contain long-range
dependencies and rapid variations that favor different state-update
patterns. AirFlow therefore combines station pollutant
normalization with two temporal streams. As shown in
Figure~\ref{fig:overview}, the normalized sequence is projected to a
$d$-dimensional latent space and passed through $K$ Hierarchical
Dual-Stream (HDS) blocks. Each block contains \LStream{} (\LSAbb{}), \RStream{}(\RSAbb{}), and Gated
Bidirectional Cross-Attention (GBCA) module. A prediction head maps the
final representation to the future target sequence.

\subsection{Station-Wise Pollutant Normalization}
\label{sec:normalization}
For pollutant $p$ at station $s$, let
$\mathbf{x}_{s,p}\in\mathbb{R}^{T}$ denote its training sequence,
where $T$ is the sequence length. The periodicity score is
\begin{equation}
\rho_{s,p}
=
\operatorname{Corr}
\left(
\mathbf{x}_{s,p}[1{:}T-h],
\mathbf{x}_{s,p}[1+h{:}T]
\right),
\label{eq:periodicity}
\end{equation}
where $h$ is the lag of one diurnal cycle and
$\operatorname{Corr}$ denotes the Pearson correlation coefficient \cite{pearson1896}.
The drift-rate score is
\begin{equation}
\nu_{s,p}
=
\frac{
\operatorname{Std}
\left(
\operatorname{MA}_{w}(\mathbf{x}_{s,p})
\right)
}{
\operatorname{Std}(\mathbf{x}_{s,p})+\varepsilon
},
\label{eq:slow_variation}
\end{equation}
where $\operatorname{MA}_{w}$ is the moving average with weekly window length
$w$, $\operatorname{Std}$ denotes standard deviation, and
$\varepsilon>0$ prevents division by zero. The score
$\nu_{s,p}$ summarizes the relative variation of the smoothed
sequence.

The normalization route is determined for each station--pollutant pair:
\begin{equation}
r_{s,p}
=
\begin{cases}
\mathrm{RevIN},
&
\rho_{s,p}<\theta_{\rho}
\ \land\
\nu_{s,p}\geq\theta_{\nu},\\
\mathrm{MinMax},
&
\text{otherwise},
\end{cases}
\label{eq:routing}
\end{equation}
where $\theta_{\rho}$ and $\theta_{\nu}$ are routing thresholds. The rule gives priority to periodicity because instance-wise centering may weaken the cross-window level information associated with a stable recurring pattern.

For input window $i$ from station $s$, let
$\mathbf{x}_{s,i,p}\in\mathbb{R}^{L}$ denote pollutant $p$. Its
normalized sequence is
\begin{equation}
\widetilde{\mathbf{x}}_{s,i,p}
=
\left\{
\begin{array}{ll}
\dfrac{
\mathbf{x}_{s,i,p}-x_{\min,s,p}
}{
x_{\max,s,p}-x_{\min,s,p}+\varepsilon
},
&
r_{s,p}=\mathrm{MinMax},\\[10pt]
\gamma_p
\dfrac{
\mathbf{x}_{s,i,p}-\mu_{s,i,p}
}{
\sigma_{s,i,p}
}
+\beta_p,
&
r_{s,p}=\mathrm{RevIN},
\end{array}
\right.
\label{eq:normalization}
\end{equation}
where $x_{\min,s,p}$ and $x_{\max,s,p}$ are the station 
training extrema, $\mu_{s,i,p}$ and $\sigma_{s,i,p}$ are the mean and
standard deviation of the input window, $\gamma_p$ and $\beta_p$
are learnable affine parameters. MinMax uses the station-specific training range for every input window, preserving absolute concentration levels and cross-window comparability. RevIN standardizes each window with its own mean and variance, reducing local level and scale shifts \cite{revin}. AirFlow prefers MinMax when a stable periodic pattern provides a reliable cross-window reference, and uses RevIN when weak periodicity is accompanied by stronger slow distribution movement. The normalized channels are concatenated
and projected to the initial hidden representation
$\mathbf{H}_0\in\mathbb{R}^{n\times L\times d}$, where $n$ is the
batch size and $d$ is the hidden dimension. The corresponding target statistics are retained for
inverse transformation. 

\subsection{Hierarchical Dual-Stream Encoder}
\label{sec:hds}
AirFlow stacks $K$ residual HDS blocks to refine the
latent representation. Let $\mathbf{H}_k$ denote the input to the
$k$-th block, with $\mathbf{H}_0$ produced by the input projection and
$k=0,\ldots,K-1$. 

\subsubsection{Structural Memory Stream}
\label{sec:sms}

A learned projection of $\mathbf{H}_{k}$ produces two representations. The convolution-enhanced representation is
$\mathbf{U}=\operatorname{SiLU}
(\operatorname{DWConv}(\Pi_u(\mathbf{H}_{k})))$,
and the context-preserving representation is
$\mathbf{Z}=\operatorname{SiLU}
(\Pi_z(\mathbf{H}_{k}))$, where $\Pi_u$ and $\Pi_z$ are learned projections and
$\operatorname{DWConv}$ is depth-wise temporal convolution. The selective propagation operator $\mathcal{P}$ is applied to both representations, that is  $\mathbf{U}'=\mathcal{P}(\mathbf{U})$ and
$\mathbf{Z}'=\mathcal{P}(\mathbf{Z})$. To define the operator, let $\mathbf{V} \in \{\mathbf{U}, \mathbf{Z}\}$ denote either input representation. The input at position $j$ generates a
positive discretization vector
$\boldsymbol{\Delta}_j\in\mathbb{R}^{d_e}$ and vectors
$\mathbf{B}_j,\mathbf{C}_j\in\mathbb{R}^{d_s}$, where $d_e$ is the
expanded width and $d_s$ is the state dimension. Let
$\mathbf{A}\in\mathbb{R}^{d_e\times d_s}$ denote the transition
coefficients, with
$\mathbf{A}=-\exp(\mathbf{A}_{\log})$. We define
$\overline{\mathbf{A}}_j
=\exp(\operatorname{Diag}(\boldsymbol{\Delta}_j)\mathbf{A})$
and
$\overline{\mathbf{B}}_j
=\boldsymbol{\Delta}_j\mathbf{B}_j^{\top}$.
The state
$\mathbf{M}_j\in\mathbb{R}^{d_e\times d_s}$ is updated by
\begin{equation}
\mathbf{M}_j
=
\overline{\mathbf{A}}_j\odot\mathbf{M}_{j-1}
+
\operatorname{Diag}(\mathbf{V}_j)\overline{\mathbf{B}}_j,
\label{eq:state_update}
\end{equation}
where $\operatorname{Diag}(\cdot)$ forms a diagonal matrix and
$\odot$ denotes element-wise multiplication. The propagated output is
\begin{equation}
\mathcal{P}(\mathbf{V})_j
=
\mathbf{M}_j\mathbf{C}_j
+
\mathbf{D}\odot\mathbf{V}_j,
\label{eq:state_readout}
\end{equation}
where $\mathbf{D}\in\mathbb{R}^{d_e}$ is the direct term. The product
$\mathbf{M}_j\mathbf{C}_j$ is matrix--vector multiplication. The
symbols $\mathbf{A}$, $\mathbf{B}$, $\mathbf{C}$, and $\mathbf{D}$
follow the standard state-space roles \cite{mamba,mamba2}.

The \LSAbb{} output is
\begin{equation}
\mathbf{S}
=
\mathbf{W}_{o}
\left[
\mathbf{U}'\odot\mathbf{Z}
+
\mathbf{Z}'\odot\mathbf{U}
\right],
\label{eq:sms_output}
\end{equation}
where $\mathbf{W}_{o}$ projects the expanded representation back to
width $d$. The symmetric interaction combines local filtering with
selective propagation without introducing a predefined decomposition.

\subsubsection{\RStream{}}
\label{sec:tds}

\RSAbb{} maintains a set of latent features with different learnable response scales. At position $j$, the candidate state is 
$\widetilde{\mathbf{h}}_j=
\tanh(\mathbf{W}_{x}\mathbf{H}_{k,j}
+\mathbf{W}_{h}\mathbf{h}_{j-1}+\mathbf{b}_{h})$, its response coefficients are computed as $\boldsymbol{\alpha} = \sigma(\boldsymbol{\eta} \oslash \boldsymbol{\tau})$, where $\boldsymbol{\eta}$ is learnable scalar, $\boldsymbol{\tau}$ is learnable vector, $\oslash$ is element-wise division, and $\sigma$ is sigmoid function. The state is updated by
\begin{equation}
\mathbf{h}_j = \mathbf{h}_{j-1} + \boldsymbol{\alpha} \odot(\widetilde{\mathbf{h}}_j-\mathbf{h}_{j-1}).
\label{eq:tds_update}
\end{equation}
Different latent features require different degrees of temporal persistence, some need to accumulate information over longer contexts, whereas others need to respond more rapidly to recent observations. A smaller entry of $\boldsymbol{\alpha}$ retains more of the preceding state, while a larger entry places greater weight on the current candidate state. Finally, the recurrent states are
projected to obtain
$\mathbf{T}\in\mathbb{R}^{n\times L\times d}$.

\subsubsection{Gated Bidirectional Cross-Attention}
\label{sec:gbca}

The GBCA module performs cross-stream interaction and adaptive
fusion. Specifically, the \LSAbb{} representation uses the \RSAbb{} representation as key and
value to retrieve recurrent state information, while the reverse
direction retrieves propagated context information:
\begin{equation}
\widehat{\mathbf{S}}
=
\operatorname{MHA}(\mathbf{S},\mathbf{T},\mathbf{T}),
\qquad
\widehat{\mathbf{T}}
=
\operatorname{MHA}(\mathbf{T},\mathbf{S},\mathbf{S}),
\end{equation}
where the arguments of Multi-Head Attention ($\operatorname{MHA}$) are query, key, and value \cite{transformer}.
This bidirectional retrieval allows each stream to be conditioned on
the other before their relative contributions are estimated by the token-wise gate $\mathbf{G}=\sigma(
\operatorname{MLP}
([\widehat{\mathbf{S}}\Vert\widehat{\mathbf{T}}]))$,
where $\Vert$ denotes concatenation. The fused representation is
\begin{equation}
\mathbf{F}
=
\mathbf{G}\odot\widehat{\mathbf{S}}
+
(\mathbf{1}-\mathbf{G})\odot\widehat{\mathbf{T}}.
\label{eq:gbca}
\end{equation}
The gate is estimated at each temporal position from the mutually
conditioned representations. The block output is
\begin{equation}
\mathbf{H}_{k+1}
=
\mathbf{H}_{k}
+
\Phi(\mathbf{F}),
\label{eq:block_output}
\end{equation}
where $\Phi$ denotes the feed forward transformation applied after fusion.

\subsection{Prediction}
\label{sec:objective}

After the final HDS block, the representation is decoded by mean pooling and two-layer prediction head $\mathcal{D}$:
\begin{equation}
\widehat{\mathbf{y}}
=
\mathcal{D}
\left(
\frac{1}{L}
\sum_{j=1}^{L}
\mathbf{H}_{K,j}
\right).
\label{eq:forecast_head}
\end{equation}
Training is performed in
the scale associated with the target route, and predictions are
transformed back to the original concentration scale for evaluation.

Multi-step \aqf{} requires both accurate
concentration values and consistent changes across adjacent
forecasting horizons. We therefore optimize a trajectory-aware
objective with a robust concentration term and an inter-step
variation term:
\begin{equation}
\mathcal{L}
=
\frac{1}{nO}
\sum_{i=1}^{n}
\sum_{o=1}^{O}
\psi(e_{i,o})
+
\lambda_{\Delta}\mathcal{L}_{\Delta},
\label{eq:objective}
\end{equation}
where
$e_{i,o}=\widehat{y}_{i,o}-y_{i,o}$ is the point-wise forecasting
error, and $\lambda_{\Delta}$ controls the variation term. The
penalty $\psi$ is:
\begin{equation}
\psi(e)
=
\begin{cases}
\lambda_{\mathrm{s}}
\dfrac{e^{2}}{2\kappa}
+
\lambda_{1}|e|,
&
|e|<\kappa,\\[6pt]
(\lambda_{\mathrm{s}}+\lambda_{1})|e|
-
\dfrac{\lambda_{\mathrm{s}}\kappa}{2},
&
|e|\geq\kappa,
\end{cases}
\label{eq:robust_penalty}
\end{equation}
where $\kappa>0$ is the transition point and
$\lambda_{\mathrm{s}},\lambda_{1}\geq0$ are fixed weights. This
penalty retains quadratic sensitivity to small residuals and grows
linearly for large residuals, preventing a small number of severe
pollution episodes from dominating optimization while maintaining a
direct penalty on absolute concentration deviations. Point-wise accuracy does not constrain how the
predicted trajectory changes across horizons. The inter-step variation term is
\begin{equation}
\mathcal{L}_{\Delta}
=
\frac{1}{n(O-1)}
\sum_{i=1}^{n}
\sum_{o=1}^{O-1}
\left(
\Delta\widehat{y}_{i,o}
-
\Delta y_{i,o}
\right)^2,
\label{eq:increment_loss}
\end{equation}
where $\Delta\widehat{y}_{i,o}
=
\widehat{y}_{i,o+1}-\widehat{y}_{i,o}$
and
$\Delta y_{i,o}
=
y_{i,o+1}-y_{i,o}$. 
This design couples adjacent forecasting horizons and penalizes
discrepancies in both the direction and magnitude of concentration
changes.
\vspace{3mm}
\section{Experiments}
\label{sec:experiments}

\subsection{Experimental Settings}
\noindent\textbf{Datasets}
We evaluate AirFlow on hourly air quality records from Beijing, Tianjin, and Hangzhou. All datasets are publicly available at \url{https://quotsoft.net/air}. After filtering stations with PM$_{2.5}$ missing rates below 20\%, the three datasets contain 35, 21, and 12 monitoring stations. Each dataset spans from Jan. 1, 2023 to Dec. 31, 2024. The data from each station are divided
chronologically into training, validation, and test sets with a ratio
of 8:1:1. Each station provides six pollutant measurements, including PM$_{2.5}$, PM$_{10}$, SO$_2$, NO$_2$, CO, and O$_3$. The four forecasting targets are listed in Table~\ref{tab:main}. PM$_{2.5}$ is treated as the primary forecasting target due to its critical health impacts.

\noindent\textbf{Baselines} We compare AirFlow with general sequence models and \aqf{} models. The former includes BiLSTM \cite{bilstm}, ConvLSTM \cite{convlstm},
Transformer \cite{transformer}, LNN \cite{lnn}, PatchTST \cite{patchtst}, iTransformer \cite{itransformer}, and Mamba \cite{mamba}, covering recurrent, convolutional, attention-based, and state-space architectures. The latter includes PM2.5-GNN \cite{pm25gnn}, AirFormer \cite{airformer}, STAAGCN \cite{staagcn}, AirPhyNet \cite{airphynet}, MGSFformer \cite{mgsfformer}, and Air-DualODE \cite{airdualode}, which introduce spatial dependency learning, atmospheric priors, or auxiliary
environmental context.

\noindent\textbf{Metrics}
Forecasts are evaluated in the original concentration scale using root mean square error (RMSE), Mean Absolute Error (MAE), and Coefficient of Determination (R$^2$). These metrics quantify the model's error magnitude, overall accuracy, and goodness of fit.

\subsection{Main Results}
\label{sec:main}
To demonstrate the effectiveness and robustness of AirFlow, we compared its performance with various state-of-the-art baselines. For fair comparisons, all baselines and AirFlow are evaluated under an identical experimental setup. As shown in Table~\ref{tab:main}, several interesting observations can be drawn. General-purpose sequence models exhibit limited performance. Although Mamba improves computational efficiency, it does not outperform recurrent baselines. The vanilla Transformer underperforms due to the lack of temporal inductive bias, whereas its variants (e.g., iTransformer and PatchTST) achieve better results by augmenting variable-wise dependencies or local temporal representation. LNN has relatively fewer neurons but is originally developed for causal decision-making under irregular sampling, thus failing to achieve competitive performance on this task.

In contrast, domain-specific models designed for spatiotemporal \aqf{} exhibit clear advantages over general-purpose sequence models. Especially the physics-guided model Air-DualODE retains RMSE advantage on Tianjin PM$_{2.5}$ and AQI. It is also worth noting that on the NO$_2$ forecasting task, where temporal fluctuations are driven by regular anthropogenic activities, the performance differences among most models remain relatively small.

AirFlow ranks first in 34 of the 36 comparisons while operating only on station-level multivariate observations. On the Beijing dataset, it reduces RMSE over the strongest baseline (Air-DualODE) by 5.26\% for PM$_{2.5}$, 5.24\% for PM$_{10}$, 1.48\% for NO$_2$, and 2.24\% for AQI. On the Hangzhou dataset, which represents a different geographic domain characterized by subtropical monsoon climate and high-tech manufacturing emissions, AirFlow reduces RMSE over the Air-DualODE by 6.05\%, 11.11\%, 1.63\%, and 7.71\% for PM$_{2.5}$, PM$_{10}$, NO$_2$, and AQI. These results indicate that in pollutant-only scenarios, prioritizing station pollutant dynamics can achieve strong forecasting accuracy without additional graph propagation.

\begin{table*}[!t]
    \centering
    \caption{Experimental results comparing AirFlow with baseline models across four target variables, averaged over three runs. Lower RMSE and MAE and higher R$^2$ indicate better performance. Best results are highlighted in \textbf{bold}.}
    \label{tab:main}
    \resizebox{\textwidth}{!}{
    \begin{tabular}{llcccccccccccc}
        \toprule
        \multirow{2}{*}{Dataset} & \multirow{2}{*}{Model} & \multicolumn{3}{c}{PM$_{2.5}$} & \multicolumn{3}{c}{PM$_{10}$} & \multicolumn{3}{c}{NO$_2$} & \multicolumn{3}{c}{AQI} \\
        \cmidrule(lr){3-5} \cmidrule(lr){6-8} \cmidrule(lr){9-11} \cmidrule(lr){12-14}
        & & RMSE $\downarrow$ & MAE $\downarrow$ & R$^2$ $\uparrow$ & RMSE $\downarrow$ & MAE $\downarrow$ & R$^2$ $\uparrow$ & RMSE $\downarrow$ & MAE $\downarrow$ & R$^2$ $\uparrow$ & RMSE $\downarrow$ & MAE $\downarrow$ & R$^2$ $\uparrow$ \\
        \midrule
        \multirow{14}{*}{\shortstack{Beijing}} 
        & BiLSTM (\citeyear{bilstm}) & 16.44 & 9.01 & 0.8113 & 28.89 & 15.06 & 0.7451 & 12.79 & 8.68 & 0.6823 & 22.14 & 12.08 & 0.7616 \\
        & ConvLSTM (\citeyear{convlstm}) & 16.48 & 9.08 & 0.8104 & 30.51 & 15.41 & 0.7157 & 12.76 & 8.70 & 0.6835 & 21.46 & 12.12 & 0.7761 \\
        & Transformer (\citeyear{transformer}) & 17.33 & 10.34 & 0.7903 & 27.13 & 16.18 & 0.7751 & 12.80 & 8.58 & 0.6814 & 21.16 & 11.83 & 0.7854 \\
        & PM2.5-GNN (\citeyear{pm25gnn}) & 15.84 & 8.42 & 0.8262 & 24.85 & 13.88 & 0.8182 & 12.82 & 8.89 & 0.6788 & 20.12 & 11.44 & 0.8096 \\
        & LNN (\citeyear{lnn}) & 17.02 & 8.67 & 0.7978 & 26.10 & 14.29 & 0.7919 & 12.83 & 8.59 & 0.6803 & 21.73 & 11.75 & 0.7703 \\
        & PatchTST (\citeyear{patchtst}) & 16.12 & 8.83 & 0.8186 & 30.29 & 14.03 & 0.7197 & 12.75 & 8.59 & 0.6872 & 20.09 & 11.43 & 0.7937 \\
        & AirFormer (\citeyear{airformer}) & 15.68 & 8.40 & 0.8284 & 26.45 & 13.96 & 0.7862 & 12.79 & 8.61 & 0.6815 & 21.60 & 11.30 & 0.7730 \\
        & STAAGCN (\citeyear{staagcn}) & 15.98 & 8.41 & 0.8132 & 24.81 & 13.90 & 0.8120 & 12.82 & 8.72 & 0.6806 & 20.29 & 11.48 & 0.7999 \\
        & iTransformer (\citeyear{itransformer}) & 15.88 & 8.58 & 0.8239 & 29.89 & 13.92 & 0.7270 & 12.95 & 8.78 & 0.6742 & 20.95 & 11.46 & 0.7866 \\      
        & Mamba (\citeyear{mamba}) & 16.64 & 8.62 & 0.8068 & 26.19 & 14.46 & 0.7904 & 12.81 & 8.64 & 0.6814 & 22.09 & 11.99 & 0.7626 \\
        & AirPhyNet (\citeyear{airphynet}) & 15.93 & 8.87 & 0.8127 & 33.18 & 18.32 & 0.7002 & 14.17 & 9.71 & 0.5915 & 20.49 & 11.49 & 0.8088 \\
        & MGSFformer (\citeyear{mgsfformer}) & 15.62 & 8.29 & 0.8297 & 25.42 & 14.12 & 0.8098 & 12.76 & 8.60 & 0.6822 & 20.32 & 11.39 & 0.8007 \\
        & Air-DualODE (\citeyear{airdualode}) & 15.22 & 8.18 & 0.8348 & 23.67 & 13.38 & 0.8310 & 12.84 & 8.88 & 0.6794 & 19.68 & 11.21 & 0.8108 \\
        & AirFlow (Ours) & \textbf{14.42} & \textbf{7.47} & \textbf{0.8513} & \textbf{22.43} & \textbf{13.02} & \textbf{0.8443} & \textbf{12.65} & \textbf{8.55} & \textbf{0.6889} & \textbf{19.24} & \textbf{10.87} & \textbf{0.8184} \\
        \midrule
        \multirow{14}{*}{\shortstack{Tianjin}} 
        & BiLSTM (\citeyear{bilstm}) & 20.09 & 11.91 & 0.6828 & 28.26 & 17.81 & 0.6717 & 14.79 & 9.87 & 0.6681 & 24.12 & 14.85 & 0.6670 \\
        & ConvLSTM (\citeyear{convlstm}) & 19.95 & 11.86 & 0.6867 & 28.46 & 18.40 & 0.6670 & 14.71 & 10.04 & 0.6716 & 23.74 & 14.62 & 0.6774 \\
        & Transformer (\citeyear{transformer}) & 21.67 & 13.08 & 0.6311 & 30.02 & 19.49 & 0.6296 & 14.96 & 9.91 & 0.6601 & 24.39 & 14.73 & 0.6595 \\
        & PM2.5-GNN (\citeyear{pm25gnn}) & 19.22 & 11.45 & 0.6897 & 27.88 & 17.12 & 0.6945 & 14.98 & 10.01 & 0.6623 & 23.02 & 13.98 & 0.6874 \\
        & LNN (\citeyear{lnn}) & 20.06 & 11.70 & 0.6839 & 29.19 & 18.34 & 0.6497 & 14.78 & 9.82 & 0.6687 & 24.07 & 14.47 & 0.6683 \\
        & PatchTST (\citeyear{patchtst}) & 19.63 & 11.78 & 0.6988 & 28.08 & 17.28 & 0.6760 & 14.66 & 9.85 & 0.6739 & 23.44 & 13.91 & 0.6856 \\
        & AirFormer (\citeyear{airformer}) & 20.39 & 11.89 & 0.6734 & 28.43 & 17.49 & 0.6678 & 14.59 & 9.86 & 0.6767 & 23.38 & 14.02 & 0.6871 \\
        & STAAGCN (\citeyear{staagcn}) & 19.87 & 11.66 & 0.6897 & 29.04 & 18.35 & 0.6533 & 14.85 & 10.09 & 0.6653 & 24.02 & 14.45 & 0.6692 \\
        & iTransformer (\citeyear{itransformer}) & 19.58 & 11.79 & 0.6921 & 28.40 & 17.91 & 0.6685 & 14.61 & 9.82 & 0.6762 & 24.69 & 14.77 & 0.6510 \\
        & Mamba (\citeyear{mamba}) & 19.82 & 11.68 & 0.6913 & 29.31 & 18.54 & 0.6468 & 15.05 & 10.11 & 0.6563 & 24.39 & 14.83 & 0.6595 \\
        & AirPhyNet (\citeyear{airphynet}) & 20.18 & 12.07 & 0.6799 & 29.98 & 18.98 & 0.6305 & 15.23 & 11.76 & 0.6488 & 23.88 & 14.58 & 0.6735 \\
        & MGSFformer (\citeyear{mgsfformer}) & 19.70 & 11.54 & 0.6966 & 27.89 & 17.11 & 0.6859 & 14.63 & 9.88 & 0.6772 & 23.06 & 14.01 & 0.6908 \\
        & Air-DualODE (\citeyear{airdualode}) & \textbf{18.78} & 10.98 & 0.6974 & 27.65 & 16.87 & 0.7001 & 15.11 & 10.44 & 0.6537 & \textbf{22.51} & 13.52 & 0.7012 \\
        & AirFlow (Ours) & 18.87 & \textbf{10.97} & \textbf{0.7124} & \textbf{26.87} & \textbf{16.81} & \textbf{0.7046} & \textbf{14.54} & \textbf{9.78} & \textbf{0.6793} & 22.58 & \textbf{13.48} & \textbf{0.7023} \\
        \midrule
        \multirow{14}{*}{\shortstack{Hangzhou}} 
        & BiLSTM (\citeyear{bilstm}) & 13.34 & 8.02 & 0.6865 & 18.45 & 11.72 & 0.6817 & 11.33 & 7.92 & 0.6107 & 16.64 & 9.94 & 0.6546 \\
        & ConvLSTM (\citeyear{convlstm}) & 13.96 & 8.75 & 0.6831 & 18.51 & 11.76 & 0.6789 & 11.69 & 8.22 & 0.5855 & 16.09 & 9.98 & 0.6769 \\
        & Transformer (\citeyear{transformer}) & 15.96 & 8.92 & 0.5654 & 19.77 & 12.17 & 0.6345 & 11.28 & 7.94 & 0.6146 & 17.34 & 10.37 & 0.6246 \\
        & PM2.5-GNN (\citeyear{pm25gnn}) & 15.23 & 9.86 & 0.6211 & 18.44 & 10.98 & 0.6812 & 11.54 & 8.18 & 0.6001 & 17.45 & 10.32 & 0.6311 \\
        & LNN (\citeyear{lnn}) & 14.22 & 8.83 & 0.6802 & 19.44 & 12.32 & 0.6468 & 11.12 & 7.76 & 0.6249 & 17.12 & 10.45 & 0.6341 \\
        & PatchTST (\citeyear{patchtst}) & 16.36 & 10.47 & 0.5439 & 22.92 & 14.96 & 0.5109 & 13.16 & 9.35 & 0.4729 & 20.05 & 12.78 & 0.5001 \\
        & AirFormer (\citeyear{airformer}) & 17.78 & 10.43 & 0.5075 & 24.57 & 15.20 & 0.4356 & 14.39 & 10.13 & 0.3710 & 22.95 & 13.59 & 0.3435 \\
        & STAAGCN (\citeyear{staagcn}) & 16.69 & 10.92 & 0.5255 & 24.74 & 16.45 & 0.4300 & 13.78 & 10.26 & 0.4227 & 20.54 & 13.85 & 0.4758 \\
        & iTransformer (\citeyear{itransformer}) & 17.04 & 11.01 & 0.5053 & 23.71 & 15.63 & 0.4768 & 14.03 & 9.62 & 0.4009 & 21.27 & 13.65 & 0.4378 \\
        & Mamba (\citeyear{mamba}) & 13.01 & 8.11 & 0.7012 & 18.43 & 11.32 & 0.6826 & 11.04 & 7.68 & 0.6304 & 16.47 & 9.89 & 0.6613 \\
        & AirPhyNet (\citeyear{airphynet}) & 16.89 & 11.76 & 0.5675 & 19.04 & 11.95 & 0.6609 & 12.62 & 8.67 & 0.5165 & 15.80 & 9.76 & 0.6891 \\
        & MGSFformer (\citeyear{mgsfformer}) & 17.00 & 10.43 & 0.5075 & 24.08 & 15.40 & 0.4580 & 14.55 & 10.31 & 0.3575 & 20.91 & 13.15 & 0.4551 \\
        & Air-DualODE (\citeyear{airdualode}) & 13.22 & 8.04 & 0.7026 & 18.99 & 12.02 & 0.6629 & 11.03 & 7.91 & 0.6306 & 16.08 & 10.06 & 0.6780 \\
        & AirFlow (Ours) & \textbf{12.42} & \textbf{7.28} & \textbf{0.7455} & \textbf{16.88} & \textbf{10.56} & \textbf{0.7224} & \textbf{10.85} & \textbf{7.43} & \textbf{0.6453} & \textbf{14.84} & \textbf{9.02} & \textbf{0.7105} \\
        \bottomrule
    \end{tabular}
    }
\end{table*}

\subsection{Ablation Study}
\begin{table}[htbp]
    \centering
    \caption{Ablation study on model components. \cmark{} and \xmark{} indicate whether a module is included or removed. AN: \Norm{}; CE: Convolution-Enhanced Branch in \LSAbb{}; CP: Context-Preserving Branch in \LSAbb{}. \RSAbb{}: \RStream{}; GBCA: Cross Attention. Best results are in \textbf{bold}.}
    \label{tab:ablation}

    \renewcommand{\arraystretch}{1.1}
    \setlength{\tabcolsep}{2pt}

    \resizebox{\columnwidth}{!}{
    \begin{tabular}{ccccccccc}
        \toprule
        Variant & AN & CE & CP & \RSAbb{} & GBCA & RMSE $\downarrow$ & MAE $\downarrow$ & R$^2$ $\uparrow$ \\
        \midrule
        V0 & \xmark & \cmark & \cmark & \cmark & \cmark & 16.68 & 9.12 & 0.8097 \\
        V1 & \cmark & \cmark & \xmark & \xmark & \xmark & 16.64 & 8.62 & 0.8068 \\
        V2 & \cmark & \xmark & \cmark & \xmark & \xmark & 16.81 & 8.70 & 0.8026 \\
        V3 & \cmark & \cmark & \cmark & \xmark & \xmark & 15.65 & 8.43 & 0.8276 \\
        V4 & \cmark & \xmark & \xmark & \cmark & \xmark & 15.85 & 8.62 & 0.8201 \\
        V5 & \cmark & \cmark & \xmark & \cmark & \cmark & 15.46 & 8.01 & 0.8323 \\
        V6 & \cmark & \xmark & \cmark & \cmark & \cmark & 15.33 & 7.94 & 0.8359 \\
        V7 & \cmark & \cmark & \cmark & \cmark & \xmark & 15.45 & 8.21 & 0.8334 \\
        AirFlow & \cmark & \cmark & \cmark & \cmark & \cmark & \textbf{14.42} & \textbf{7.47} & \textbf{0.8513} \\
        \bottomrule
    \end{tabular}
    }
\end{table}
To evaluate the contribution of components in AirFlow, we conduct ablation studies on the Beijing PM$_{2.5}$ task. 

As shown in Table~\ref{tab:ablation}, AirFlow achieves the best performance across all evaluation metrics, demonstrating that each component contributes positively to the final performance. $V_0$ removes AN, i.e. uses fixed MinMax, leads to 22.09\% increase in MAE and 4.89\% decrease in R$^2$, showing that input representation is sensitive to which station-level statistics are retained before temporal encoding. Removing \RSAbb{} ($V_3$) increases MAE by 12.85\%, while removing \LSAbb{} ($V_4$) leads to a larger increase of 15.39\%. These results indicate that modeling either long-term temporal correlations or abrupt change patterns is insufficient to match the performance of the full model. Within \LSAbb{}, removing the context-preserving branch ($V_5$) increases MAE by 7.23\%, while removing the convolution-enhanced branch ($V_6$) leads to 6.29\% increase.

Replacing GBCA with a simpler gate control ($V_7$) increases MAE by 9.91\% and reduces R$^2$ by 2.1\% relative to the AirFlow, indicating that fine-grained feature interactions are more effective than coarse weighting for multi-scale feature fusion.

\subsection{Computational Efficiency}
\label{sec:efficiency}

\begin{figure}[t]
    \centering

    \includegraphics[width=\columnwidth]{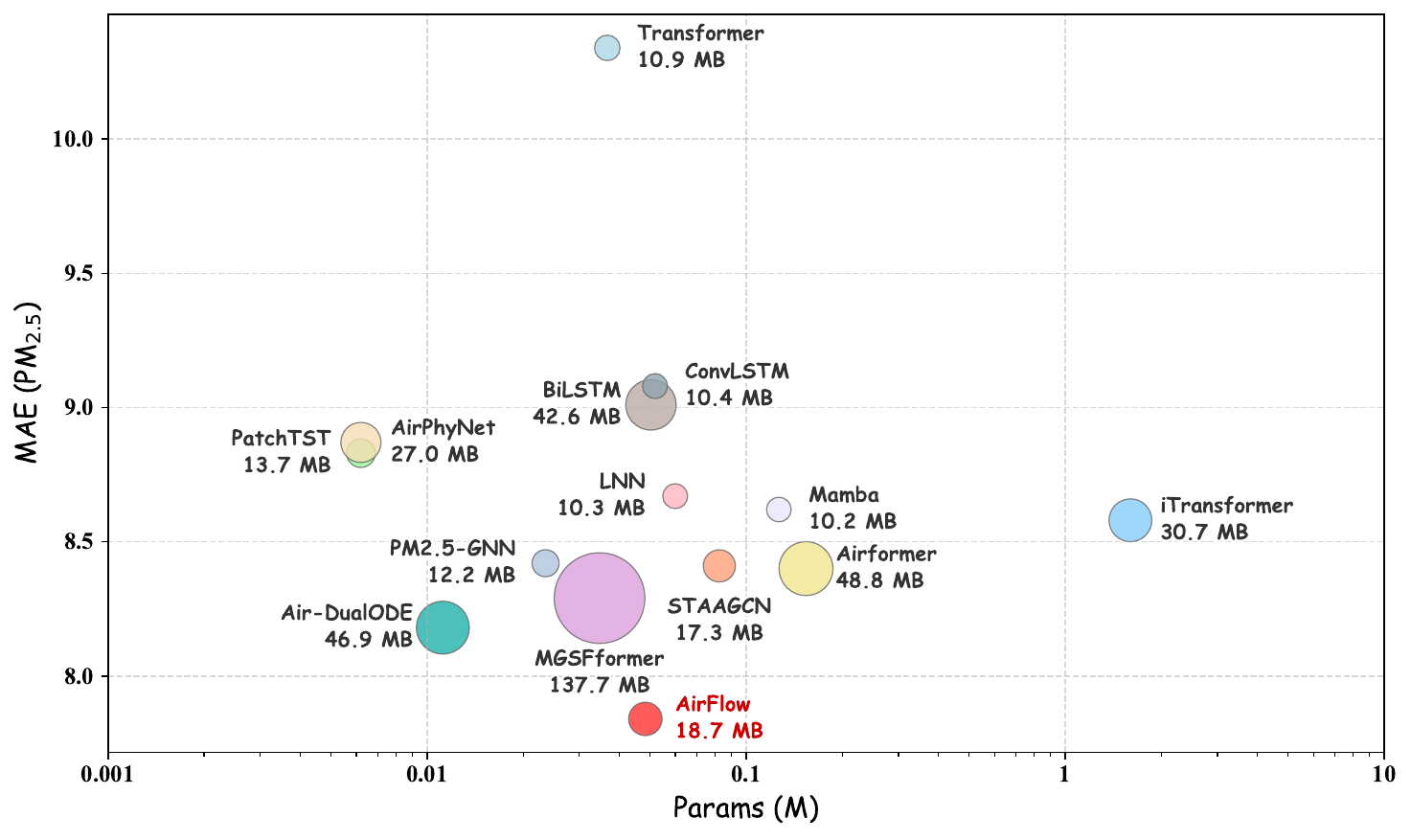}

    \vspace{0em}

    \includegraphics[width=\columnwidth]{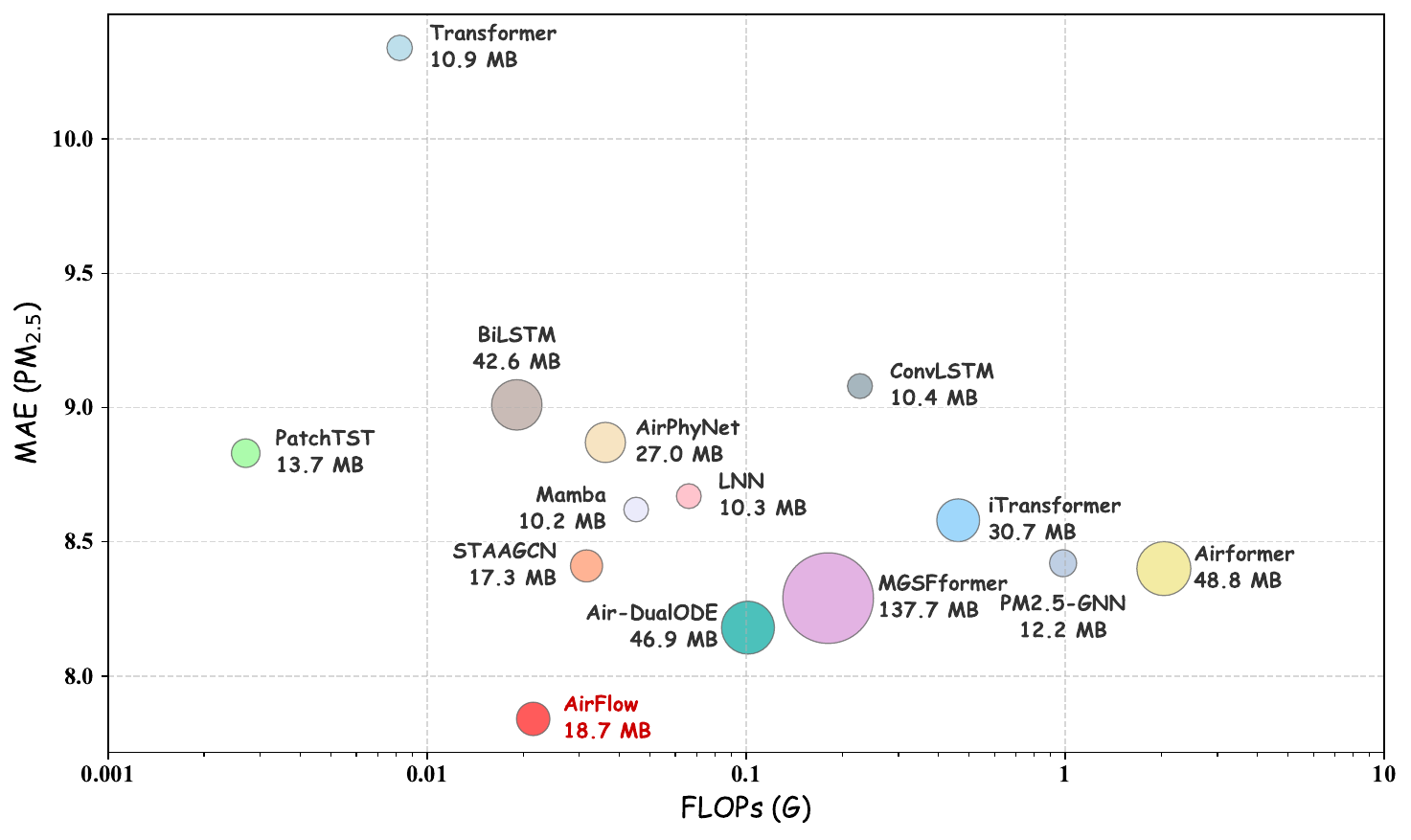}

    \caption{Computational efficiency comparison based on parameters
    (upper panel) and FLOPs (lower panel). Bubble size reflects peak
    memory usage. MAE is reported on the Beijing PM$_{2.5}$
    forecasting task.}
    \label{fig:overall_efficiency}
\end{figure}

For real-world air quality early-warning systems, computational complexity and memory footprint are as critical as forecasting accuracy. As summarized in Figure~\ref{fig:overall_efficiency}, AirFlow achieves a superior balance between modeling complexity and computational feasibility. Specifically, its parameter count (0.0483M) is lightweight compared to sequence models like iTransformer (1.6010M), and its FLOPs are approximately 1/95 of the spatiotemporal baseline AirFormer (0.0215G vs. 2.0378G). Furthermore, AirFlow maintains a remarkably low peak memory footprint (18.7 MB), avoiding the exorbitant memory costs of spatial models like MGSFformer (137.7 MB). 

\section{Limitations and Future Work}
Although AirFlow effectively captures the multi-scale temporal evolution of air quality data through variable-adaptive normalization and hierarchical dual-stream architecture, this study is currently constrained by the limited external covariates available in the datasets. Therefore, the potential intervention mechanisms of cross-modal factors on the underlying dynamics have not yet been fully explored. In the future, we expect to extend our research to urban sensing scenarios supported by multi-source heterogeneous big data. We aim to investigate how to integrate physical priors with multi-modal features, thereby further enhancing the model's reasoning capabilities within complex urban atmospheric systems.

\section{Conclusion}
In this paper, we propose AirFlow, a lightweight pollutant-aware forecasting framework under incomplete transport context. AirFlow utilizes statistic-guided normalization routing to preserves cross-window concentration references for statistically stable channels while reducing local level and scale shifts for drifting channels. The multi-rate latent state modeling identify which historical dependencies should continue to propagate and how quickly different features update within the recurrent state. Gated cross-attention enables fine-grained interaction between the two streams, allowing rapid concentration changes to be interpreted against the relevant historical context before producing an integrated forecasting representation. Extensive experiments on three real-world datasets across multiple cities and pollutants demonstrate that AirFlow can provide accurate and efficient \aqf{}.


\section*{Acknowledgements}
This work was supported by Deep Earth Probe and Mineral Resources Exploration - National Science and Technology Major Project (2025ZD1007008).

\clearpage
\bibliographystyle{plainnat}
\bibliography{cas-refs}

\appendix
\section{Appendix}
\subsection{Experimental Protocol and Reproducibility}
\label{sec:additional experimental settings}
\begin{figure*}[htbp]
    \centering
    \begin{subfigure}[b]{0.32\textwidth}
        \centering
        \includegraphics[width=\linewidth, height=4cm]{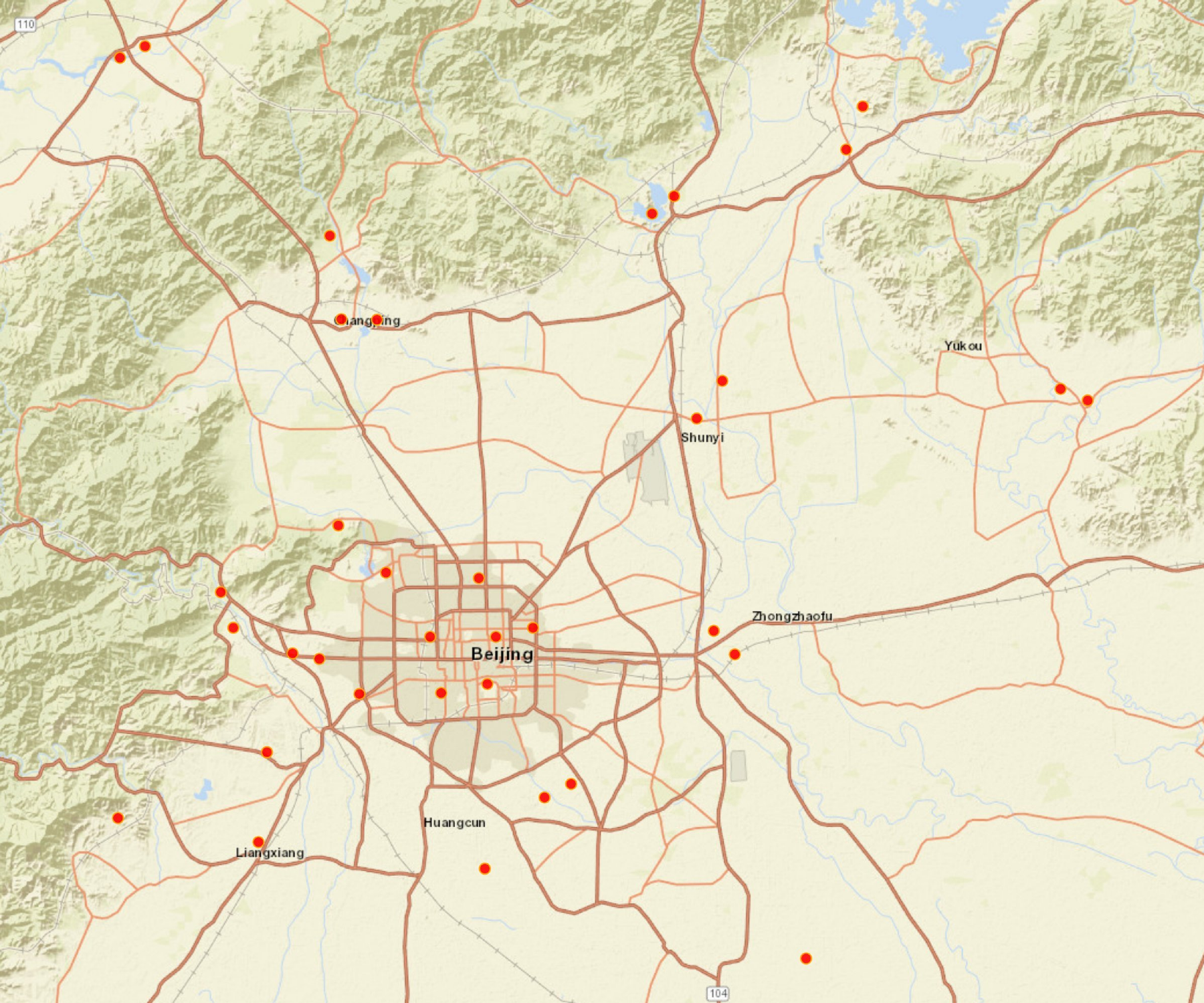}
        \caption{Beijing (Map)}
        \label{fig:beijing-map}
    \end{subfigure}
    \hfill
    \begin{subfigure}[b]{0.32\textwidth}
        \centering
        \includegraphics[width=\linewidth, height=4cm]{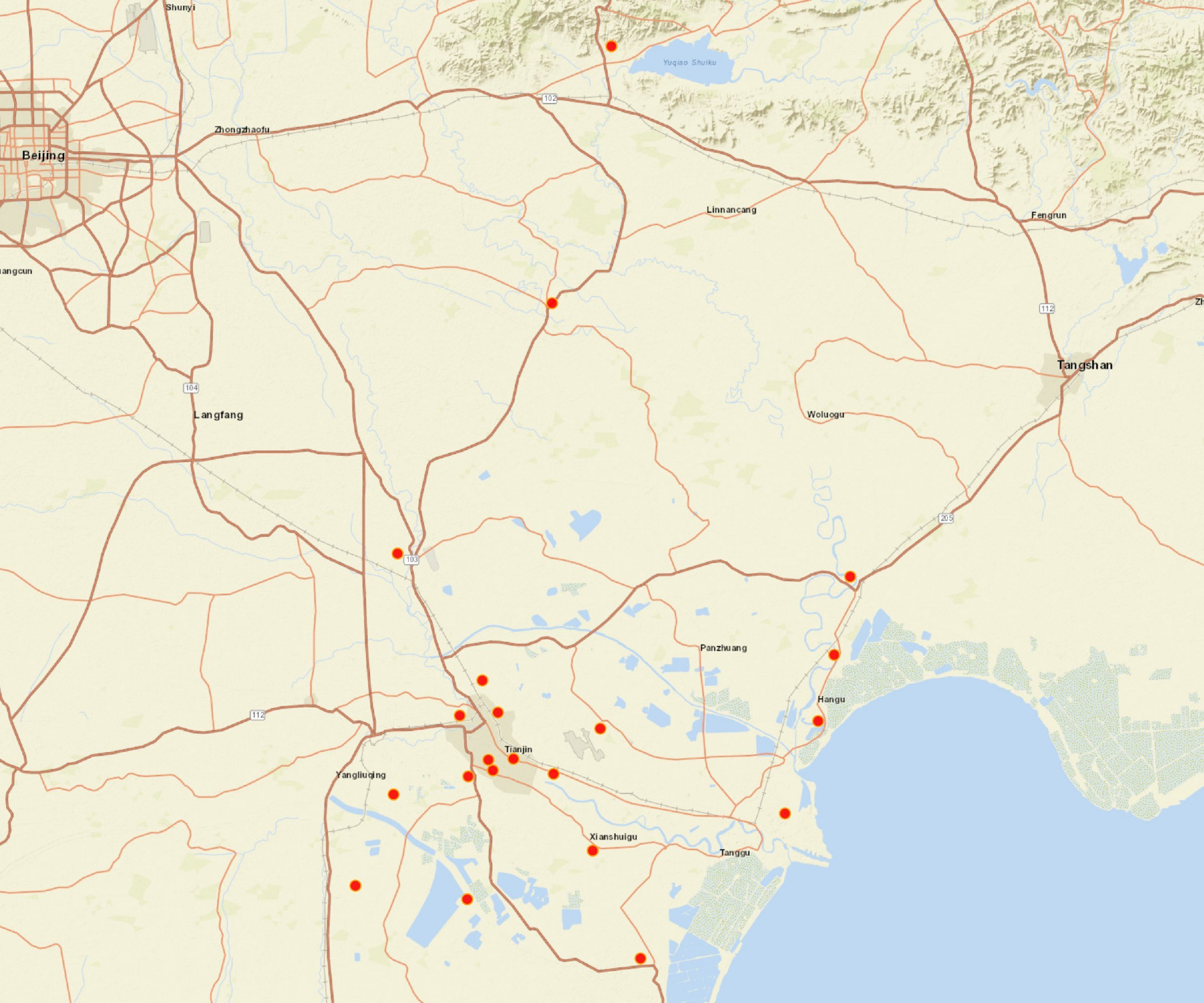}
        \caption{Tianjin (Map)}
        \label{fig:beijing-sat}
    \end{subfigure}
    \hfill
    \begin{subfigure}[b]{0.32\textwidth}
        \centering
        \includegraphics[width=\linewidth, height=4cm]{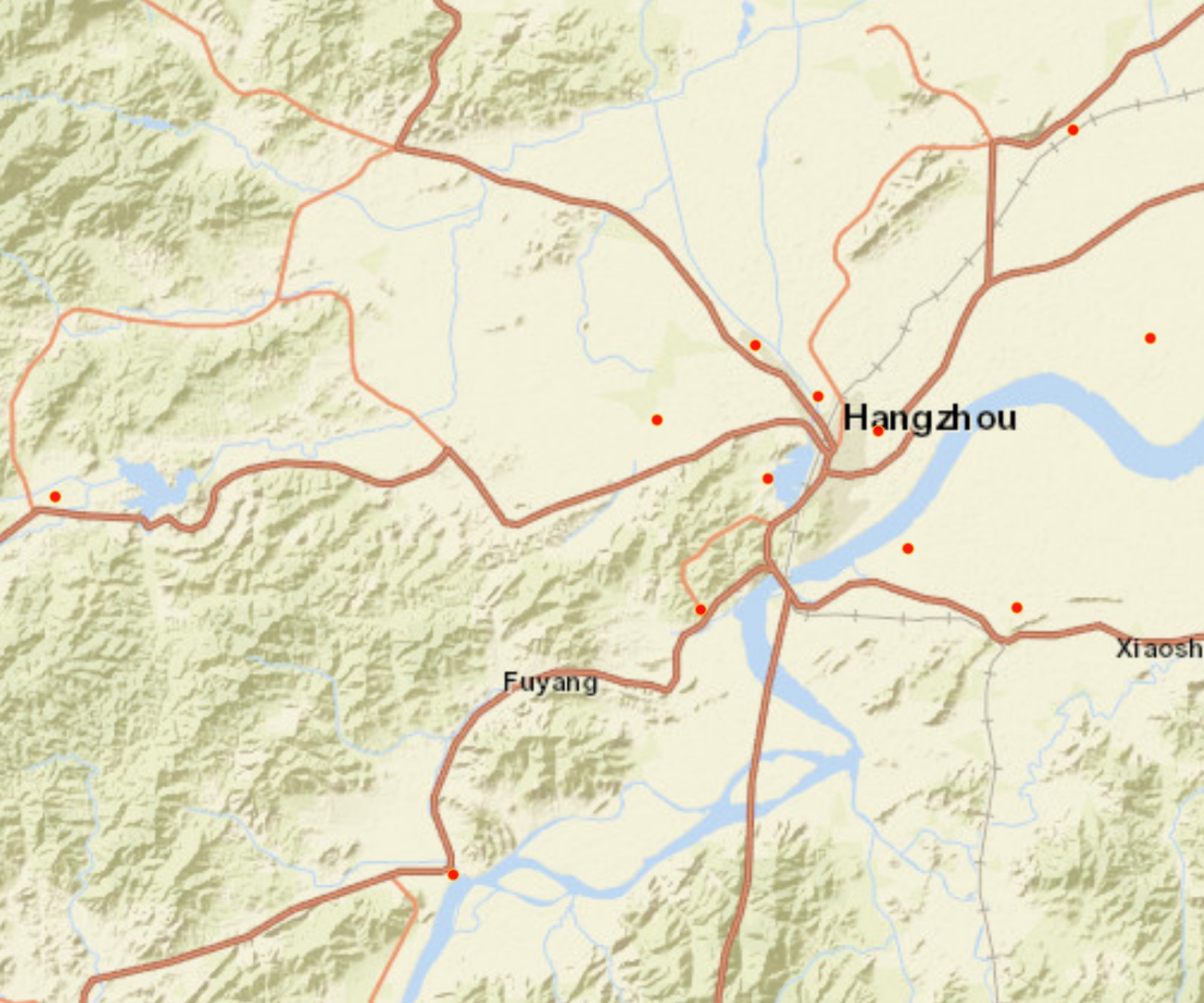} 
        \caption{Hangzhou (Map)}
        \label{fig:beijing-terrain}
    \end{subfigure}

    \vspace{0.3cm} 

    \begin{subfigure}[b]{0.32\textwidth}
        \centering
        \includegraphics[width=\linewidth, height=4cm]{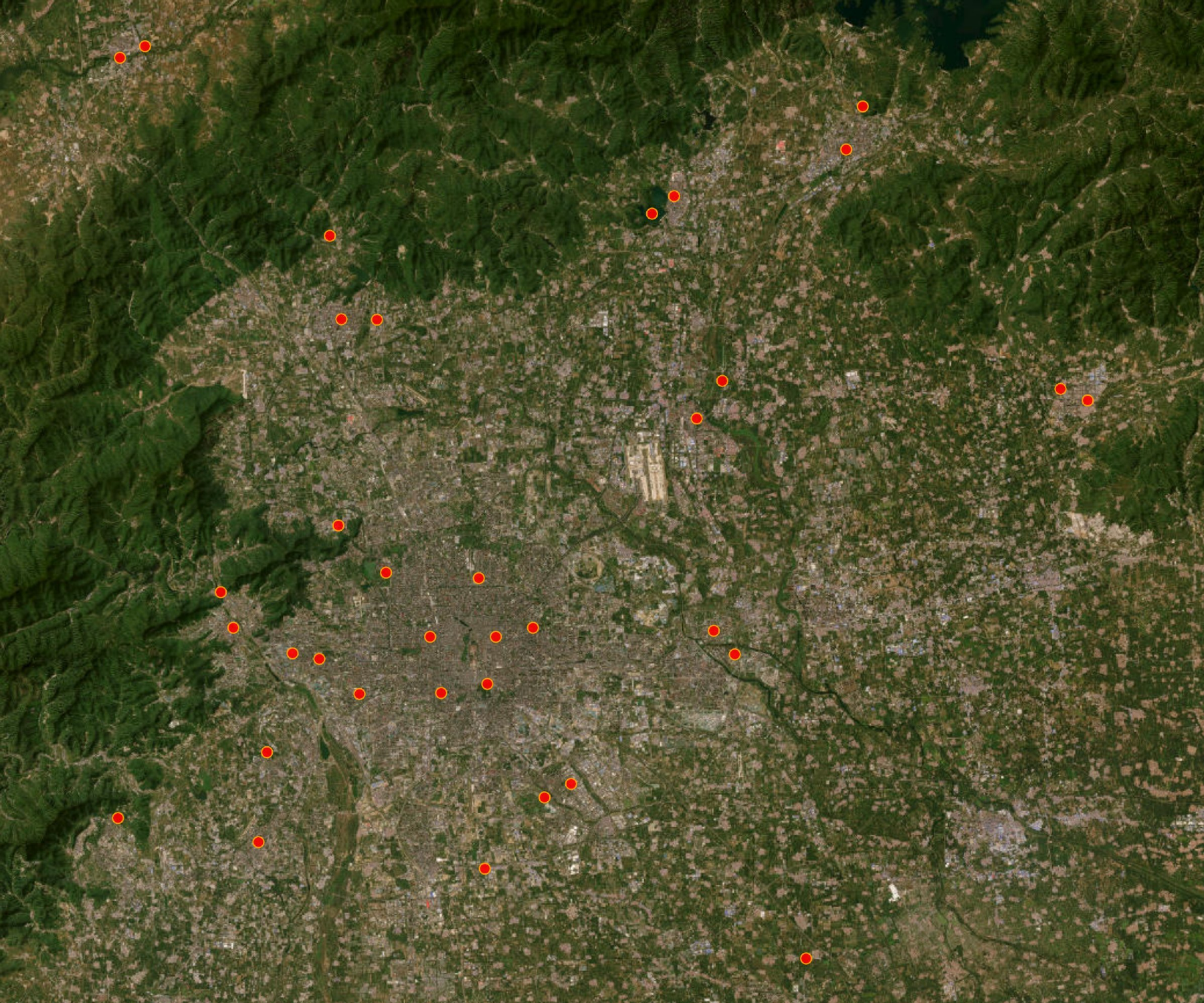}
        \caption{Beijing (Satellite)}
        \label{fig:tianjin-map}
    \end{subfigure}
    \hfill
    \begin{subfigure}[b]{0.32\textwidth}
        \centering
        \includegraphics[width=\linewidth, height=4cm]{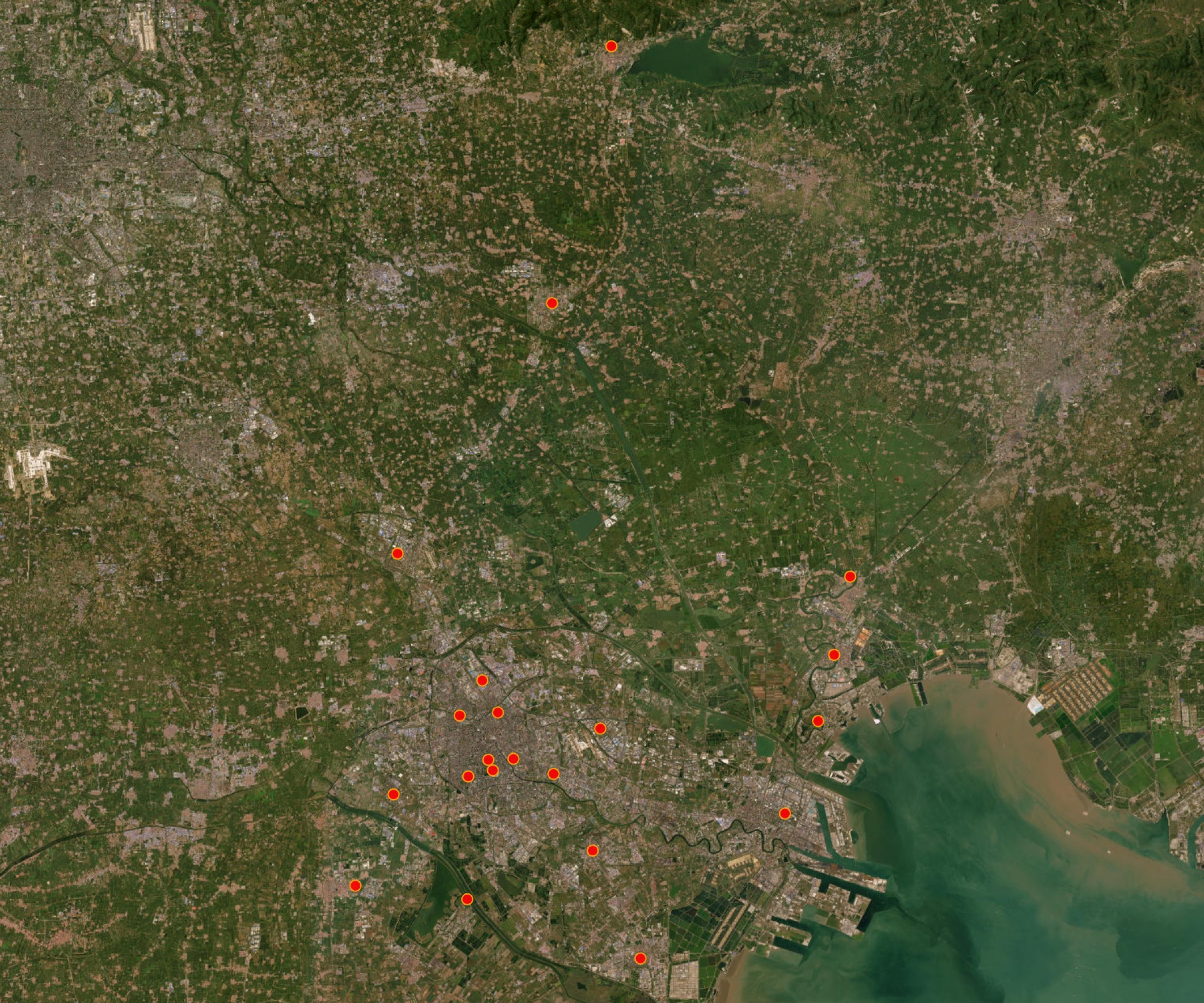}
        \caption{Tianjin (Satellite)}
        \label{fig:tianjin-sat}
    \end{subfigure}
    \hfill
    \begin{subfigure}[b]{0.32\textwidth}
        \centering
        \includegraphics[width=\linewidth, height=4cm]{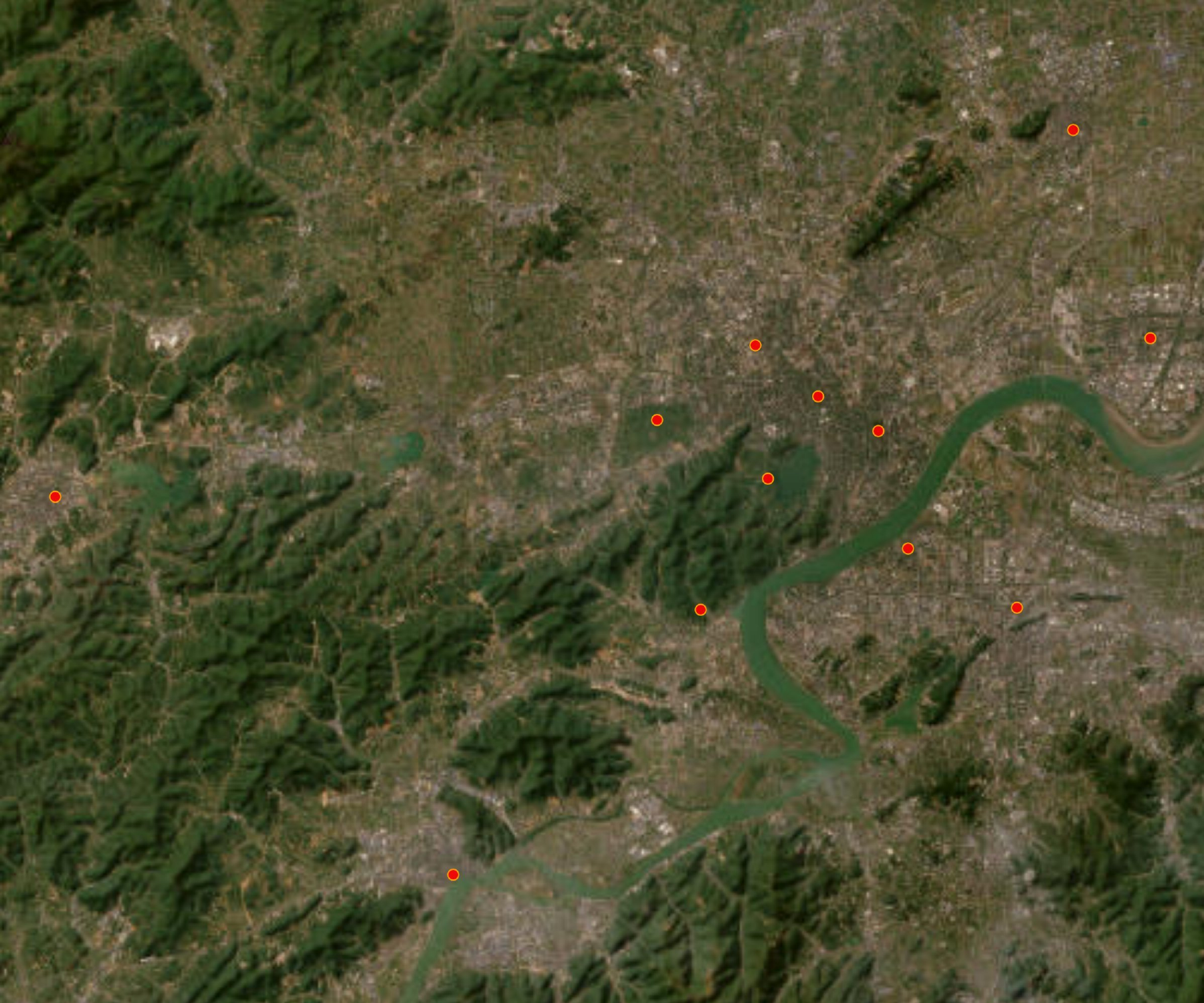} %
        \caption{Hangzhou (Satellite)}
        \label{fig:tianjin-terrain}
    \end{subfigure}

    \caption{Geographical distribution of air quality monitoring stations in Beijing, Tianjin, and Hangzhou. The first row (a)-(c) displays the map views, while the second row (d)-(f) presents the corresponding satellite imagery. Red dots represent the specific deployment locations of the stations.}
    \label{fig:station}
\end{figure*}

\noindent\textbf{Datasets Statistics and Station Coverage}
Statistical summaries of the Beijing, Tianjin and Hangzhou datasets are reported in Table~\ref{tab:station_info}, while the spatial distribution of monitoring stations is shown in Figure~\ref{fig:station}.

\begin{table}[htbp]
    \centering
    \caption{Statistical information of air quality datasets for Beijing, Tianjin, and Hangzhou.}
    \label{tab:station_info}
    \begin{tabular}{llccc}
        \toprule
        Dataset & Pollutant & Unit & Mean & Std. \\
        \midrule
        \multirow{7}{*}{Beijing} 
        & PM$_{2.5}$ & $\mu g/m^3$ & 32.69 & 34.38 \\
        & PM$_{10}$ & $\mu g/m^3$ & 66.00 & 85.71 \\
        & SO$_2$ & $\mu g/m^3$ & 2.82 & 2.11 \\
        & NO$_2$ & $\mu g/m^3$ & 24.57 & 20.62 \\
        & CO & $mg/m^3$ & 0.5 & 0.3 \\
        & O$_3$ & $\mu g/m^3$ & 64.23 & 49.52 \\
        & AQI & - & 61.23 & 51.23 \\
        \midrule
        \multirow{7}{*}{Tianjin} 
        & PM$_{2.5}$ & $\mu g/m^3$ & 40.91 & 40.45 \\
        & PM$_{10}$ & $\mu g/m^3$ & 78.86 & 80.78 \\
        & SO$_2$ & $\mu g/m^3$ & 7.36 & 4.60 \\
        & NO$_2$ & $\mu g/m^3$ & 33.66 & 24.12 \\
        & CO & $mg/m^3$ & 0.69 & 0.38 \\
        & O$_3$ & $\mu g/m^3$ & 72.04 & 53.66 \\
        & AQI & - & 72.31 & 54.51 \\
        \midrule
        \multirow{7}{*}{Hangzhou} 
        & PM$_{2.5}$ & $\mu g/m^3$ & 30.76 & 25.89 \\
        & PM$_{10}$ & $\mu g/m^3$ & 51.12 & 39.93 \\
        & SO$_2$ & $\mu g/m^3$ & 6.43 & 1.92 \\
        & NO$_2$ & $\mu g/m^3$ & 28.93 & 19.35 \\
        & CO & $mg/m^3$ & 0.66 & 0.19 \\
        & O$_3$ & $\mu g/m^3$ & 62.99 & 46.26 \\
        & AQI & - & 51.91 & 33.05 \\
        \bottomrule
    \end{tabular}
\end{table}

\noindent\textbf{Implementation Details and Hyperparameters}
All experiments were conducted on two NVIDIA RTX A5000 GPUs (24~GB VRAM) hosted on workstations equipped with Intel Core i7-14700KF CPUs. 
The software environment comprised Python~3.8.20, PyTorch~2.2.2, and CUDA~12.1. All baselines are evaluated using the same pollutant variables, historical input windows, target sequences, station set, and chronological train/validation/test splits. For baselines that model spatial relationships, we additionally provide the latitude and longitude of each monitoring station. 
The hyperparameter configurations are summarized in Table~\ref{tab:hyperparameters}.

\begin{table*}[bp]
    \centering
    \caption{Key hyperparameter settings of AirFlow.}
    \label{tab:hyperparameters}
    \begin{tabular}{ccc}
        \toprule
        Config & Description & Values \\
        \midrule
        Optimizer & The algorithm used for parameter optimization. & AdamW \\
        Learning rate scheduler & The strategy for adjusting the learning rate during training. & SGDR \\
        Learning rate & The starting value for the learning rate. & 0.0002 \\
        Weight decay & The L2 regularization coefficient to prevent overfitting. & 0.0001 \\
        Multi-head num & The number of parallel attention mechanisms in the multi-head attention layer. & 4 \\
        Number of layers & The number of stacked HDS layers in the model. & 3 \\
        Dropout & The probability of randomly dropping out units during training. & 0.1 \\
        Batch size & The number of training samples processed per iteration. & 32 \\
        Epoch & The total number of complete passes through the training dataset. & 60 \\
        Periodicity threshold & $\theta_{\rho}$ in the normalization routing rule. & 0.425 \\
        Drift-rate threshold & $\theta_\nu$ in the normalization routing rule. & 0.4 \\
        Diurnal lag & Lag $h$ used to compute the periodicity score. & 24 \\
        Drift window & Moving-average window $w$ used to compute the drift-rate score. &
        168 \\
        Hidden dimension & Latent representation width $d$ used throughout the HDS blocks. & 32 \\
        Expanded dimension & Expanded width $d_e$ used in the \LSAbb{}. & 64 \\
        State dimension & State dimension $d_s$ of the selective propagation operator. & 16 \\
        DWConv kernel size & Kernel size of the depth-wise temporal convolution in the \LSAbb{}. & 4 \\
        Discretization rank & Rank of the projection used to parameterize the discretization vector in the \LSAbb{}. & 4 \\
        Prediction-head width & Hidden width of the two-layer prediction head. & 16 \\
        \RSAbb{} hidden dimension & Hidden-state width $d_r$ of the \RSAbb{}. & 16 \\
        Step-scale initialization & Initial value of the learnable scalar $\eta$ controlling the response coefficients. & 0.1 \\
        Concentration loss weights & Weights $\lambda_s$ and  $\lambda_1$ of the SmoothL1 and absolute error terms. & 0.8, 0.15 \\
        Variation loss weight & Weight $\lambda_\Delta$ of the inter-step variation loss. & 0.05 \\
        Robust transition point & Transition point $\kappa$ of the SmoothL1 penalty & 1.0\\
        \bottomrule
    \end{tabular}
\end{table*}

\subsection{Mechanism and Design Analysis}
\noindent\textbf{Normalization Routing Analysis}
\label{sec:normalization_analysis}
We compare fixed normalization with statistic-guided routing to determine whether the gain arises from the normalization operator itself or from pollutant-dependent assignments. The statistics used for routing are computed only from the training split. The thresholds are selected on the validation split and fixed to $(\theta_{\rho},\theta_{\nu})=(0.425,0.40)$ for test evaluation.

As shown in Table~\ref{tab:normalization_strategy}, RevIN
provides a strong baseline, while the selected routing configuration yields a further improvement. This result suggests that removing window level and
scale shifts is beneficial for most non-stationary sequences, but applying the same transformation to every
pollutants may discard useful concentration
references. The routing rule retains these references only
when the training statistics indicate sufficiently stable
periodicity or limited drift.

\begin{figure}[htbp]
    \centering
    \includegraphics[width=\columnwidth]{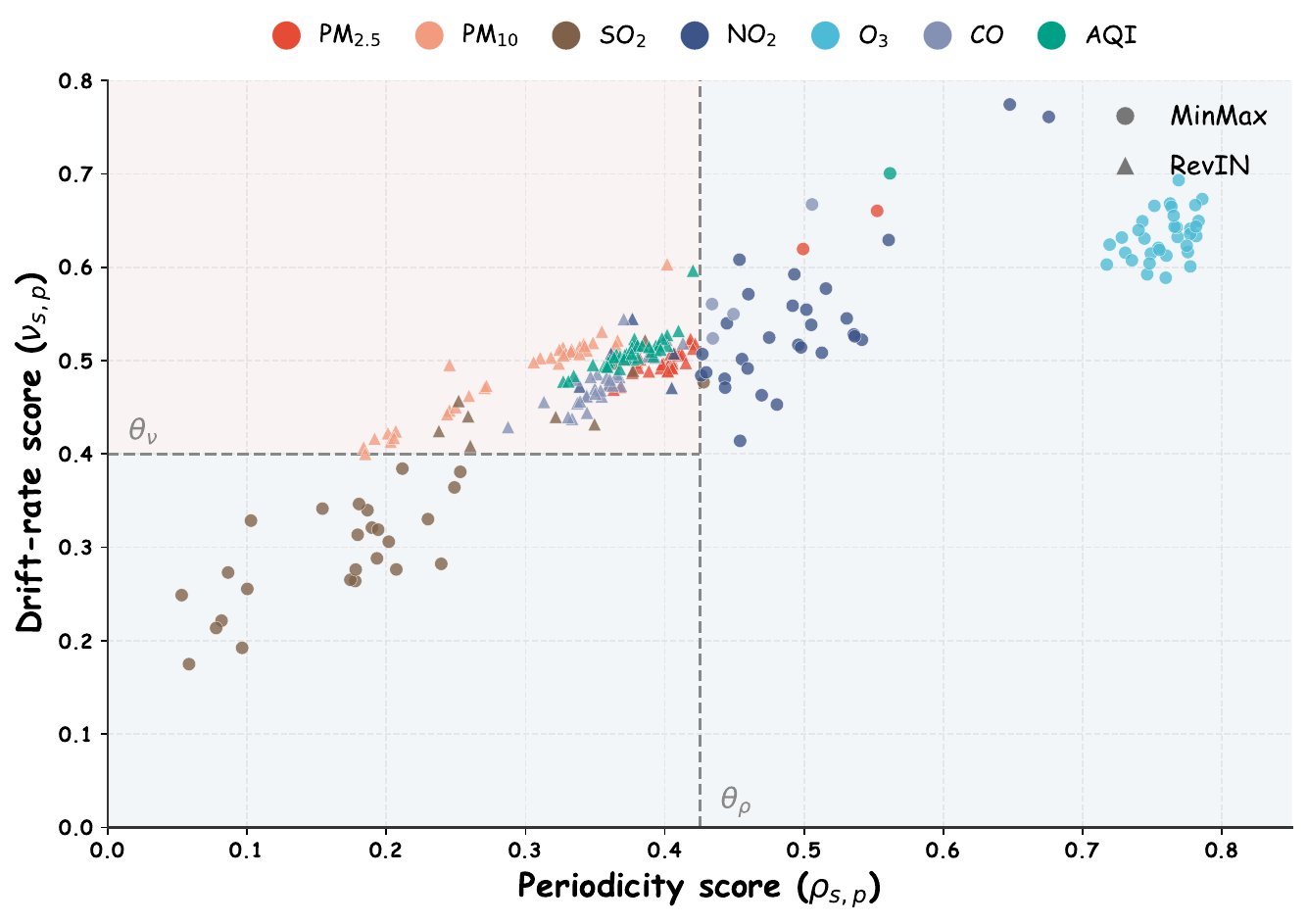}
    \caption{Pollutant normalization assignment. Each point represents one station–pollutant pair. Colors distinguish variables, while marker
shapes indicate the assigned normalization operator. Pairs
satisfying $\rho_{s,p}<\theta_{\rho}$ and
$\nu_{s,p}\geq\theta_{\nu}$ are routed to RevIN, the
remaining routed to MinMax.}
    \label{fig:norm_assignment}
\end{figure}

The threshold pairs do not produce a monotonic performance
trend. A small threshold change may leave most assignments unchanged or switch several pollutants located near a decision boundary. Figure~\ref{fig:norm_assignment} further illustrates that station--pollutant pairs occupy different regions of the periodicity or drift space, including multiple observations close to the selected boundaries.

\begin{table}[h] 
\centering \caption{Comparison of fixed normalization strategies and
    statistic-guided routing under different candidate threshold.} \label{tab:normalization_strategy} \setlength{\tabcolsep}{3pt} \resizebox{\columnwidth}{!}{ \begin{tabular}{cccccc} \toprule Strategy & $\theta_{\rho}$ & $\theta_{\nu}$ & RMSE $\downarrow$ & MAE $\downarrow$ & R$^2 \uparrow$ \\ \midrule All MinMax & -- & -- & 16.68 & 9.12 & 0.8097 \\ All RevIN & -- & -- & 14.74 & 7.61 & 0.8457 \\ \midrule \multirow{6}{*}{\shortstack{Statistic-guided\\Routing}} & 0.425 & 0.50 & 16.12 & 8.98 & 0.8143 \\ & 0.425 & 0.45 & 14.79 & 7.75 & 0.8412 \\ & \textbf{0.425} & \textbf{0.40}
            & \textbf{14.42} & \textbf{7.47} & \textbf{0.8513} \\ & 0.400 & 0.50 & 16.54 & 9.04 & 0.8111 \\ & 0.400 & 0.45 & 15.48 & 8.33 & 0.8301 \\ & 0.400 & 0.40 & 15.76 & 8.49 & 0.8217 \\ \bottomrule \end{tabular} }
\end{table}

\noindent\textbf{Spatial Augmentation Analysis}
\label{sec:spatial_ablation}
To investigate whether introducing spatial inductive biases might lead to a mismatch in the absence of reliable transport priors, we explored several spatial modeling strategies. (1) Spatial Loss: Adding a penalty term to encourage forecasting consistency between proximal stations;
(2) Coordinate Embedding: Incorporating station latitude and longitude as additional input features;
(3) GCN Fusion: Introducing Graph Convolutional Network (GCN) after the dual-stream fusion to aggregate neighborhood information.

The results in Table~\ref{tab:spatial_results} show that the examined spatial augmentations do not yield useful gains under the current coverage. This suggests that, under limited station observations and incomplete transport-related priors, increasing spatial modeling complexity may introduce additional dependencies without addressing abrupt temporal changes in pollutant trajectories. These results motivate us to reconsider whether reliable forecasting can be achieved by modeling station-level pollutant evolution without relying on additional spatial graph propagation.
\begin{table*}[htbp]
    \centering
    \caption{Performance comparison of different spatial dependency modeling against AirFlow.}
    \label{tab:spatial_results}
    \begin{tabular}{lccccccccc}
        \toprule
        \multirow{2}{*}{Model Variant} & \multicolumn{3}{c}{Beijing} & \multicolumn{3}{c}{Tianjin} & \multicolumn{3}{c}{Hangzhou} \\
        \cmidrule(lr){2-4} \cmidrule(lr){5-7} \cmidrule(lr){8-10}
        & RMSE $\downarrow$ & MAE $\downarrow$ & R$^2\uparrow$ & RMSE $\downarrow$ & MAE $\downarrow$ & R$^2\uparrow$ & RMSE $\downarrow$ & MAE $\downarrow$ & R$^2\uparrow$ \\
        \midrule
        Spatial Loss & 25.21 & 13.97 & 0.5564 & 31.27 & 19.04 & 0.2316 & 19.84 & 12.82 & 0.3285 \\
        Coord. Embed & 15.29 & 7.95 & 0.8369 & 19.57 & 11.33 & 0.6991 & 12.70 & 7.50 & 0.7247 \\
        GCN Fusion & 15.19 & 8.04 & 0.8390 & 19.56 & 11.47 & 0.6992 & 12.89 & 7.64 & 0.7162 \\
        AirFlow & \textbf{14.42} & \textbf{7.47} & \textbf{0.8513} & \textbf{18.87} & \textbf{10.97} & \textbf{0.7124} & \textbf{12.42} & \textbf{7.28} & \textbf{0.7455} \\
        \bottomrule
    \end{tabular}
\end{table*}

\noindent\textbf{Forecasting Horizon Analysis}
\label{sec:pre_horizons}
\begin{table*}[htbp]
    \centering
    \caption{Performance comparison under different input lengths and forecasting horizons. Best results are in \textbf{bold}.}
    \label{tab:horizons} 
    \setlength{\tabcolsep}{3pt}
    \resizebox{\textwidth}{!}{
    \begin{tabular}{lcccccccccccc}
        \toprule
        \multirow{2}{*}{Model} & \multicolumn{3}{c}{12h$\to$6h} & \multicolumn{3}{c}{24h$\to$6h} & \multicolumn{3}{c}{12h$\to$12h} & \multicolumn{3}{c}{24h$\to$12h} \\
        \cmidrule(lr){2-4} \cmidrule(lr){5-7} \cmidrule(lr){8-10} \cmidrule(lr){11-13}
        & RMSE $\downarrow$ & MAE $\downarrow$ & R$^2\uparrow$ & RMSE $\downarrow$ & MAE $\downarrow$ & R$^2\uparrow$ & RMSE $\downarrow$ & MAE $\downarrow$ & R$^2\uparrow$ & RMSE $\downarrow$ & MAE $\downarrow$ & R$^2\uparrow$ \\
        \midrule
        BiLSTM & 16.44 & 9.01 & 0.8113 & 18.89 & 9.58 & 0.7522 & 23.52 & 13.03 & 0.6148 & 24.51 & 13.63 & 0.5842 \\
        ConvLSTM & 16.48 & 9.08 & 0.8104 & 18.34 & 9.59 & 0.7665 & 24.28 & 13.6 & 0.5896 & 23.56 & 13.68 & 0.6158 \\
        Transformer & 17.33 & 10.34 & 0.7903 & 18.47 & 9.76 & 0.7630 & 23.96 & 13.15 & 0.6002 & 25.32 & 14.05 & 0.5564 \\
        PM2.5-GNN & 15.84 & 8.42 & 0.8262 & 16.88 & 8.97 & 0.8059 & 22.27 & 12.33 & 0.6511 & 23.01 & 12.88 & 0.6287 \\
        LNN & 17.02 & 8.67 & 0.7978 & 18.27 & 9.36 & 0.7683 & 22.71 & 12.59 & 0.6409 & 24.15 & 13.89 & 0.5964 \\
        PatchTST& 16.12 & 8.83 & 0.8186 & 17.43 & 9.08 & 0.7891 & 22.53 & 12.66 & 0.6467 & 23.31 & 13.06 & 0.6238 \\
        AirFormer & 15.68 & 8.40 & 0.8284 & 17.53 & 8.71 & 0.7867 & 22.10 & 12.21 & 0.6641 & 24.01 & 13.01 & 0.6001 \\
        STAAGCN & 15.98 & 8.41 & 0.8132 & 17.97 & 8.73 & 0.7701 & 22.23 & 12.25 & 0.6467 & 23.11 & 12.73 & 0.6242 \\
        iTransformer & 15.88 & 8.58 & 0.8239 & 16.47 & 8.91 & 0.8118 & 22.49 & 12.35 & 0.6530 & 23.33 & 12.82 & 0.6232 \\
        Mamba & 16.64 & 8.62 & 0.8068 & 19.18 & 9.66 & 0.7445 & 24.43 & 13.42 & 0.5845 & 24.49 & 13.43 & 0.5848 \\
        AirPhyNet & 15.93 & 8.87 & 0.8127 & 17.30 & 9.04 & 0.8145 & 22.63 & 13.28 & 0.6317 & 22.61 & 13.40 & 0.6343 \\
        MGSFformer & 15.62 & 8.29 & 0.8297 & 16.08 & 8.67 & 0.8192 & 22.21 & 12.33 & 0.6502 & 23.55 & 13.32 & 0.6175 \\
        Air-DualODE & 15.22 & 8.18 & 0.8348 & 16.32 & 8.68 & 0.8188 & 22.43 & 12.65 & 0.6422 & 22.51 & 12.45 & 0.6387 \\
        AirFlow (Ours) & \textbf{14.42} & \textbf{7.47} & \textbf{0.8513} & \textbf{15.73} & \textbf{8.24} & \textbf{0.8283} & \textbf{21.84} & \textbf{11.47} & \textbf{0.6679} & \textbf{22.47} & \textbf{11.89} & \textbf{0.6460} \\
        \bottomrule
    \end{tabular}
    }
\end{table*}

\begin{figure*}[htbp] 
    \centering
    
    \newcommand{\taylorwidth}{0.24\linewidth}
    
    
    \begin{minipage}{\taylorwidth} 
        \centering 
        \includegraphics[height=0.9cm]{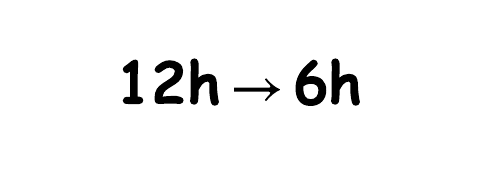} 
    \end{minipage}
    \hfill
    \begin{minipage}{\taylorwidth} 
        \centering 
        \includegraphics[height=0.9cm]{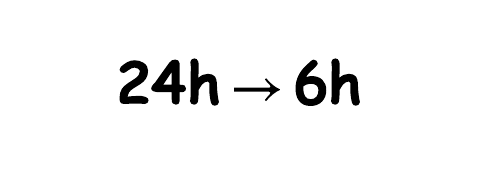} 
    \end{minipage}
    \hfill
    \begin{minipage}{\taylorwidth} 
        \centering 
        \includegraphics[height=0.9cm]{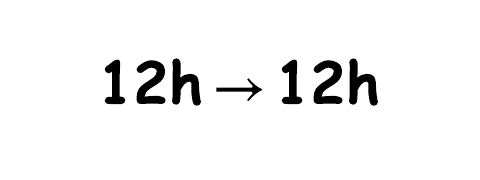} 
    \end{minipage}
    \hfill
    \begin{minipage}{\taylorwidth} 
        \centering 
        \includegraphics[height=0.9cm]{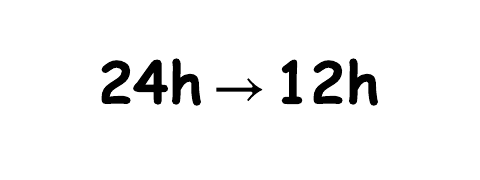} 
    \end{minipage}
    
    
    \begin{subfigure}{\taylorwidth}
        \centering
        \includegraphics[width=\linewidth]{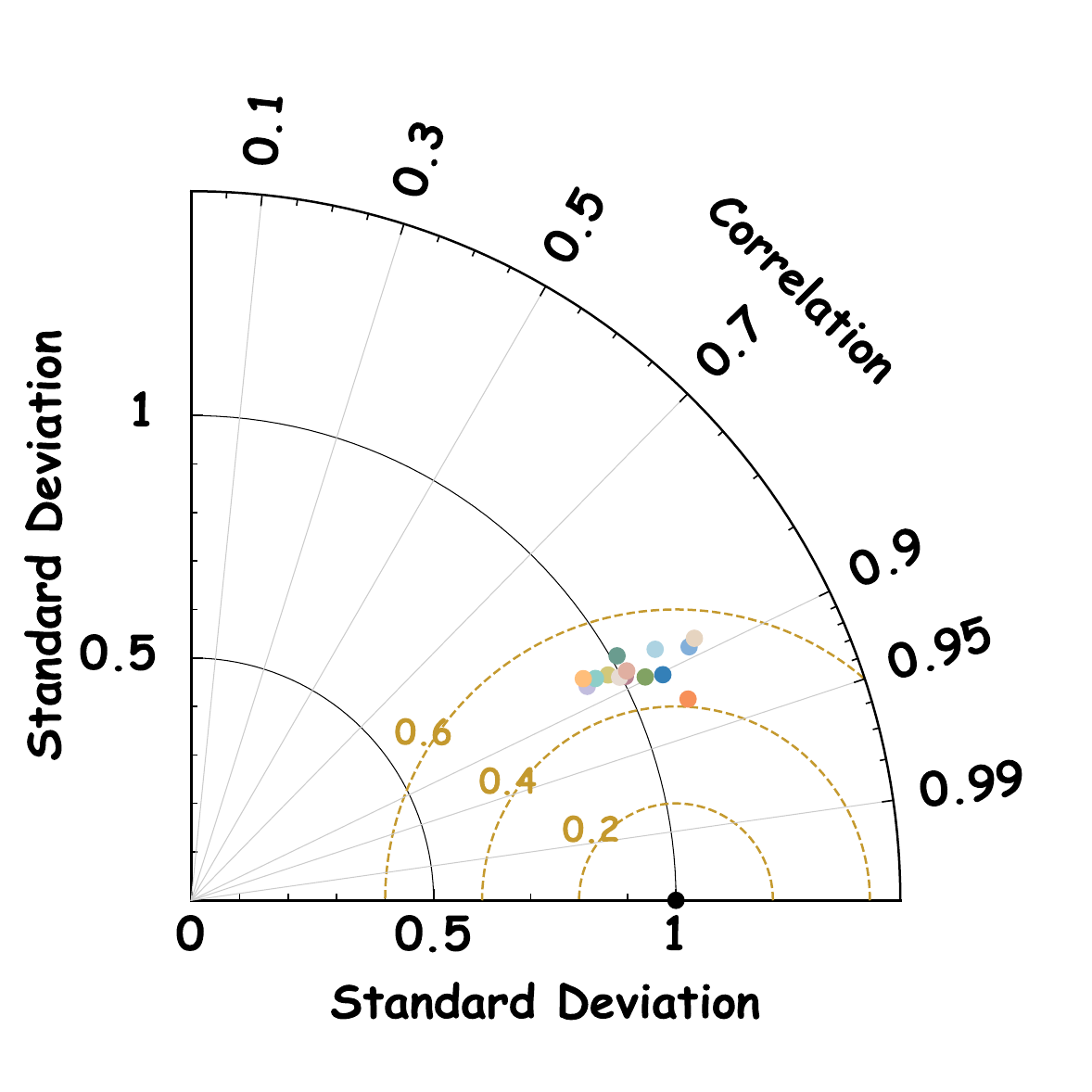} 
    \end{subfigure}
    \hfill
    \begin{subfigure}{\taylorwidth}
        \centering
        \includegraphics[width=\linewidth]{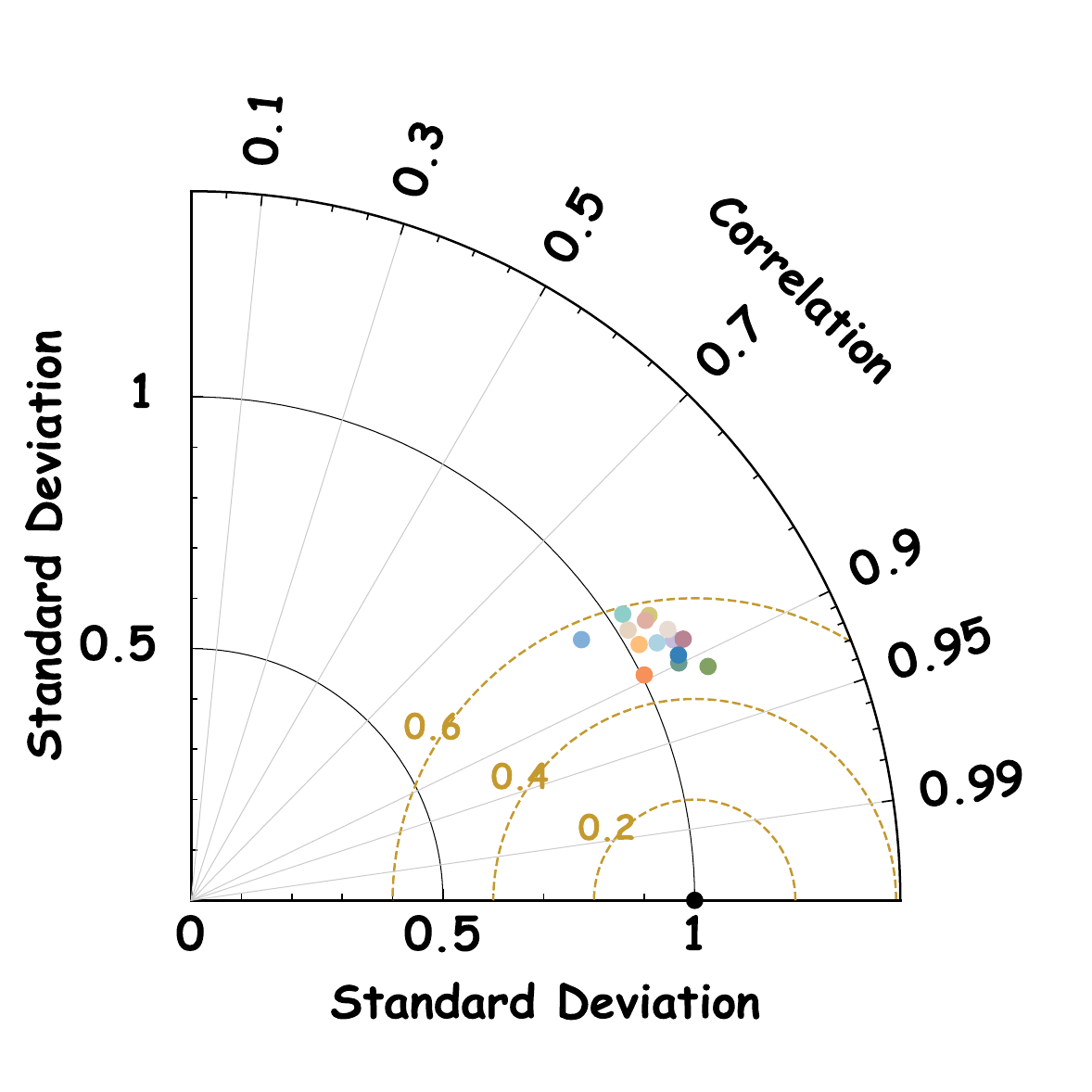}
    \end{subfigure}
    \hfill
    \begin{subfigure}{\taylorwidth}
        \centering
        \includegraphics[width=\linewidth]{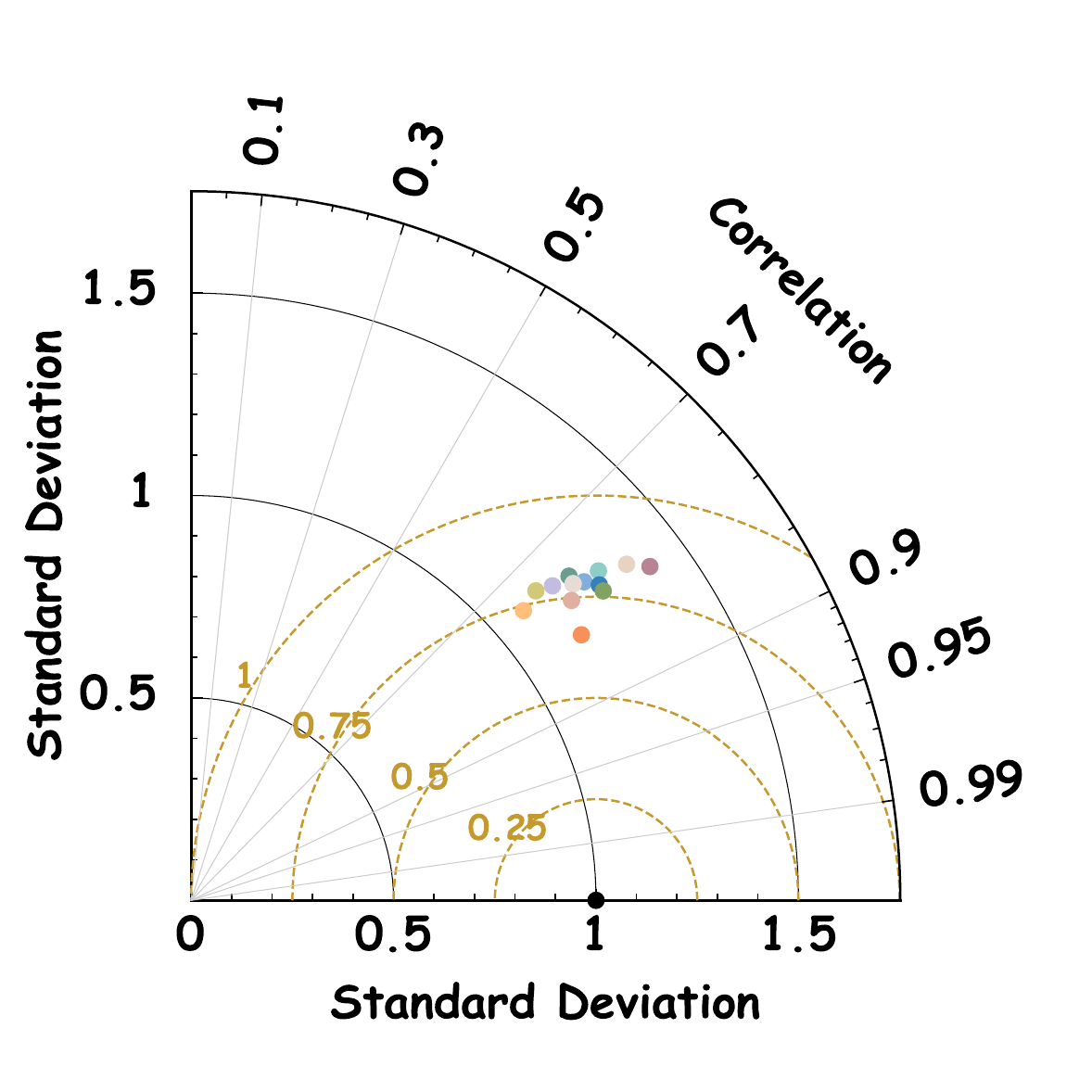}
    \end{subfigure}
    \hfill
    \begin{subfigure}{\taylorwidth}
        \centering
        \includegraphics[width=\linewidth]{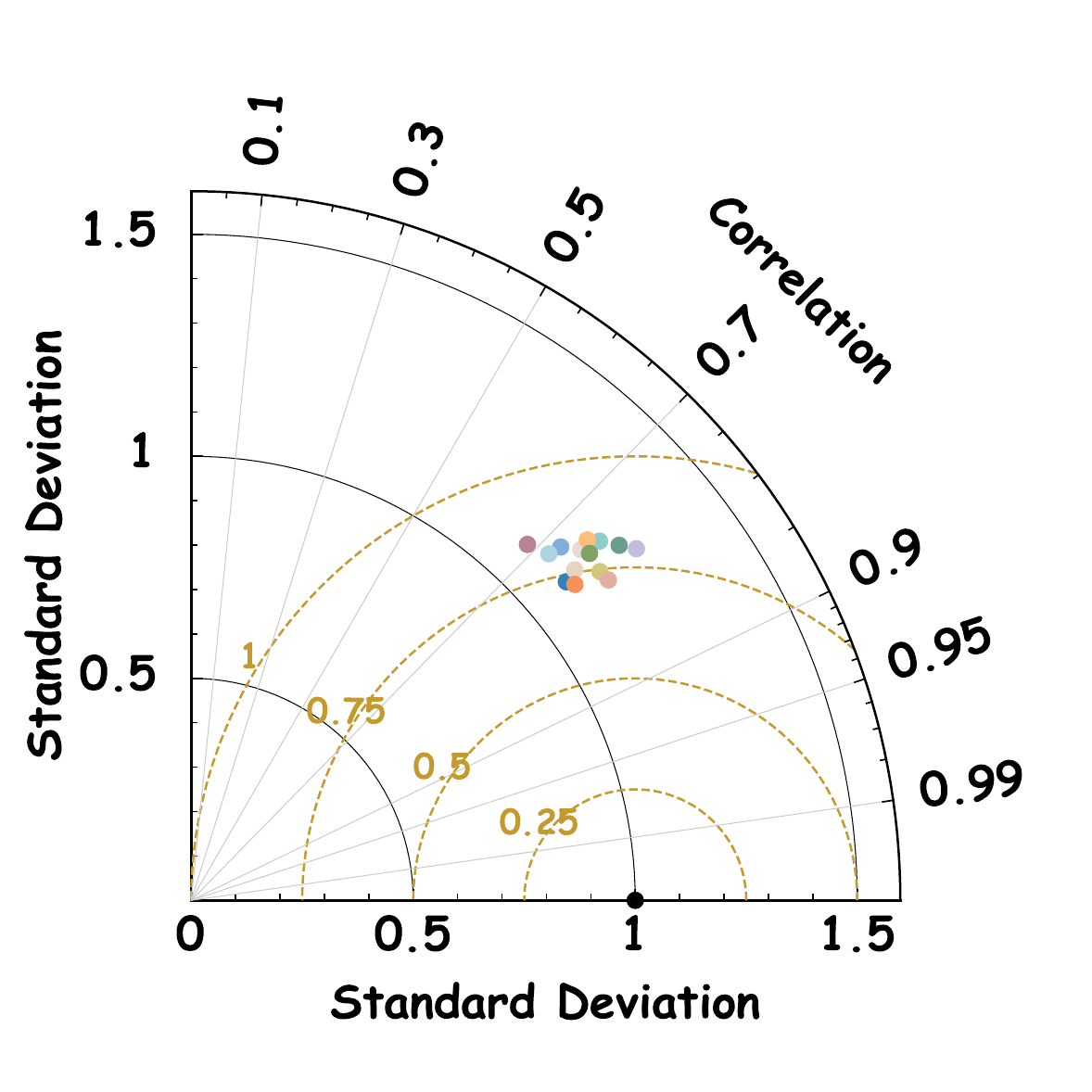}
    \end{subfigure}

    \begin{subfigure}{\linewidth}
        \centering
        \includegraphics[width=1\linewidth]{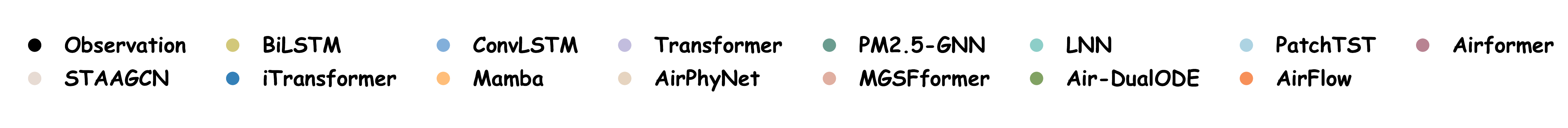} 
    \end{subfigure}

    \caption{Taylor diagram comparing the forecasting performance of different models on the Beijing PM$_{2.5}$ forecasting task under varying input lengths and forecasting horizons.
    The diagram visualizes the correlation coefficient, relative standard deviation, and centered root mean square difference between model predictions and observations.
    The Observation point is the ideal forecasting with perfect correlation, matched variability, and zero centered difference. Models located closer to the Observation point indicate better overall agreement with the ground truth.}
    \label{fig:taylor}
\end{figure*}

To validate the robustness of AirFlow under different input lengths and forecasting horizons, we compared it with baselines on the Beijing PM$_{2.5}$ forecasting task. As shown in Table~\ref{tab:horizons}, although extending the prediction horizon from 6h to 12h increases the errors of all models, AirFlow retains the good performance, reducing 6.06\% MAE than the state-of-the-art baseline.

When fixing the forecasting horizon at 12h, increasing the input length from 12h to 24h leads to different performance trends across models. While longer input sequences provide richer historical context, they may also introduce redundancy and noise. Under this setting, AirFlow also shows good performance, with RMSE increasing only marginally from 21.84 to 22.47 (2.88\%), whereas AirFormer experiences a larger increase from 22.10 to 24.01 (8.64\%). This suggests that AirFlow can effectively retain informative dependencies while suppressing irrelevant historical variations. 

Figure~\ref{fig:taylor} provides Taylor diagram summarizing the forecasting performance under different input lengths and forecasting horizons on the Beijing PM$_{2.5}$ dataset.
The diagram summarizes multiple complementary evaluation aspects, including correlation, relative variability, and centered root mean square difference. AirFlow appears closer to the observation reference point, indicating better alignment with the ground truth in these evaluation aspects.

\subsection{Generalization Analysis}
\noindent\textbf{Cross-Period Case Study}
To evaluate temporal generalization under distribution shifts, we test AirFlow on two non-overlapping periods: Oct. 20--Dec. 31, 2024 and Jan. 1--Sep. 30, 2025. The 2025 records are used only as an additional out-of-period test set. As summarized in Table~\ref{tab:test_set_char}, the later period exhibits a distinct data distribution, with lower average pollution levels but substantially larger extreme values, consistent with changes in meteorological and emission conditions. Table~\ref{tab:temporal_gen} shows that AirFlow maintains stable predictive performance across the two periods. The relative degradation in R$^2$ ranges from 6.07\% to 11.19\% in Beijing and from 4.74\% to 7.99\% in Tianjin. In Hangzhou, the degradation for PM$_{2.5}$, PM$_{10}$, and AQI ranges from 1.99\% to 11.04\%, while NO$_2$ exhibits a slight improvement. These results indicate that AirFlow remains effective under substantial cross-period distribution changes.

\begin{figure*}[htbp]
    \centering
    
    \newcommand{\sidecolwidth}{0.03\linewidth} 
    \newcommand{\maincolwidth}{0.46\linewidth} 
    
    \begin{minipage}{\sidecolwidth} \quad \end{minipage}
    \hfill
    \begin{minipage}{\maincolwidth}
        \centering 
        \includegraphics[height=0.7cm]{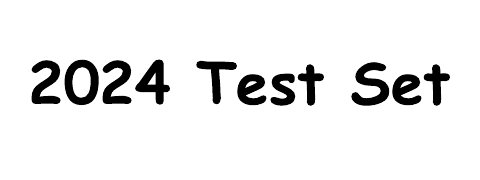}
    \end{minipage}
    \hfill
    \begin{minipage}{\maincolwidth}
        \centering 
        \includegraphics[height=0.7cm]{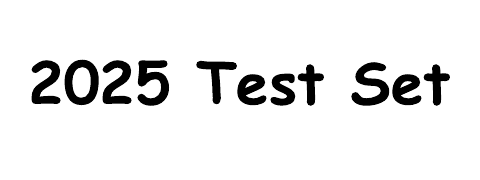}
    \end{minipage}
    

    \begin{minipage}[c]{\sidecolwidth}
        \raggedleft
        \includegraphics[width=0.7\linewidth]{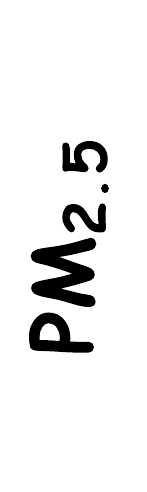}
    \end{minipage}%
    \hfill
    \begin{subfigure}[c]{\maincolwidth}
        \includegraphics[width=\linewidth]{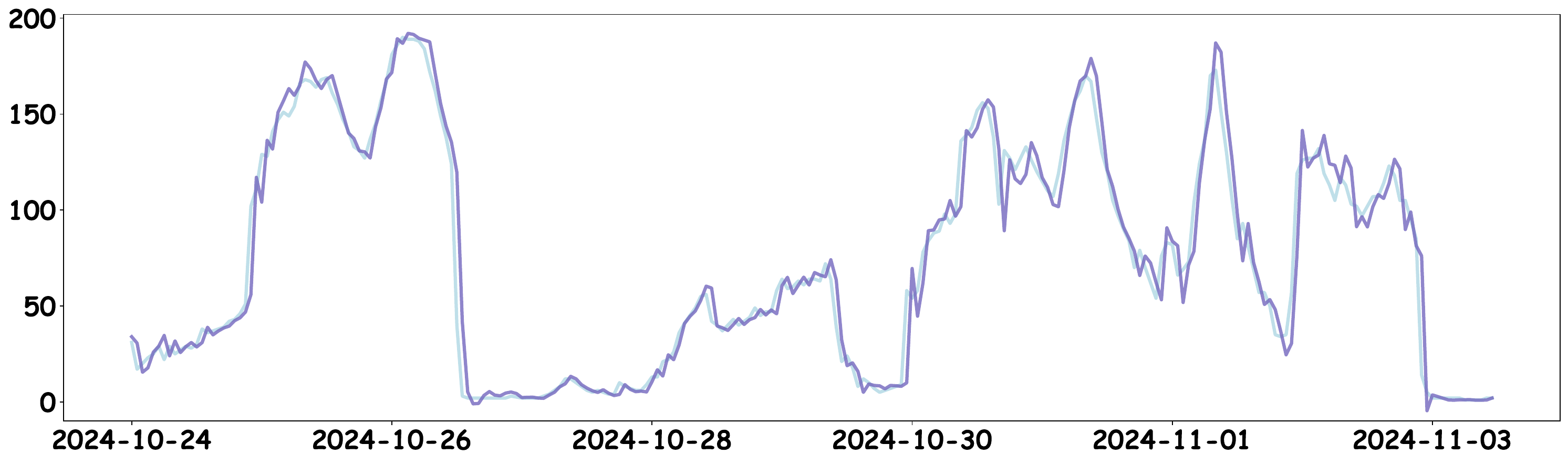}
    \end{subfigure}%
    \hfill
    \begin{subfigure}[c]{\maincolwidth}
        \includegraphics[width=\linewidth]{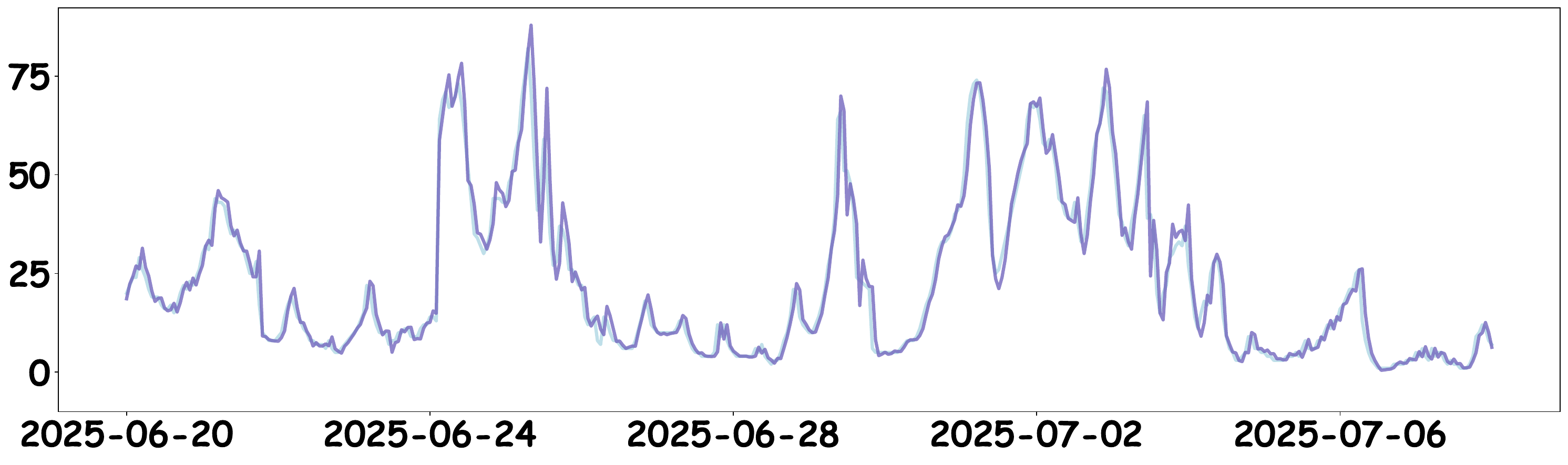}
    \end{subfigure}

    \vspace{0.1cm} 

    \begin{minipage}[c]{\sidecolwidth}
        \raggedleft
        \includegraphics[width=0.7\linewidth]{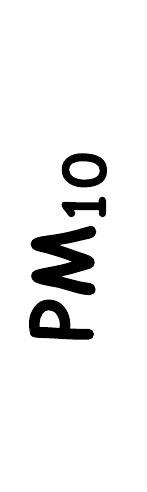}
    \end{minipage}%
    \hfill
    \begin{subfigure}[c]{\maincolwidth}
        \includegraphics[width=\linewidth]{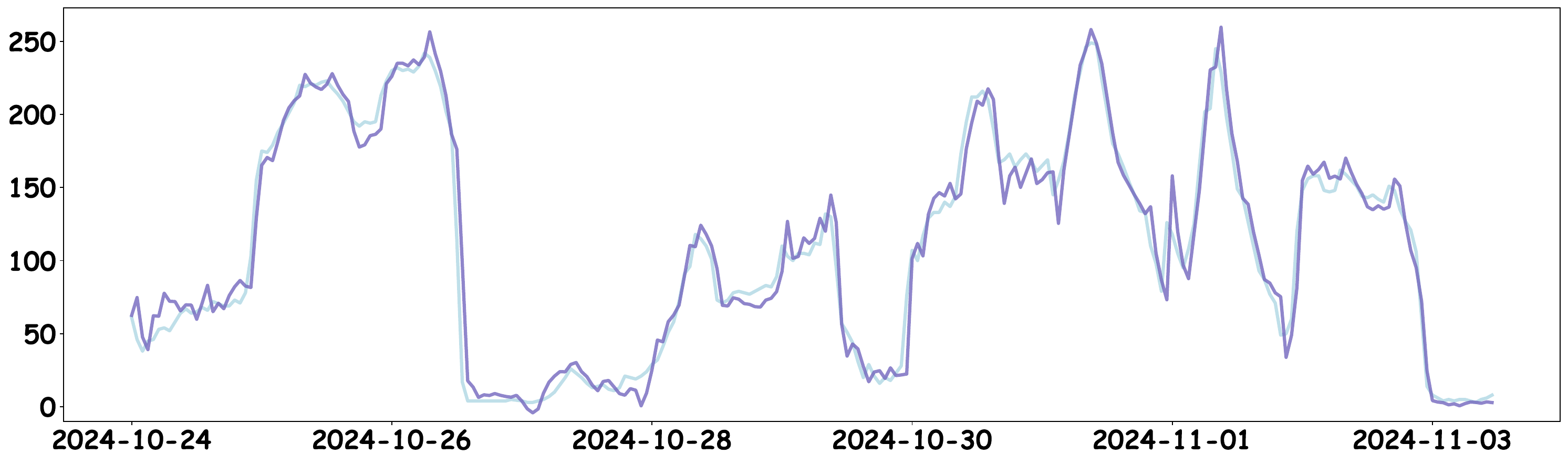}
    \end{subfigure}%
    \hfill
    \begin{subfigure}[c]{\maincolwidth}
        \includegraphics[width=\linewidth]{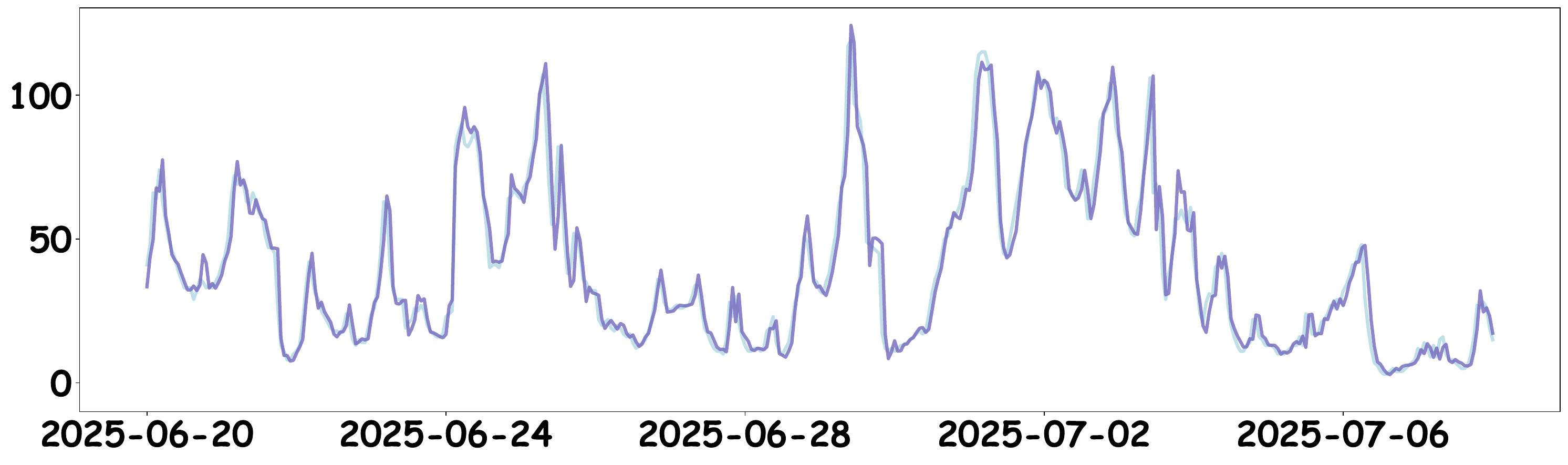}
    \end{subfigure}

    \vspace{0.1cm}

    \begin{minipage}[c]{\sidecolwidth}
        \raggedleft
        \includegraphics[width=0.7\linewidth]{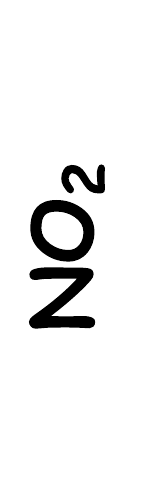}
    \end{minipage}%
    \hfill
    \begin{subfigure}[c]{\maincolwidth}
        \includegraphics[width=\linewidth]{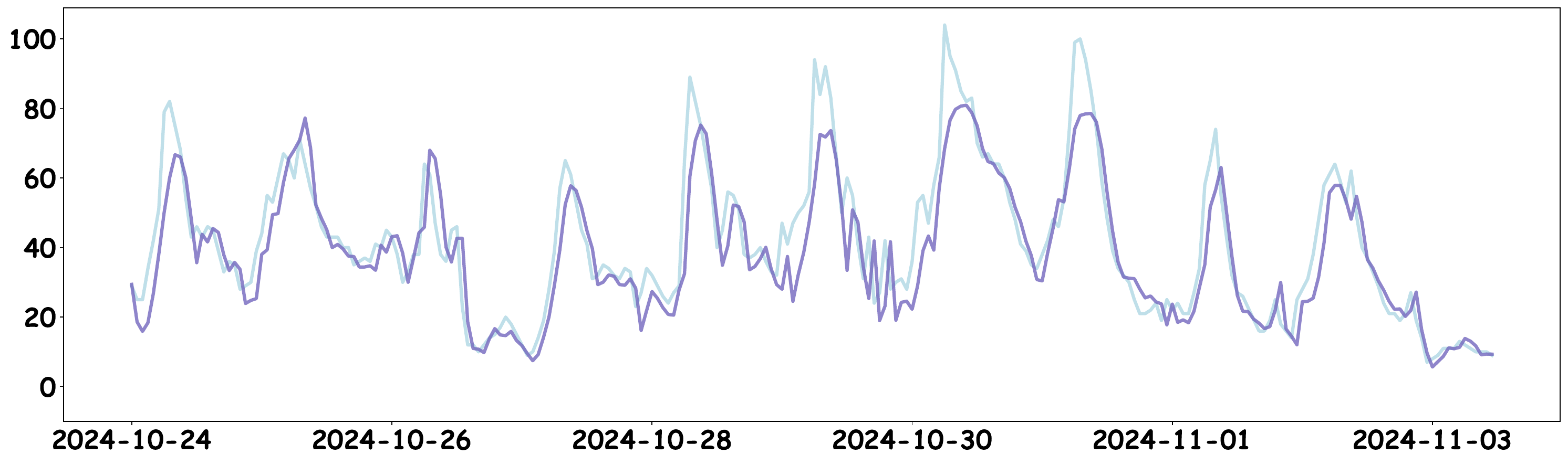}
    \end{subfigure}%
    \hfill
    \begin{subfigure}[c]{\maincolwidth}
        \includegraphics[width=\linewidth]{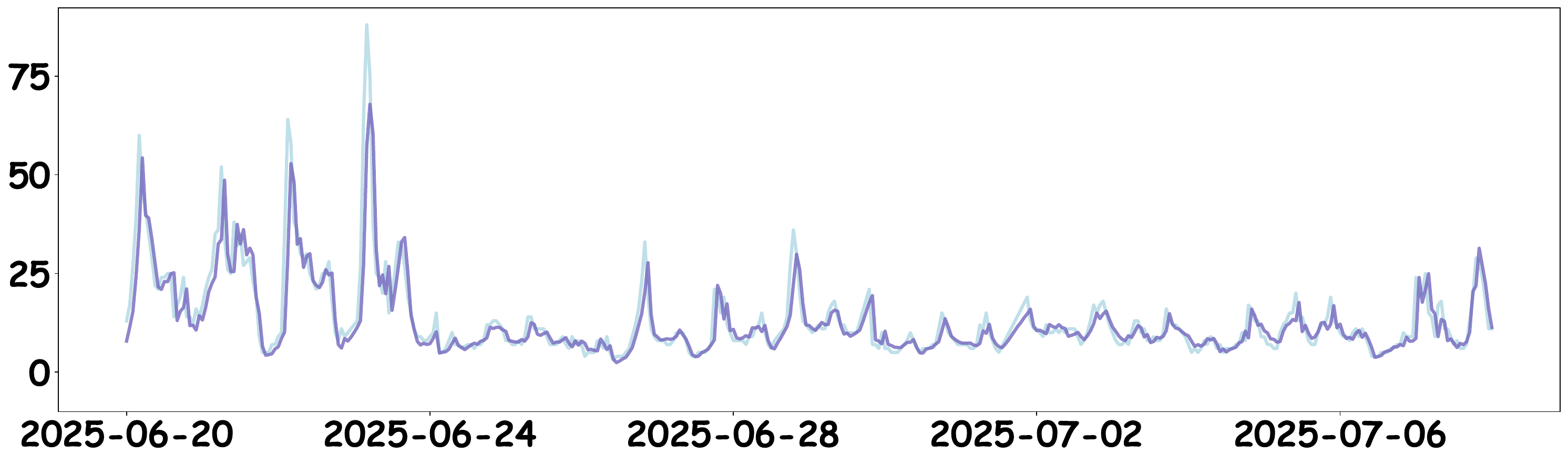}
    \end{subfigure}

    \vspace{0.1cm}

    \begin{minipage}[c]{\sidecolwidth}
        \raggedleft
        \includegraphics[width=0.7\linewidth]{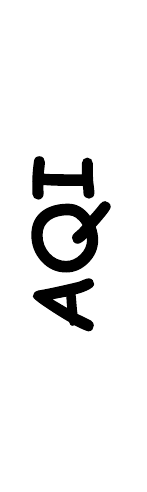}
    \end{minipage}%
    \hfill
    \begin{subfigure}[c]{\maincolwidth}
        \includegraphics[width=\linewidth]{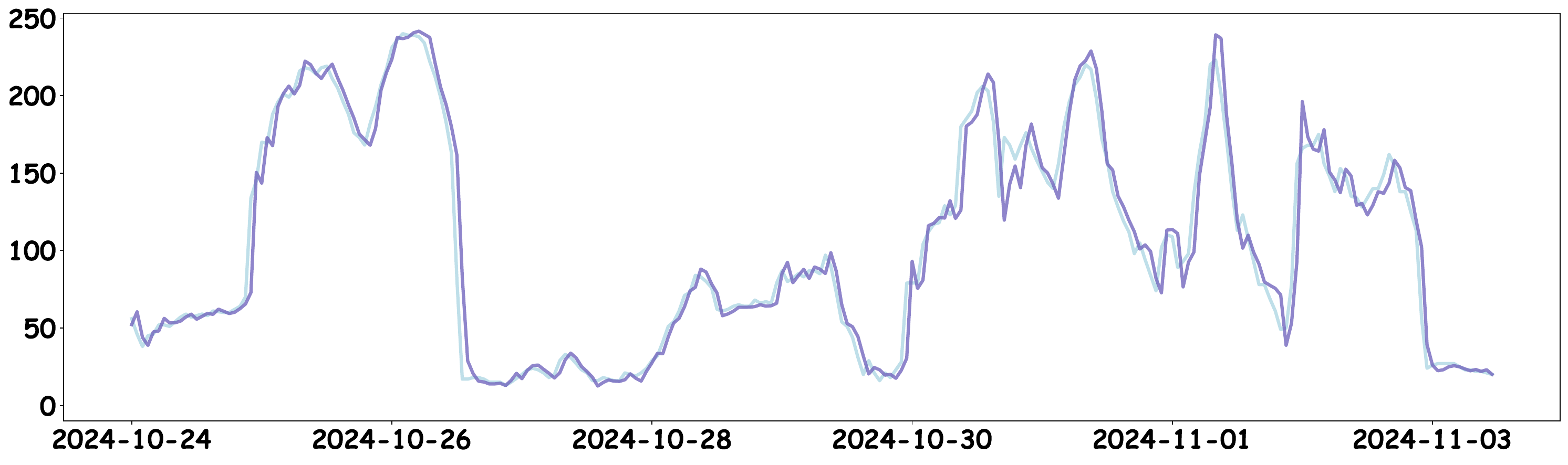}
    \end{subfigure}%
    \hfill
    \begin{subfigure}[c]{\maincolwidth}
        \includegraphics[width=\linewidth]{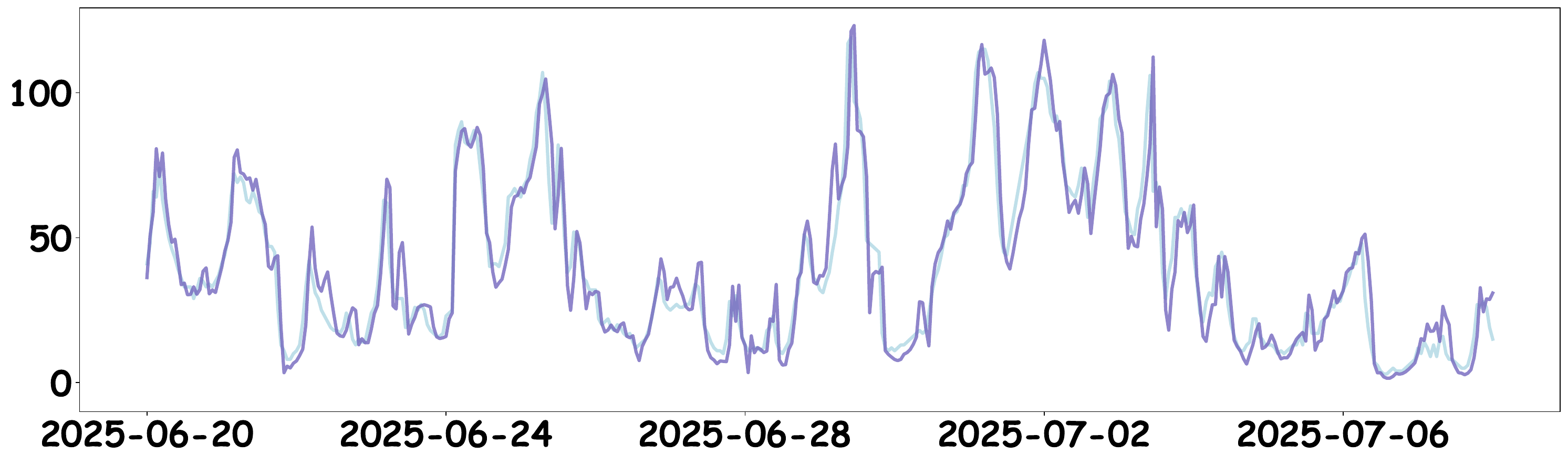}
    \end{subfigure}
    
    \vspace{0cm}

    \begin{subfigure}{\linewidth}
        \centering
        \includegraphics[height=0.7cm]{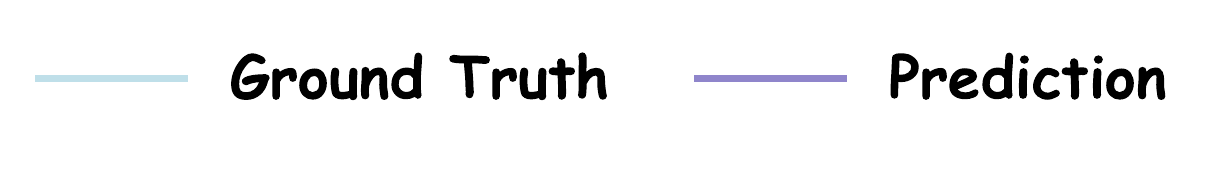} 
    \end{subfigure}

    \caption{Comparison between ground truth and predicted values for PM$_{2.5}$, PM$_{10}$, NO$_2$, and AQI on the 2024 and 2025 test sets from the Beijing dataset.}
    \label{fig:prediction_comparison}
\end{figure*}

The cross-pollutant results further reveal different levels of temporal stability. NO$_2$ is relatively stable across cities, which is consistent with its more regular activity-related emission patterns, whereas PM$_{10}$ undergoes larger performance degradation because it is more sensitive to changing dust, transport, and removal conditions. Figure~\ref{fig:prediction_comparison} provides representative forecasting trajectories. AirFlow captures abrupt surge-and-drop patterns in particulate matter, including the event around Nov. 2, 2024, while preserving the preceding concentration context. During the summer of 2025, it continues to follow the overall trajectory under frequent short-term fluctuations.

\begin{table}[!htbp]
    \centering
        \caption{Multi-pollutant Test Set Characteristics Comparison.}
        \label{tab:test_set_char}
        \resizebox{\columnwidth}{!}{%
        \begin{tabular}{cccccccc}
            \toprule
            \multirow{2}{*}{Dataset} & \multirow{2}{*}{Pollutant} & \multicolumn{3}{c}{2024 Test Set} & \multicolumn{3}{c}{2025 Test Set} \\
            \cmidrule(lr){3-5} \cmidrule(lr){6-8}
            & & Mean & Std. & Max & Mean & Std. & Max \\
            \midrule
            \multirow{4}{*}{Beijing}
            & PM$_{2.5}$ & 33.55 & 37.73 & 227 & 25.67 & 25.93 & 455 \\
            & PM$_{10}$ & 60.70 & 57.07 & 562 & 49.74 & 44.26 & 1111 \\
            & NO$_2$ & 31.46 & 22.65 & 198 & 19.62 & 18.05 & 202 \\
            & AQI & 55.45 & 45.22 & 462 & 50.32 & 34.06 & 500 \\
            \midrule
            \multirow{4}{*}{Tianjin}
            & PM$_{2.5}$ & 40.21 & 35.63 & 260 & 32.57 & 31.90 & 1048 \\
            & PM$_{10}$ & 69.14 & 49.33 & 188 & 60.70 & 48.66 & 1186 \\
            & NO$_2$ & 44.44 & 25.63 & 188 & 27.02 & 21.85 & 213 \\
            & AQI & 63.22 & 41.72 & 310 & 59.07 & 37.84 & 500 \\
            \midrule
            \multirow{4}{*}{Hangzhou}
            & PM$_{2.5}$ & 33.39 & 24.23 & 175 & 29.00 & 23.64 & 558 \\
            & PM$_{10}$ & 54.78 & 32.81 & 239 & 48.68 & 38.77 & 843 \\
            & NO$_2$ & 38.08 & 18.22 & 157 & 25.40 & 18.42 & 186 \\
            & AQI & 53.52 & 28.36 & 225 & 49.76 & 31.42 & 500 \\
            \bottomrule
        \end{tabular}%
        }
\end{table}

\label{sec:Temporal}
\begin{table}[!t]
    \centering
    \caption{Case study performance of AirFlow on different time periods.}
    \label{tab:temporal_gen}

    \setlength{\tabcolsep}{3pt}
    \resizebox{\columnwidth}{!}{
    \begin{tabular}{lccccccc}
        \toprule
        \multirow{2}{*}{Dataset} & \multirow{2}{*}{Pollutant} & \multicolumn{3}{c}{10/20--12/31 2024} & \multicolumn{3}{c}{01/01--09/30 2025} \\
        \cmidrule(lr){3-5} \cmidrule(lr){6-8}
        & & RMSE $\downarrow$ & MAE $\downarrow$ & R$^2$ $\uparrow$ & RMSE $\downarrow$ &  MAE $\downarrow$ & R$^2$ $\uparrow$ \\
        \midrule
        \multirow{4}{*}{\shortstack{Beijing}}
        & PM$_{2.5}$ & 14.42 & 7.47 & 0.8513 & 12.79 & 7.11 & 0.7582 \\
        & PM$_{10}$ & 22.43 & 13.02 & 0.8443 & 24.76 & 12.87 & 0.7532 \\
        & NO$_2$ & 12.65 & 8.55 & 0.6889 & 11.02 & 6.62 & 0.6471 \\
        & AQI & 19.24 & 10.87 & 0.8184 & 19.10 & 10.81 & 0.7268 \\
        \midrule
        \multirow{4}{*}{\shortstack{Tianjin}}
        & PM$_{2.5}$ & 18.87 & 10.97 & 0.7124 & 18.13 & 9.70 & 0.6746 \\
        & PM$_{10}$ & 26.87 & 16.81 & 0.7046 & 28.69 & 16.42 & 0.6483 \\
        & NO$_2$ & 14.54 & 9.78 & 0.6793 & 12.93 & 7.98 & 0.6471 \\
        & AQI & 22.58 & 13.48 & 0.7023 & 24.16 & 14.42 & 0.6657 \\
        \midrule
        \multirow{4}{*}{\shortstack{Hangzhou}}
        & PM$_{2.5}$ & 12.42 & 7.28 & 0.7455 & 12.23 & 6.79 & 0.7306 \\
        & PM$_{10}$ & 16.88 & 10.56 & 0.7224 & 22.17 & 10.46 & 0.6427 \\
        & NO$_2$ & 10.85 & 7.43 & 0.6453 & 9.63 & 6.06 & 0.7241 \\
        & AQI & 14.84 & 9.02 & 0.7105 & 17.15 & 8.93 & 0.6767 \\
        \bottomrule
    \end{tabular}
    }
\end{table}

\noindent\textbf{Cross-City Transfer Learning}
\label{sec:transfer_learning}
A primary challenge in \aqf{} lies in the diversity of geographical topologies, climatic conditions, and emission structures across cities. By modeling intrinsic multi-scale dependencies and temporal evolution rates, AirFlow capture universal atmospheric dynamics beyond specific regions. To further leverage this property, we adopt a cross-city transfer learning (TL) strategy, where the model is initialized with pre-trained weights from Beijing and fine-tuned on two target domains: Tianjin (physically proximate) and Hangzhou (geographically distant with distinct emission profiles).

As shown in Table~\ref{tab:transfer_results}, TL strategy yields improvements in Tianjin, suggesting that the model transfers shared regional transport dynamics and background atmospheric states within geographically proximate areas. This improvement indicates that the learned representations encode not only local correlations but also physically meaningful cross-city dependencies. In contrast, in Hangzhou, the model still maintains competitive performance across major pollutants, implying that the captured temporal evolution patterns are invariant to regional differences. The relatively slight variation observed for NO$_2$ further highlights that gaseous pollutants are more sensitive to localized emission sources such as traffic and urban layouts, making them less transferable than particulate matter. These results suggest that AirFlow learns transferable temporal dynamics while retaining sufficient flexibility to adapt to city-specific characteristics, enabling stable performance across both near-domain and far-domain scenarios.
\begin{table*}[!t]
    \centering
    \caption{Comparative performance of AirFlow and its transfer learning variant AirFlow (TL) on Tianjin and Hangzhou datasets, where models are initialized with pre-trained weights from Beijing and fine-tuned on each target city's training data.}
    \label{tab:transfer_results}
    \resizebox{\textwidth}{!}{
    \begin{tabular}{llcccccccccccc}
        \toprule
        \multirow{2}{*}{Dataset} & \multirow{2}{*}{Model} & \multicolumn{3}{c}{PM$_{2.5}$} & \multicolumn{3}{c}{PM$_{10}$} & \multicolumn{3}{c}{NO$_2$} & \multicolumn{3}{c}{AQI} \\
        \cmidrule(lr){3-5} \cmidrule(lr){6-8} \cmidrule(lr){9-11} \cmidrule(lr){12-14}
        & & RMSE $\downarrow$ & MAE $\downarrow$ & R$^2\uparrow$ & RMSE $\downarrow$ & MAE $\downarrow$ & R$^2\uparrow$ & RMSE $\downarrow$ & MAE $\downarrow$ & R$^2\uparrow$ & RMSE $\downarrow$ & MAE $\downarrow$ & R$^2\uparrow$ \\        
        \midrule
        \multirow{2}{*}{Tianjin} 
        & AirFlow & 18.87 & 10.97 & 0.7124 & 26.87 & 16.81 & 0.7046 & 14.54 & 9.78 & 0.6793 & 22.58 & 13.48 & 0.7023 \\
        & AirFlow (TL) & 18.52 & 10.76 & 0.7182 & 25.98 & 16.37 & 0.7153 & 14.32 & 9.64 & 0.6832 & 22.17 & 13.12 & 0.7111 \\
        \midrule
        \multirow{2}{*}{Hangzhou} 
        & AirFlow & 12.42 & 7.28 & 0.7455 & 16.88 & 10.56 & 0.7224 & 10.85 & 7.43 & 0.6453 & 14.84 & 9.02 & 0.7105 \\
        & AirFlow (TL) & 12.30 & 7.17 & 0.7501 & 16.64 & 10.33 & 0.7319 & 10.97 & 7.45 & 0.6421 & 14.46 & 8.77 & 0.7396 \\
        \bottomrule
    \end{tabular}
    }
\end{table*}

\subsection{Visualization of Experimental Results}
\begin{figure*}[htbp]
    \centering

    \begin{minipage}{0.04\linewidth} \quad \end{minipage}%
    \hspace{0.05cm}
    \begin{minipage}{0.44\linewidth}
        \centering
        \includegraphics[height=0.9cm]{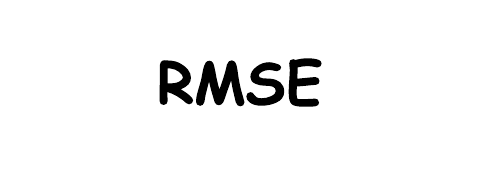}
    \end{minipage}%
    \hspace{0.05cm}
    \begin{minipage}{0.44\linewidth}
        \centering
        \includegraphics[height=0.9cm]{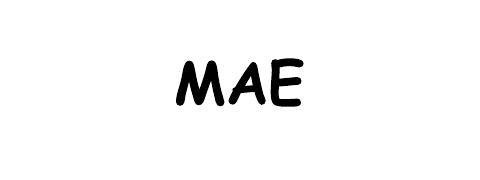}
    \end{minipage}
    
    
    \begin{minipage}[c]{0.03\linewidth} 
        \centering
        \includegraphics[width=\linewidth]{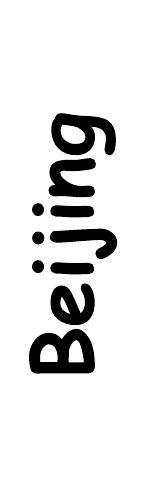}
    \end{minipage}%
    \hspace{0.05cm}
    \begin{subfigure}[c]{0.44\linewidth}
        \centering
        \includegraphics[width=\linewidth]{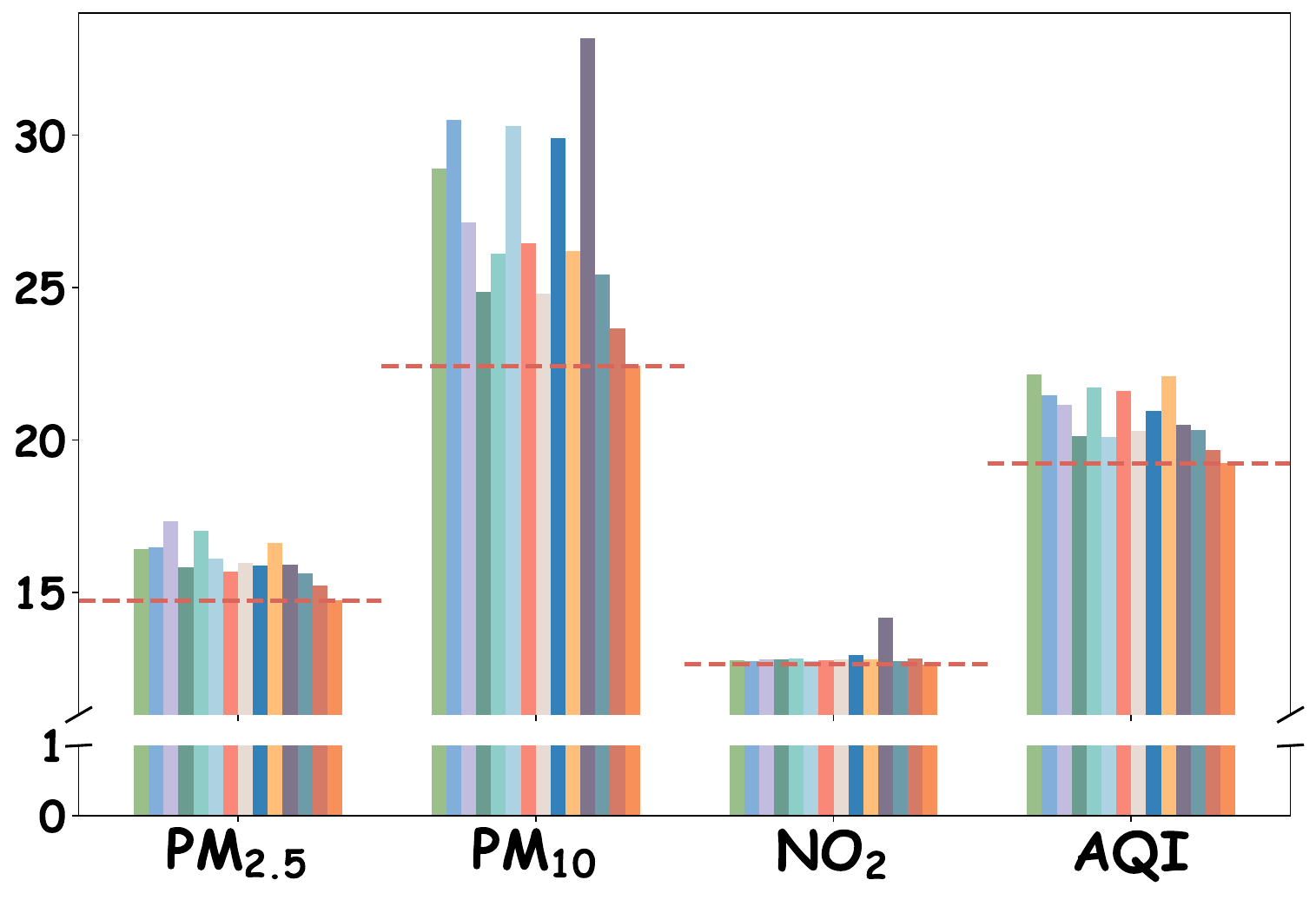}
    \end{subfigure}%
    \hspace{0.05cm}
    \begin{subfigure}[c]{0.44\linewidth}
        \centering
        \includegraphics[width=\linewidth]{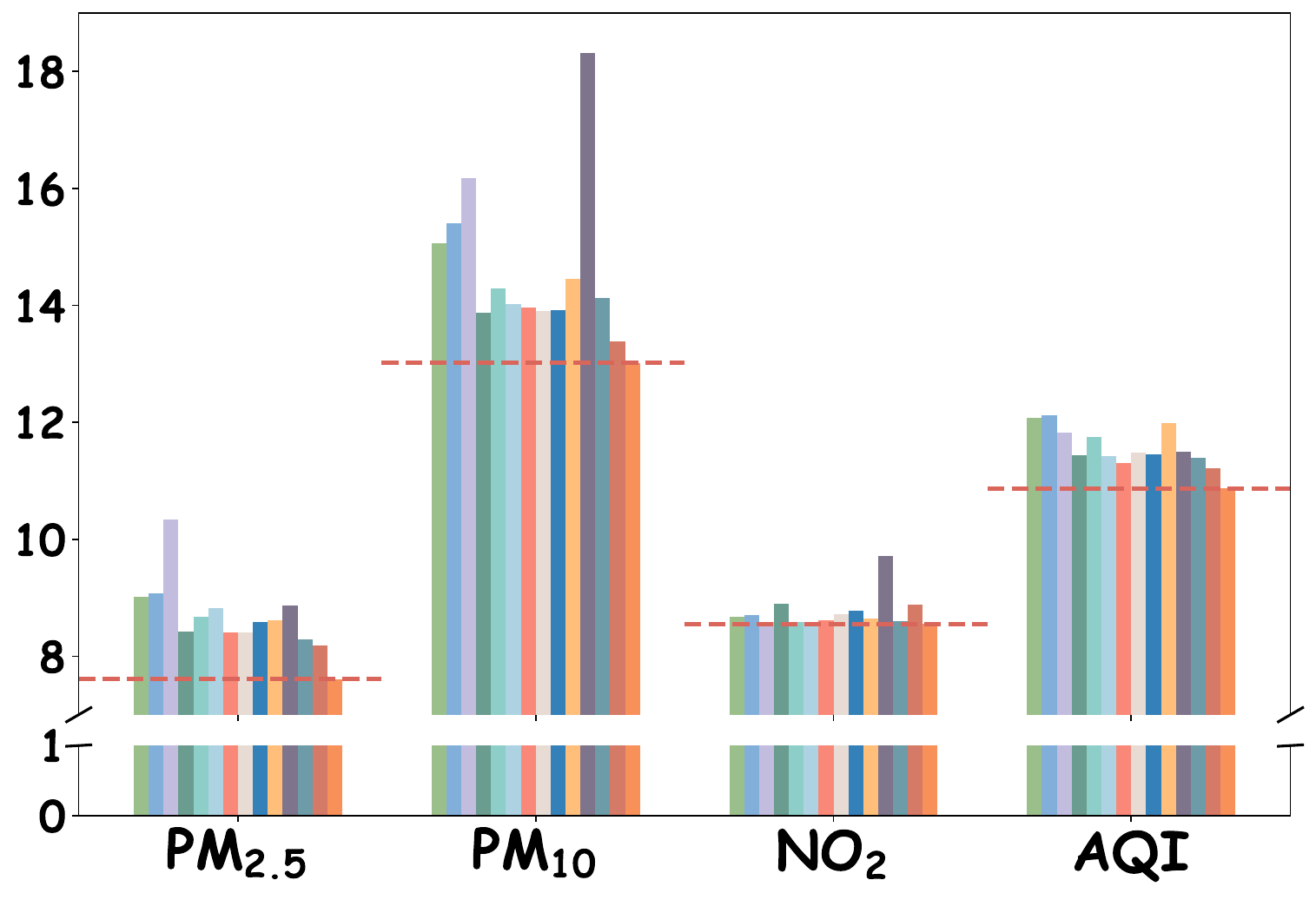}
    \end{subfigure}
    
    \vspace{0.1cm}
    
    \begin{minipage}[c]{0.03\linewidth} 
        \centering
        \includegraphics[width=\linewidth]{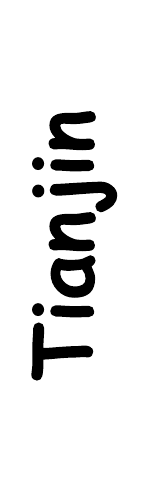}
    \end{minipage}%
    \hspace{0.05cm}
    \begin{subfigure}[c]{0.44\linewidth}
        \centering
        \includegraphics[width=\linewidth]{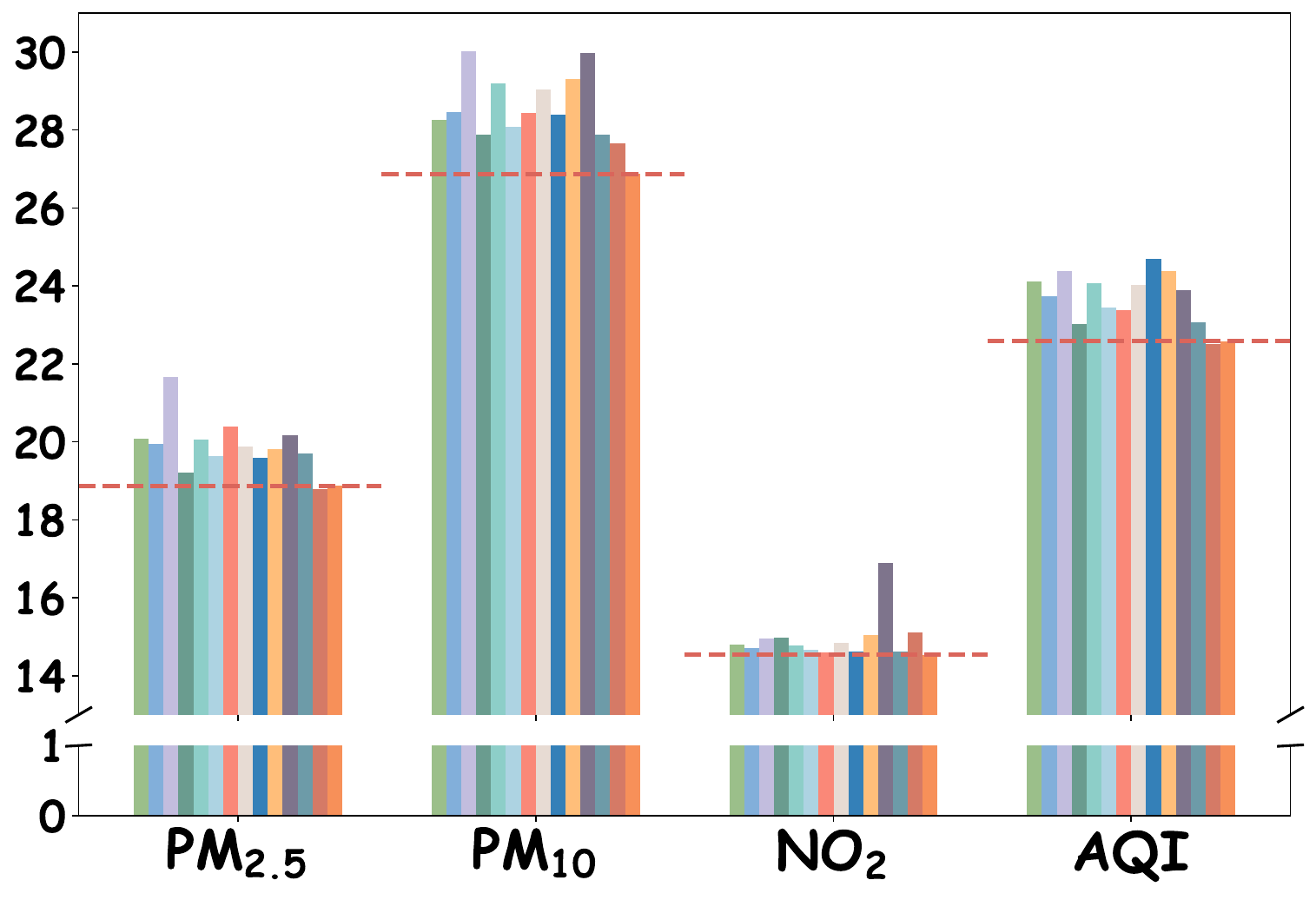}
    \end{subfigure}%
    \hspace{0.05cm}
    \begin{subfigure}[c]{0.44\linewidth}
        \centering
        \includegraphics[width=\linewidth]{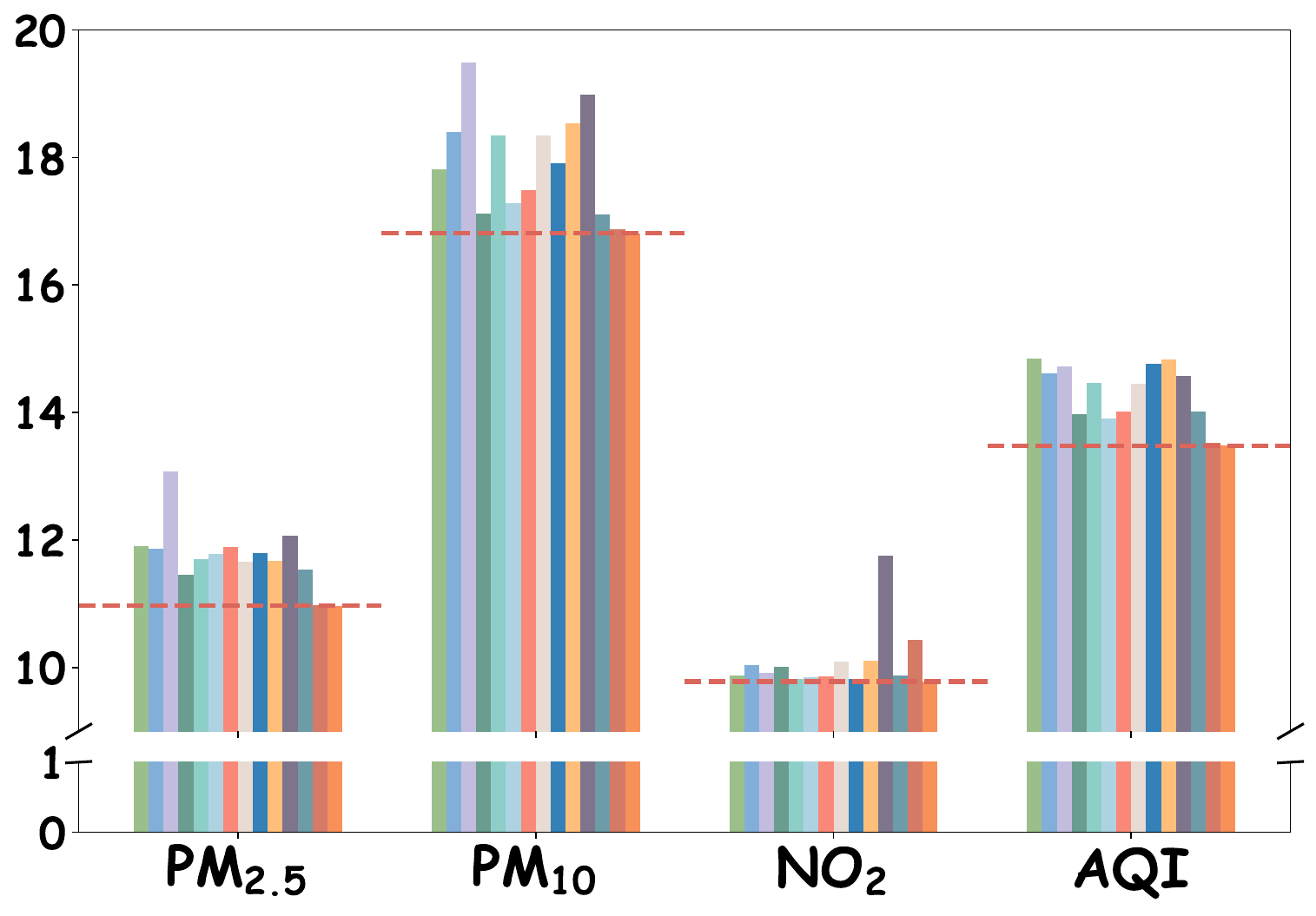}
    \end{subfigure}
    
    \vspace{0.1cm}
    \begin{minipage}[c]{0.03\linewidth} 
        \centering
        \includegraphics[width=\linewidth]{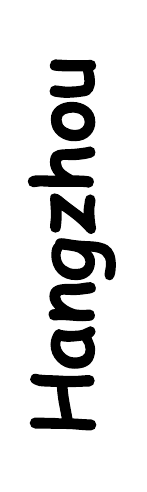}
    \end{minipage}%
    \hspace{0.05cm}
    \begin{subfigure}[c]{0.44\linewidth}
        \centering
        \includegraphics[width=\linewidth]{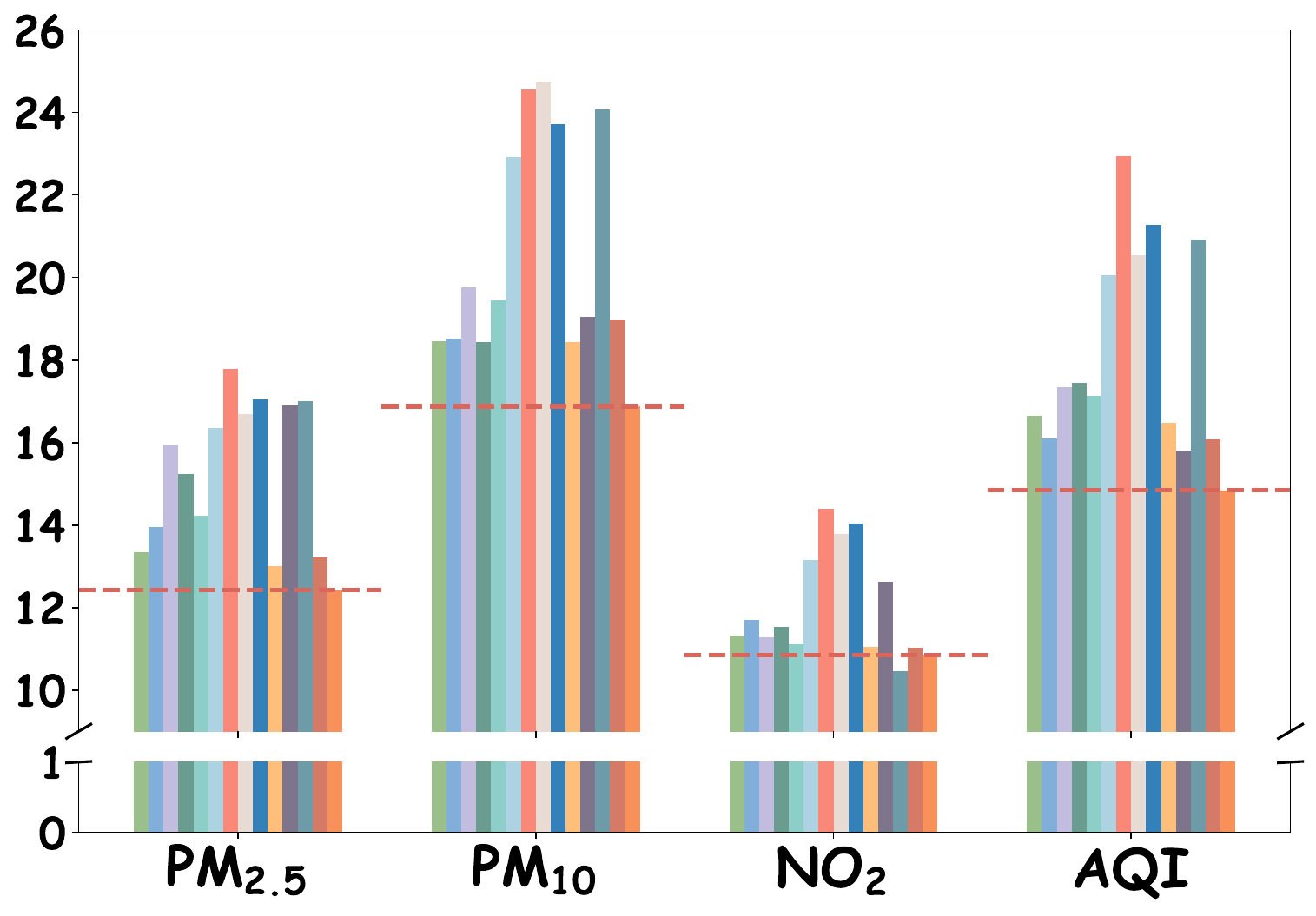}
    \end{subfigure}%
    \hspace{0.05cm}
    \begin{subfigure}[c]{0.44\linewidth}
        \centering
        \includegraphics[width=\linewidth]{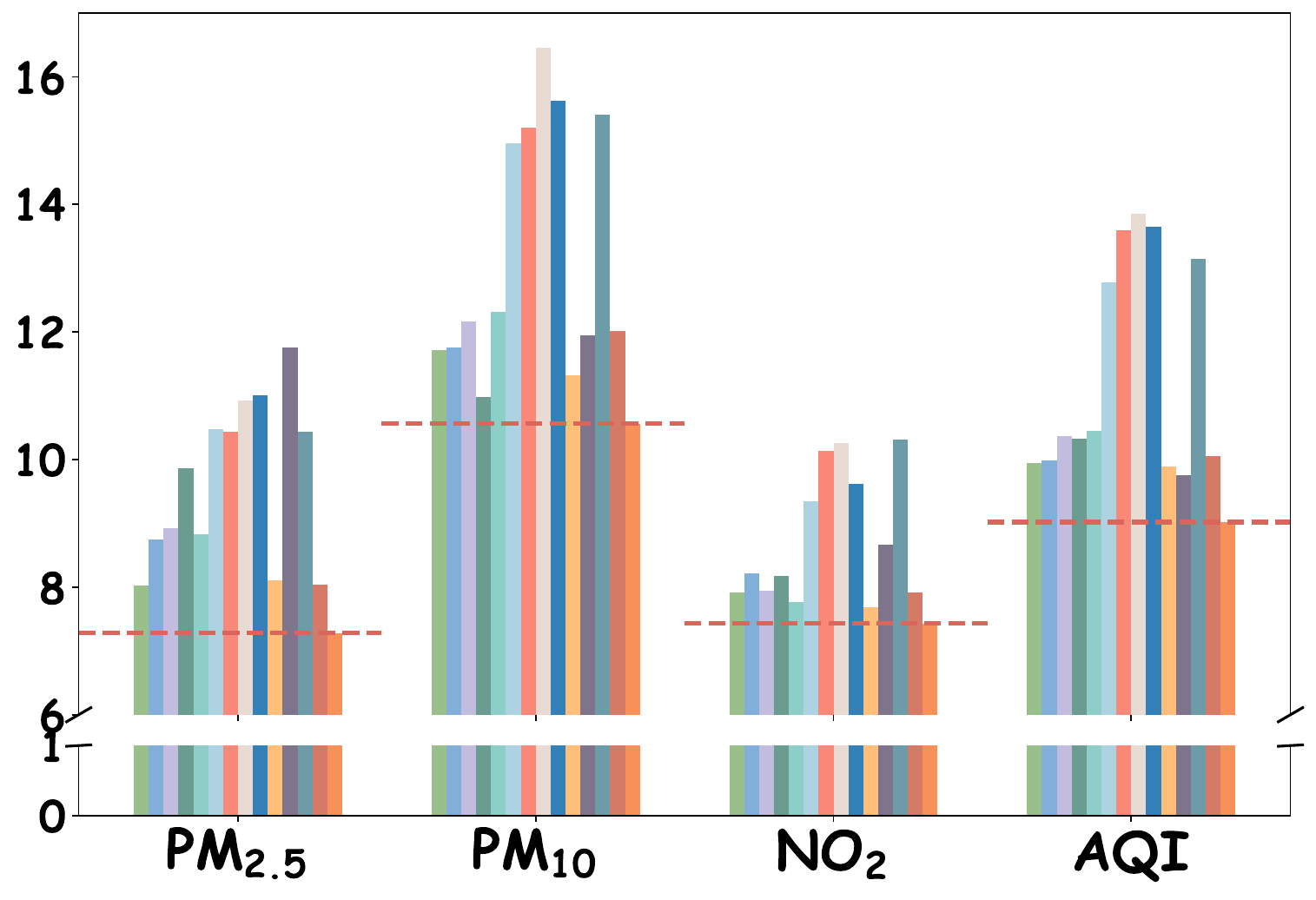}
    \end{subfigure}
    
    \vspace{0.2cm} 
    \begin{minipage}{\linewidth} 
        \centering
        \includegraphics[width=0.9\linewidth]{MAE_legend.pdf}
    \end{minipage}

    \caption{Comparative performance of baseline and proposed models based on RMSE and MAE. }
    \label{fig:mae}
    
\end{figure*}

\begin{figure*}[htbp]
    \centering
    
    \newcommand{\leftlabelwidth}{0.03\linewidth}
    \newcommand{\subfigwidth}{0.23\linewidth}
    
    \begin{minipage}{\leftlabelwidth} \quad \end{minipage} 
    \hfill
    \begin{minipage}{\subfigwidth} 
        \centering 
        \hspace{0mm} 
        \includegraphics[height=0.9cm]{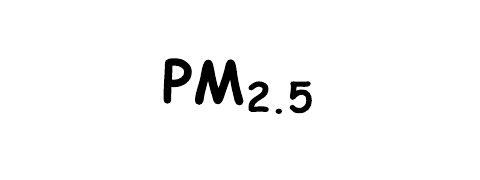} 
    \end{minipage}%
    \hfill
    \begin{minipage}{\subfigwidth} 
        \centering 
        \hspace{0mm}
        \includegraphics[height=0.9cm]{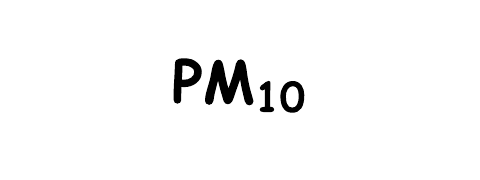} 
    \end{minipage}%
    \hfill
    \begin{minipage}{\subfigwidth} 
        \centering 
        \hspace{0mm}
        \includegraphics[height=0.9cm]{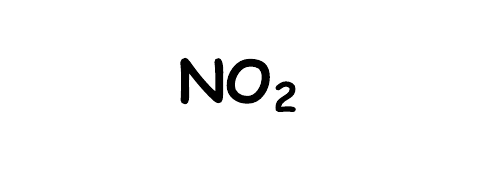} 
    \end{minipage}%
    \hfill
    \begin{minipage}{\subfigwidth} 
        \centering 
        \hspace{0mm}
        \includegraphics[height=0.9cm]{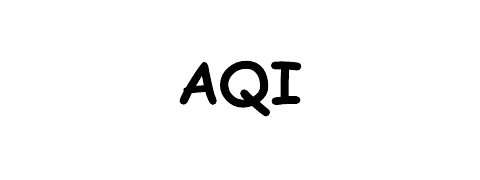} 
    \end{minipage}

    \begin{minipage}[c]{0.03\linewidth} 
        \centering
        \includegraphics[width=\linewidth]{beijing_sites.pdf}
    \end{minipage}%
    \hfill
    \begin{subfigure}[c]{\subfigwidth}
        \includegraphics[width=\linewidth]{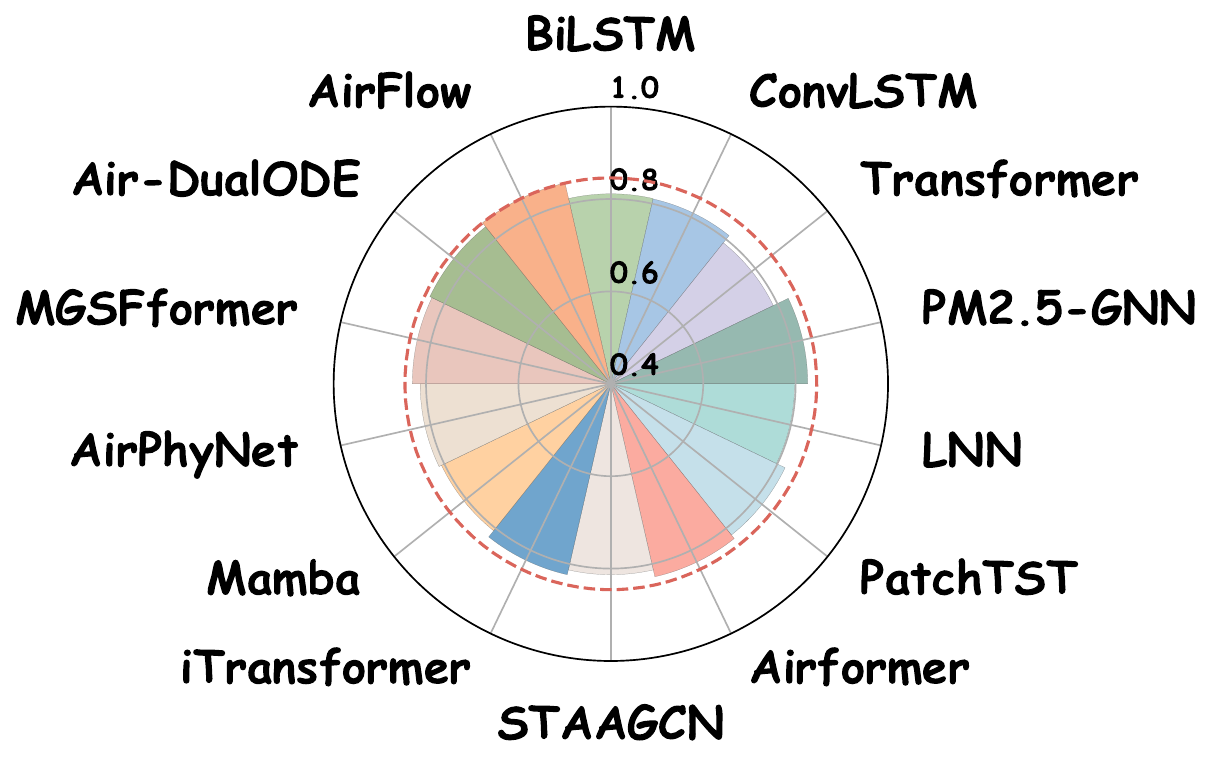}
    \end{subfigure}%
    \hfill
    \begin{subfigure}[c]{\subfigwidth}
        \includegraphics[width=\linewidth]{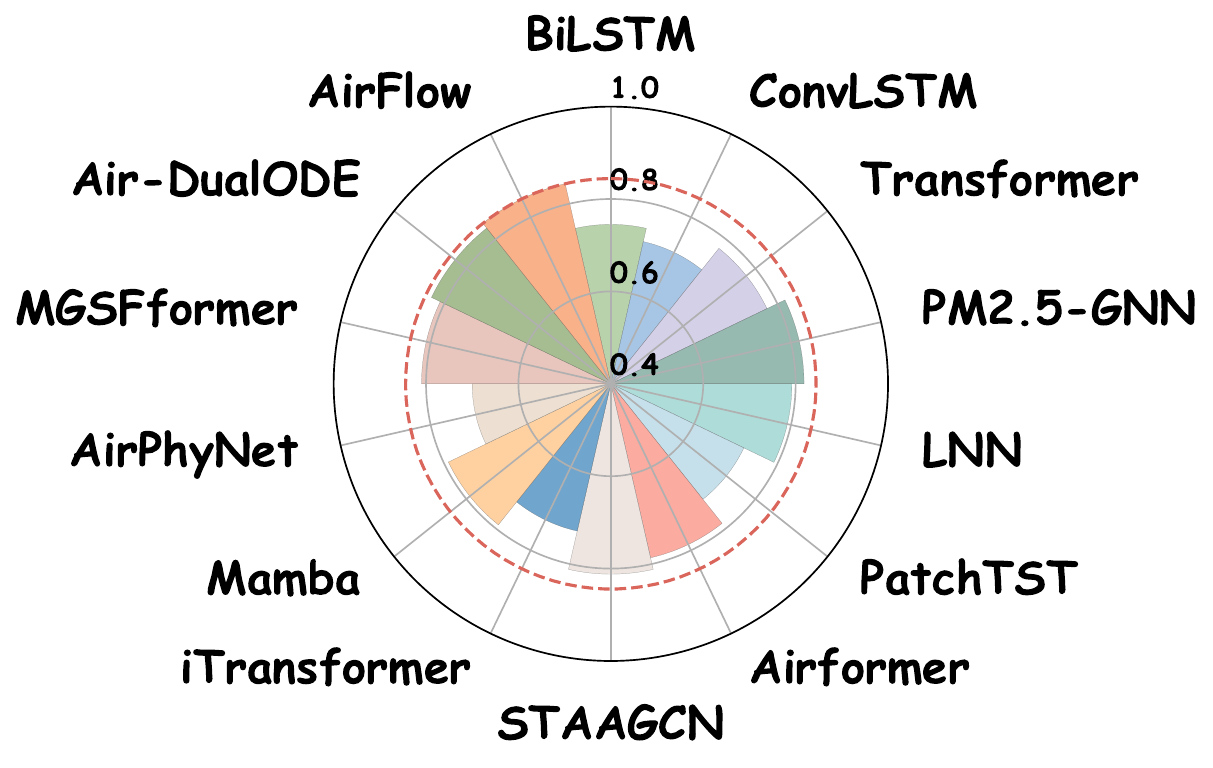}
    \end{subfigure}%
    \hfill
    \begin{subfigure}[c]{\subfigwidth}
        \includegraphics[width=\linewidth]{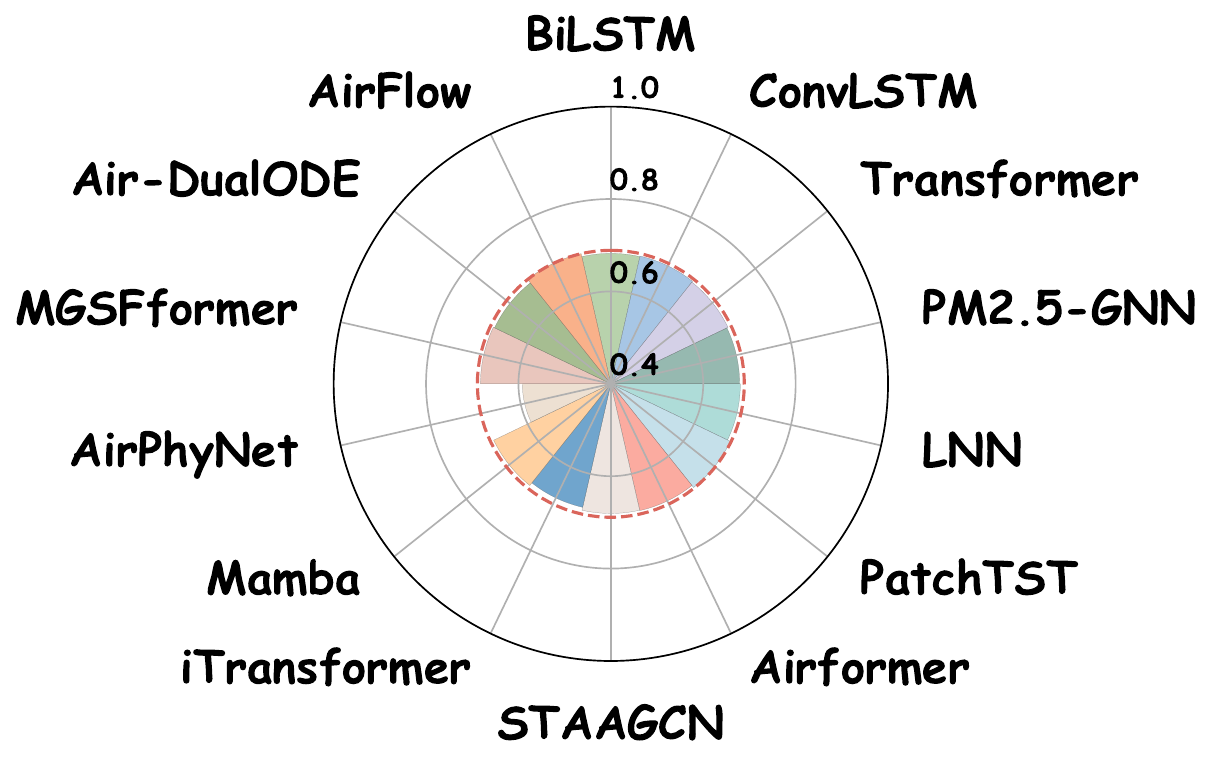}
    \end{subfigure}%
    \hfill
    \begin{subfigure}[c]{\subfigwidth}
        \includegraphics[width=\linewidth]{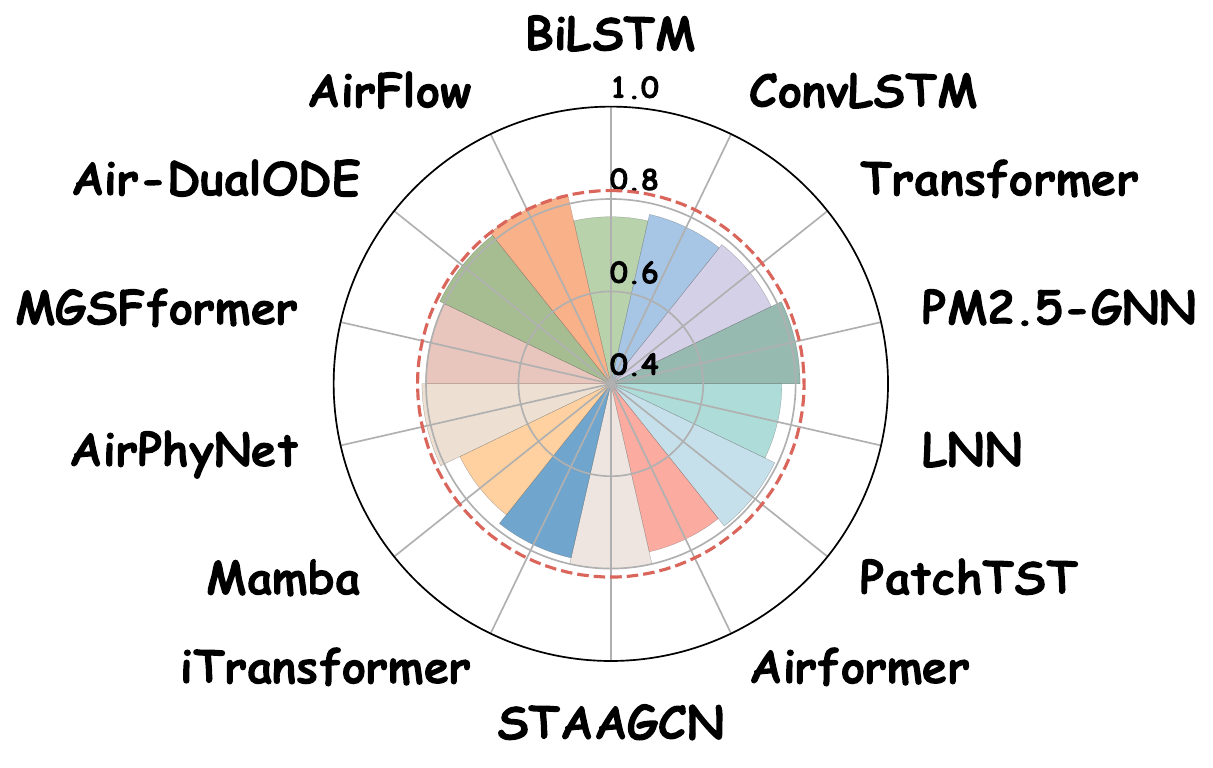}
    \end{subfigure}
    
    \vspace{0.15cm} 
    
    \begin{minipage}[c]{0.03\linewidth} 
        \centering
        \includegraphics[width=\linewidth]{tianjin_sites.pdf}
    \end{minipage}%
    \hfill
    \begin{subfigure}[c]{\subfigwidth}
        \includegraphics[width=\linewidth]{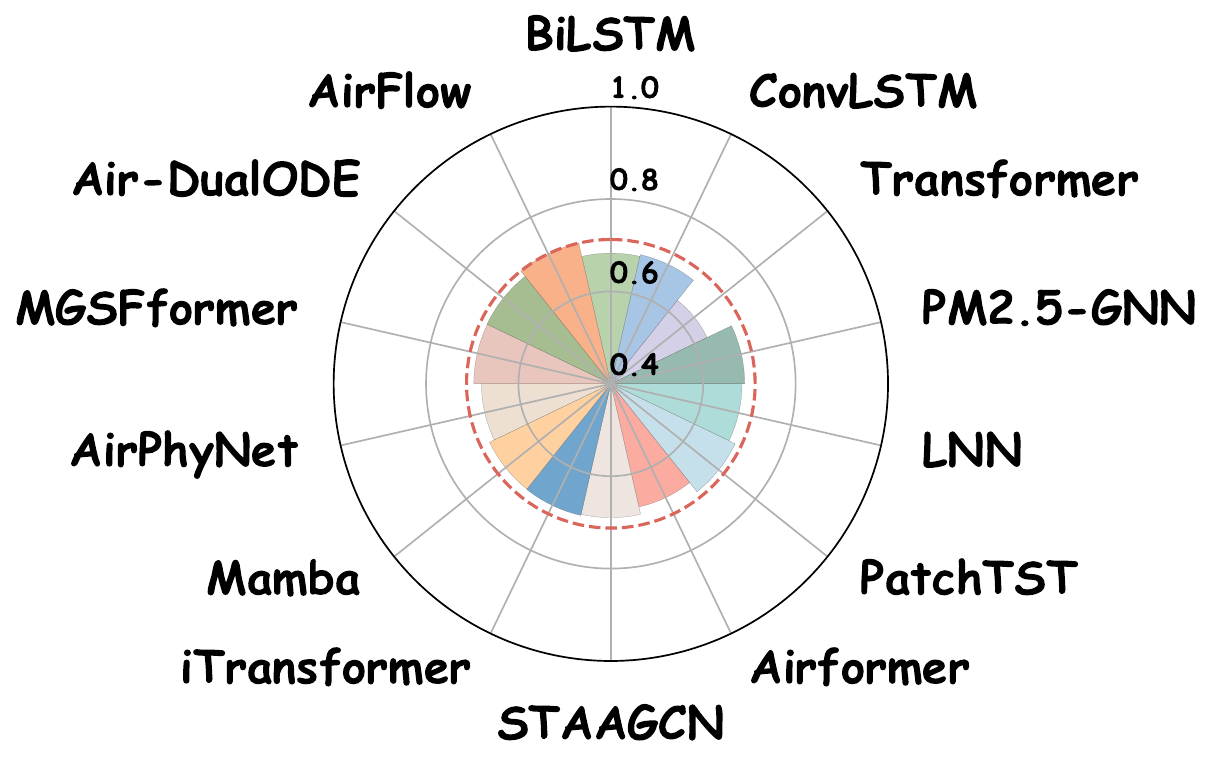}
    \end{subfigure}%
    \hfill
    \begin{subfigure}[c]{\subfigwidth}
        \includegraphics[width=\linewidth]{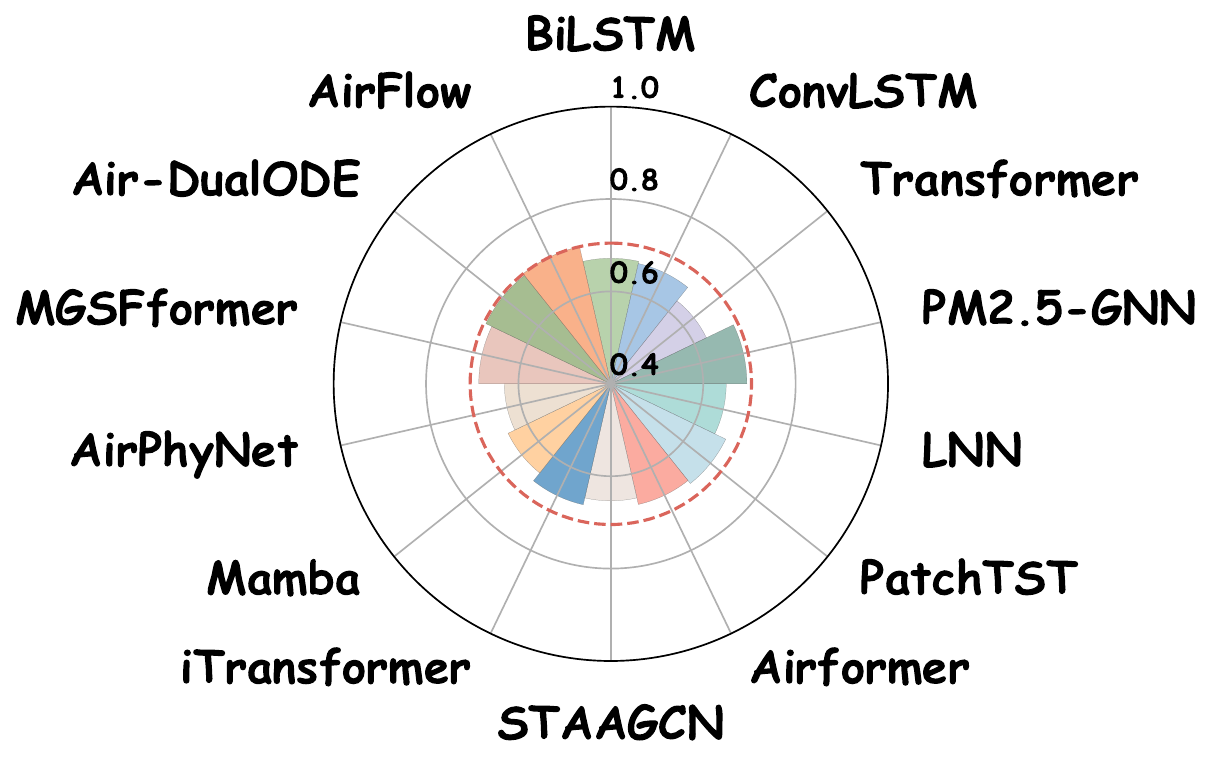}
    \end{subfigure}%
    \hfill
    \begin{subfigure}[c]{\subfigwidth}
        \includegraphics[width=\linewidth]{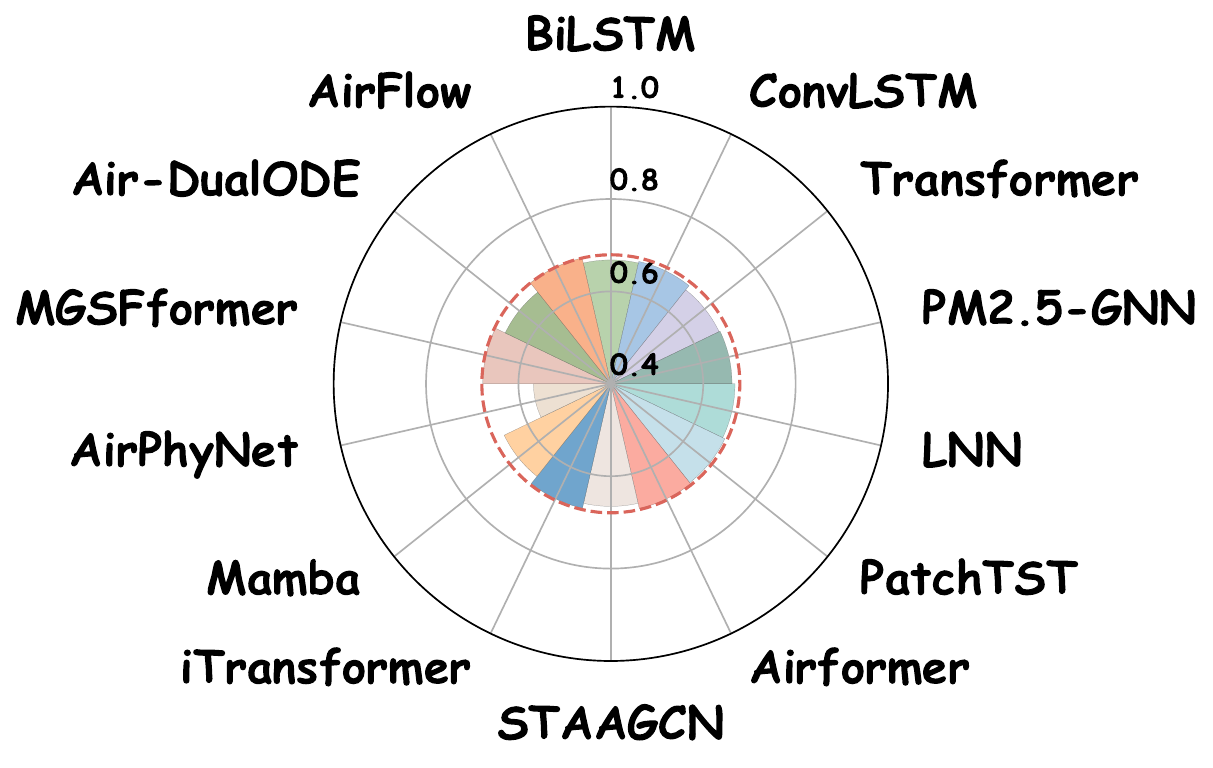}
    \end{subfigure}%
    \hfill
    \begin{subfigure}[c]{\subfigwidth}
        \includegraphics[width=\linewidth]{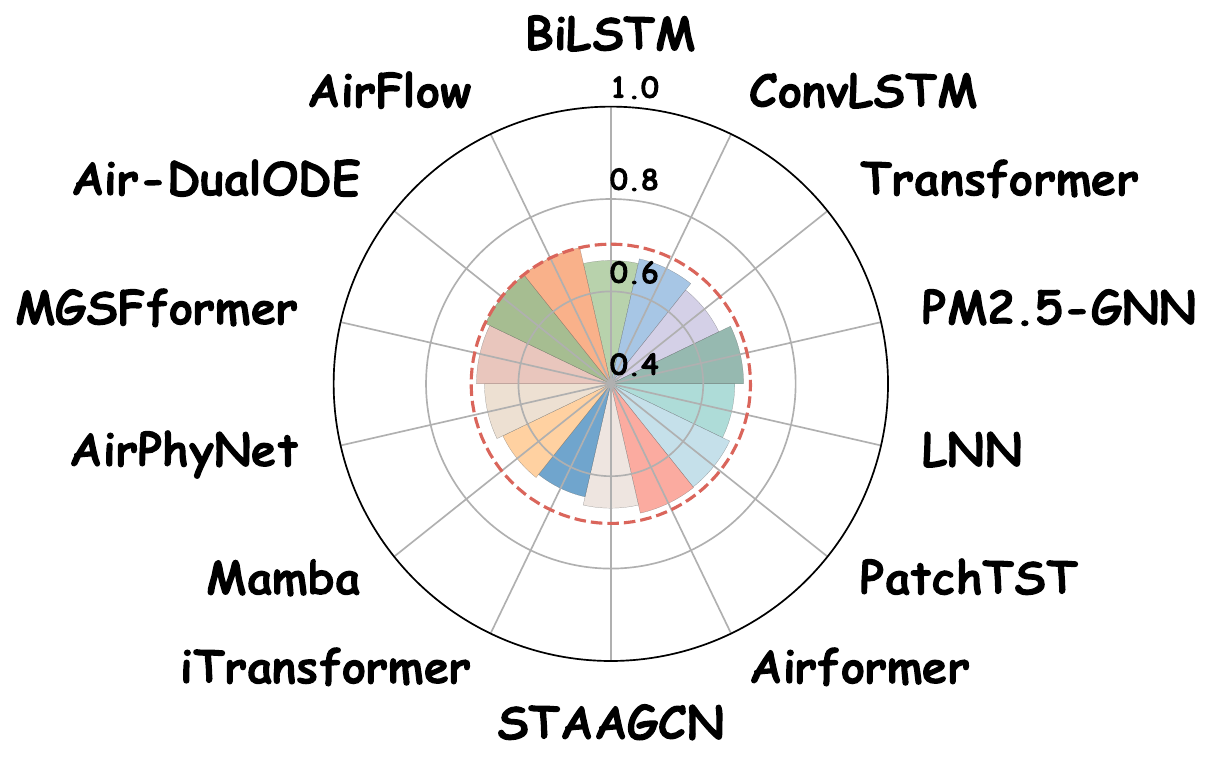}
    \end{subfigure}

    \vspace{0.15cm} 

    \begin{minipage}[c]{0.03\linewidth} 
        \centering
        \includegraphics[width=\linewidth]{hangzhou_sites.pdf}
    \end{minipage}%
    \hfill
    \begin{subfigure}[c]{\subfigwidth}
        \includegraphics[width=\linewidth]{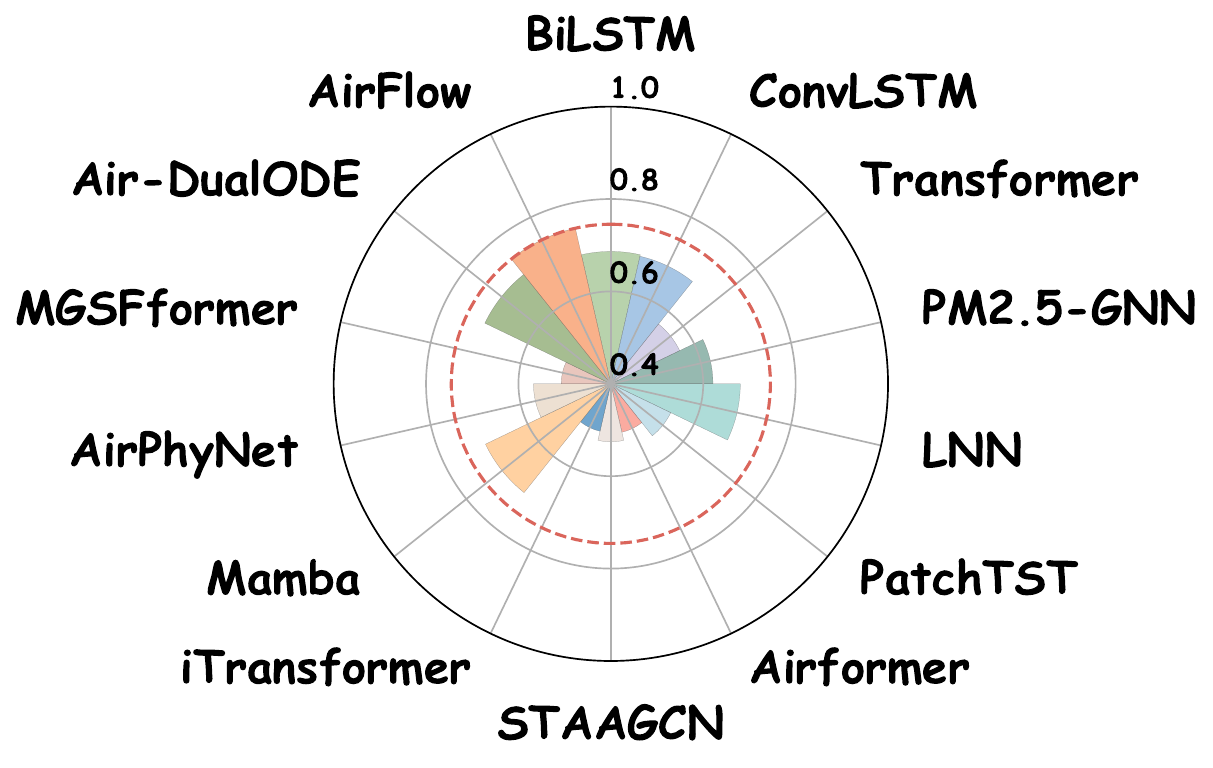}
    \end{subfigure}%
    \hfill
    \begin{subfigure}[c]{\subfigwidth}
        \includegraphics[width=\linewidth]{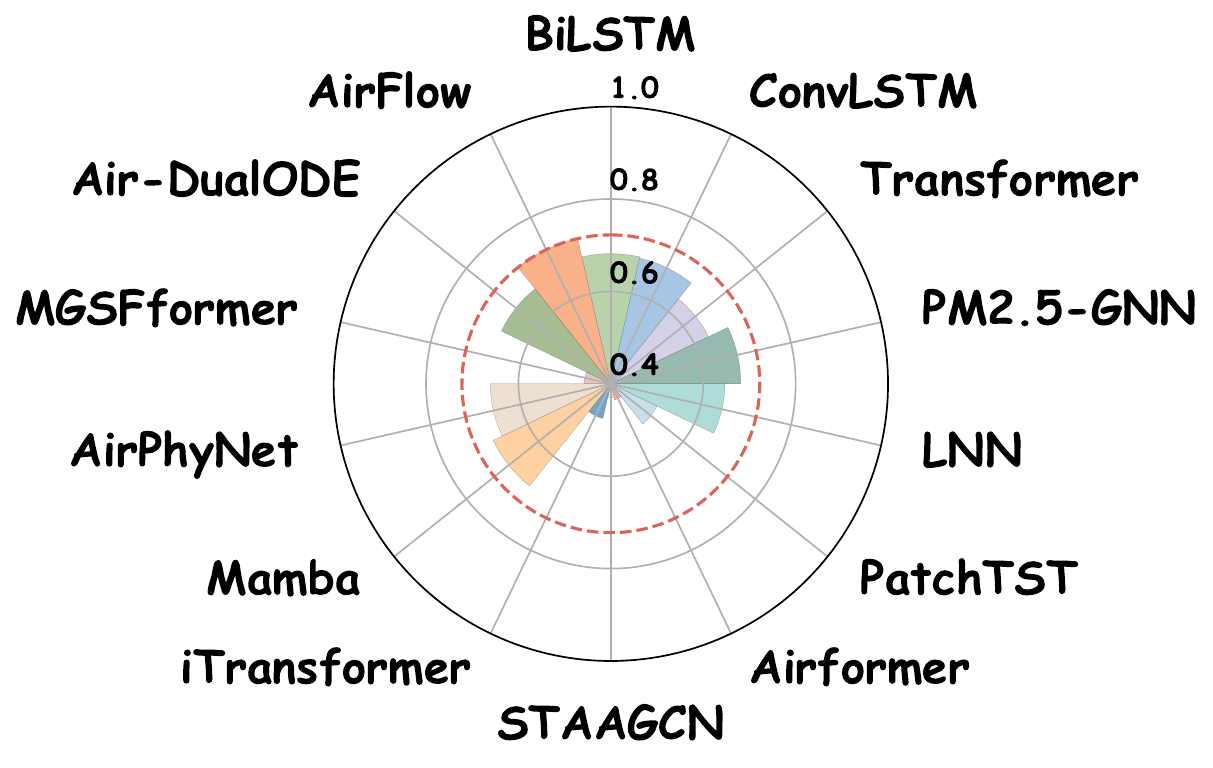}
    \end{subfigure}%
    \hfill
    \begin{subfigure}[c]{\subfigwidth}
        \includegraphics[width=\linewidth]{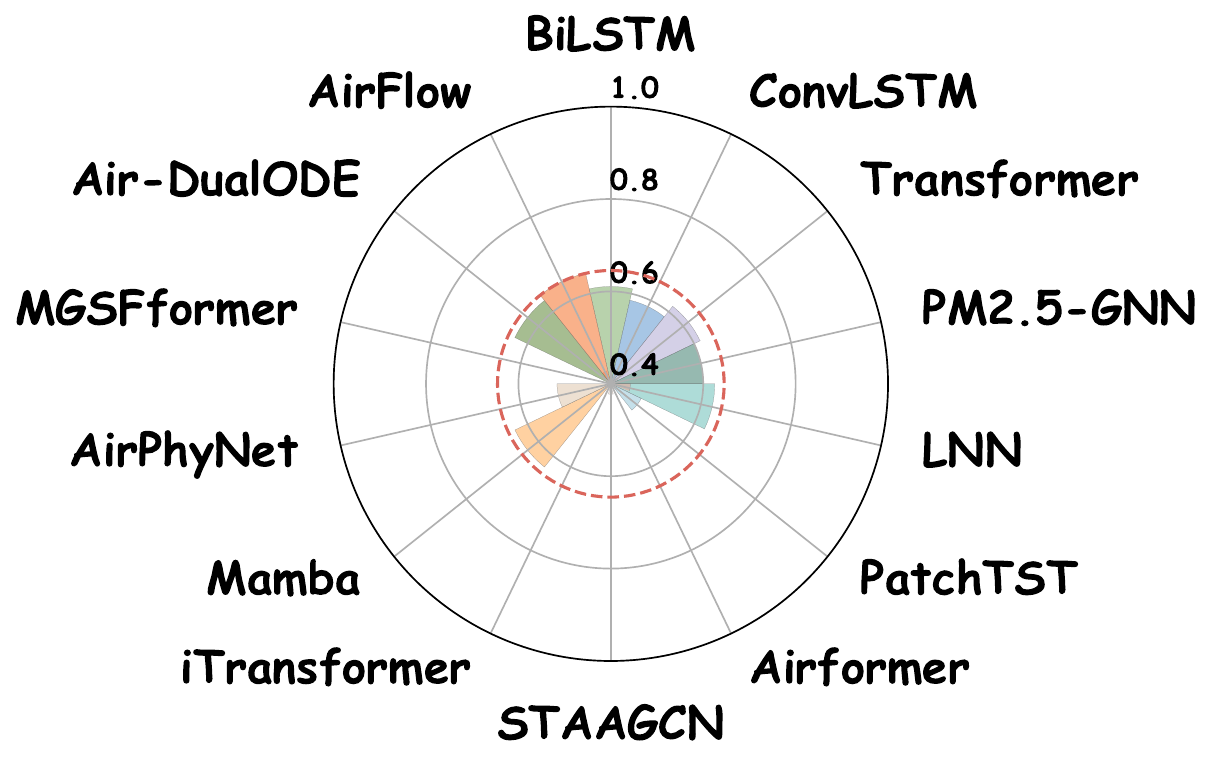}
    \end{subfigure}%
    \hfill
    \begin{subfigure}[c]{\subfigwidth}
        \includegraphics[width=\linewidth]{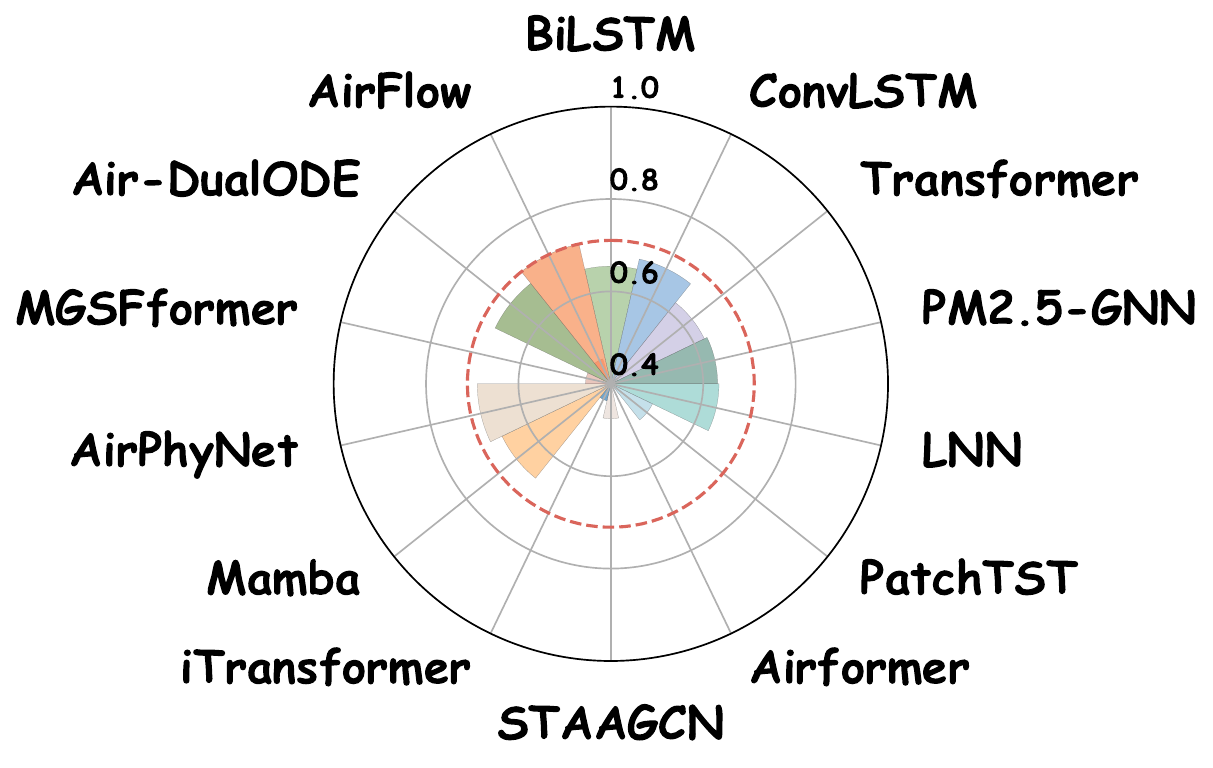}
    \end{subfigure}
    
    \caption{Comparative performance of baseline and proposed models based on R$^2$.}
    \label{fig:r2}
\end{figure*}

The figures presented in this section provide an intuitive visual comparison of model performance across different pollutants, evaluation metrics, and datasets.
Figures~\ref{fig:mae} and~\ref{fig:r2} visualize the comparative RMSE, MAE, and R$^2$ results for each model on three datasets, where the red dashed reference line marks the performance of AirFlow.

\end{document}